\documentclass[10pt,twocolumn,letterpaper]{article}

\usepackage{iccv}
\usepackage{times}
\usepackage{epsfig}
\usepackage{graphicx}
\usepackage{amsmath}
\usepackage{amssymb}
\usepackage{booktabs}
\usepackage{subfig}
\usepackage{tabularx}
\usepackage{rotating}

\makeatletter
\@namedef{ver@everyshi.sty}{}
\makeatother
\usepackage{tikz}
\usepackage{pgf, pgfsys, pgffor}
\usepackage{pgfplots}
\usepackage{multirow}
\usepackage{dsfont}

\newcolumntype{Y}{>{\centering\arraybackslash}X}
\newcommand{\eat}[1]{}
\definecolor{linkcolor}{rgb}{0.92, 0.09, 0.55}
\usepackage[breaklinks=true,bookmarks=false,colorlinks,urlcolor=linkcolor]{hyperref}

\iccvfinalcopy 

\begin{document}
\title{Spatio-Temporal Filter Adaptive Network for Video Deblurring}
\author{Shangchen Zhou$^1$\thanks{Equal contribution\ \ \ $^\dagger$Corresponding author: sdluran@gmail.com.}\ \ \ \ Jiawei Zhang$^1$$^*$\ \ \ Jinshan Pan$^2$$^\dagger$\ \ \ Haozhe Xie$^{1,3}$\ \ \ Wangmeng Zuo$^3$\ \ \ Jimmy Ren$^1$\\
$^1$SenseTime Research\quad
$^2$Nanjing University of Science and Technology, Nanjing, China\\
$^3$Harbin Institute of Technology, Harbin, China\\
{\tt\small \url{https://shangchenzhou.com/projects/stfan}}
}
\maketitle
\thispagestyle{empty}
\begin{abstract}
Video deblurring is a challenging task due to the spatially variant blur caused by camera shake, object motions, and depth variations, etc. 
Existing methods usually estimate optical flow in the blurry video to align consecutive frames or approximate blur kernels.
However, they tend to generate artifacts or cannot effectively remove blur when the estimated optical flow is not accurate.
To overcome the limitation of separate optical flow estimation, we propose a Spatio-Temporal Filter Adaptive Network (STFAN) for the alignment and deblurring in a unified framework.
The proposed STFAN takes both blurry and restored images of the previous frame as well as blurry image of the current frame as input, 
and dynamically generates the spatially adaptive filters for the alignment and deblurring. 
We then propose the new Filter Adaptive Convolutional (FAC) layer to align the deblurred features of the previous frame with the current frame and remove the spatially variant blur from the features of the current frame.
Finally, we develop a reconstruction network which takes the fusion of two transformed features to restore the clear frames.
Both quantitative and qualitative evaluation results on the benchmark datasets and real-world videos demonstrate that the proposed algorithm performs
favorably against state-of-the-art methods in terms of accuracy, speed as well as model size.
\end{abstract}
\vspace{-4mm}
\section{Introduction}
Recently, the hand-held and onboard video capturing devices have enjoyed widespread popularity, e.g., smartphone, action camera, unmanned aerial vehicle.
The camera shake and high-speed movement in dynamic scenes often generate undesirable blur and result in blurry videos.
The low-quality video not only leads to visually poor quality but also hampers some high-level vision tasks such as tracking~\cite{jin2005visual, mei2008modeling}, video stabilization~\cite{matsushita2006full} and SLAM~\cite{lee2011simultaneous}.
Thus, it is of great interest to develop an effective algorithm to deblur videos for above mentioned human perception and high-level vision tasks. 

\begin{figure}[t]\footnotesize
\centering
\renewcommand{\tabcolsep}{1.2pt}
\renewcommand{\arraystretch}{1}
\centering
\begin{tabular}{cccc}
	\includegraphics[width=0.235\linewidth]{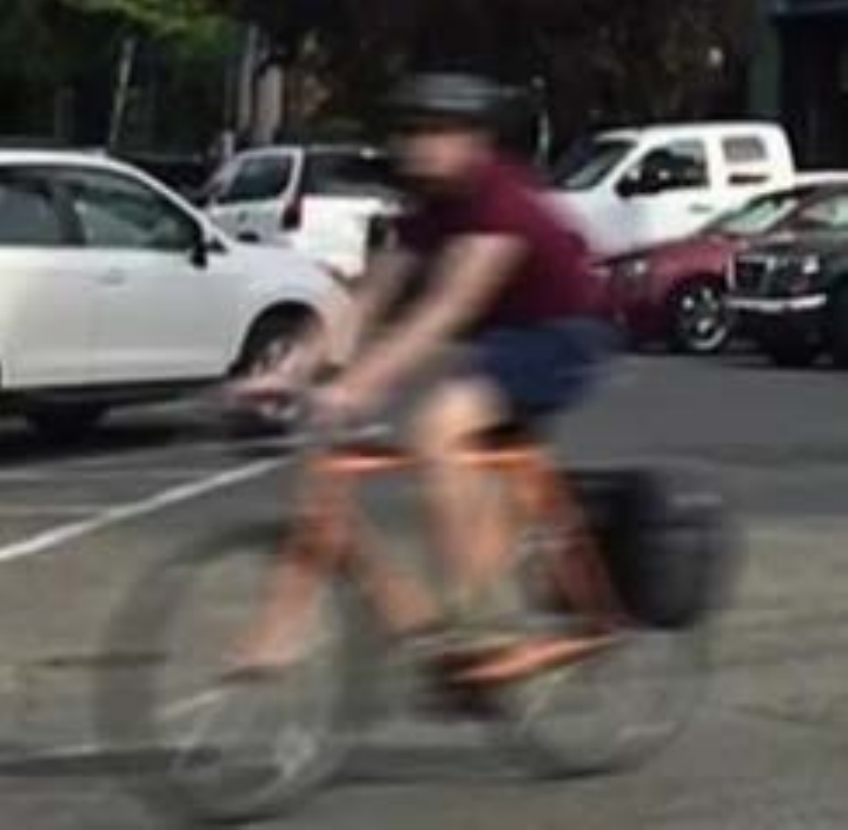} &
	\includegraphics[width=0.235\linewidth]{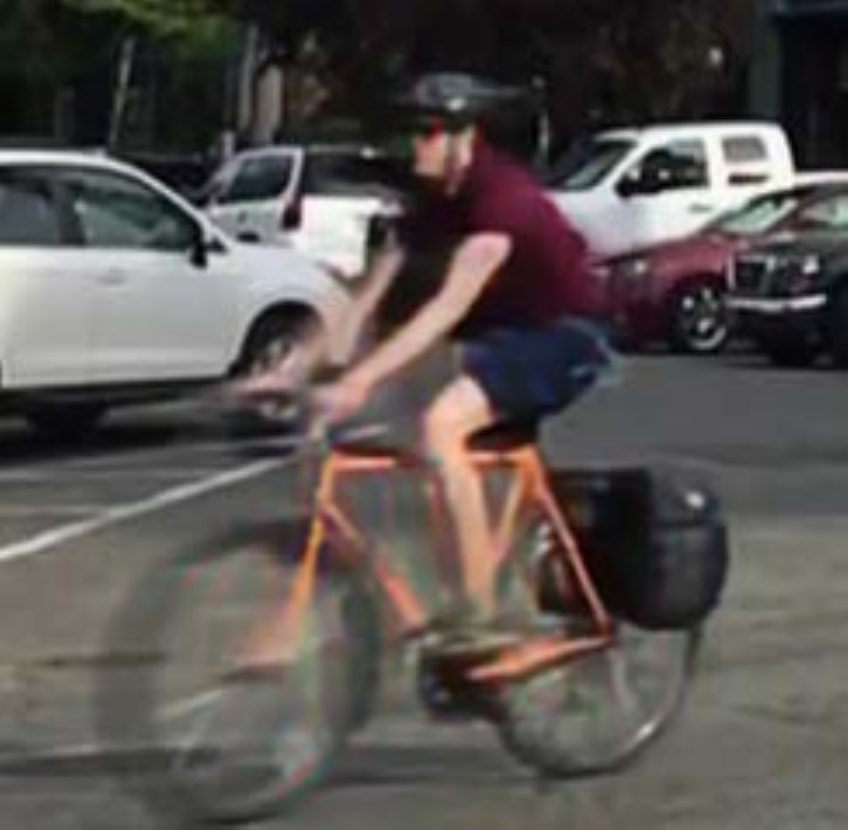} &
	\includegraphics[width=0.235\linewidth]{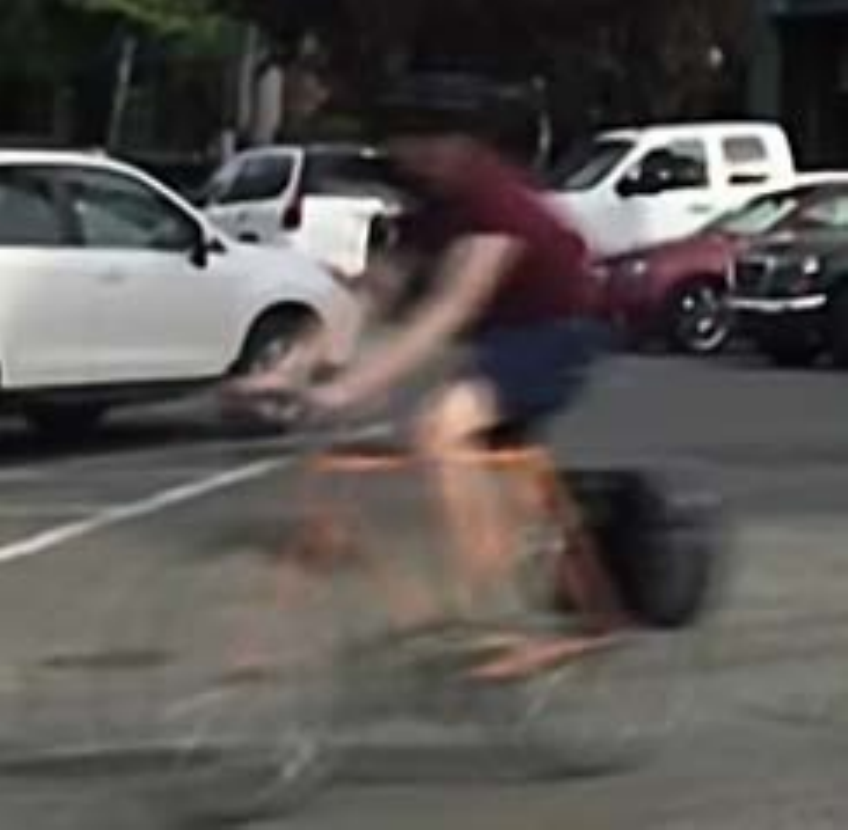} &
	\includegraphics[width=0.235\linewidth]{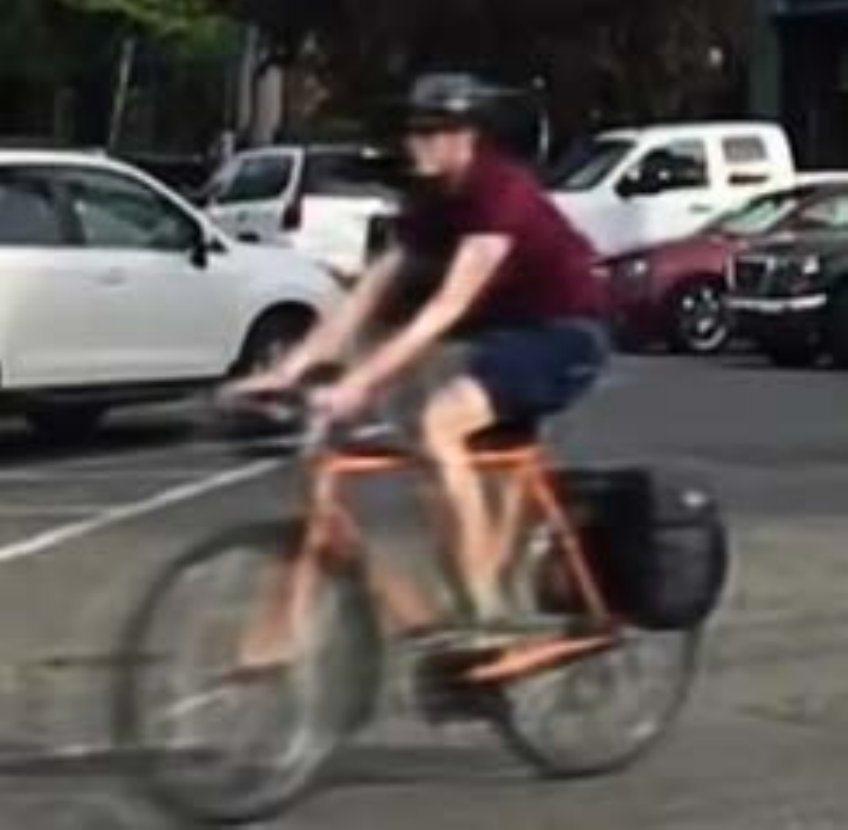}\\
	(a) Blurry frame & (b) SRN~\cite{tao2018scale}  & (c) GVD~\cite{hyun2015generalized} & (d) OVD~\cite{hyun2017online}
	\vspace{1.5pt}\\
	\includegraphics[width=0.235\linewidth]{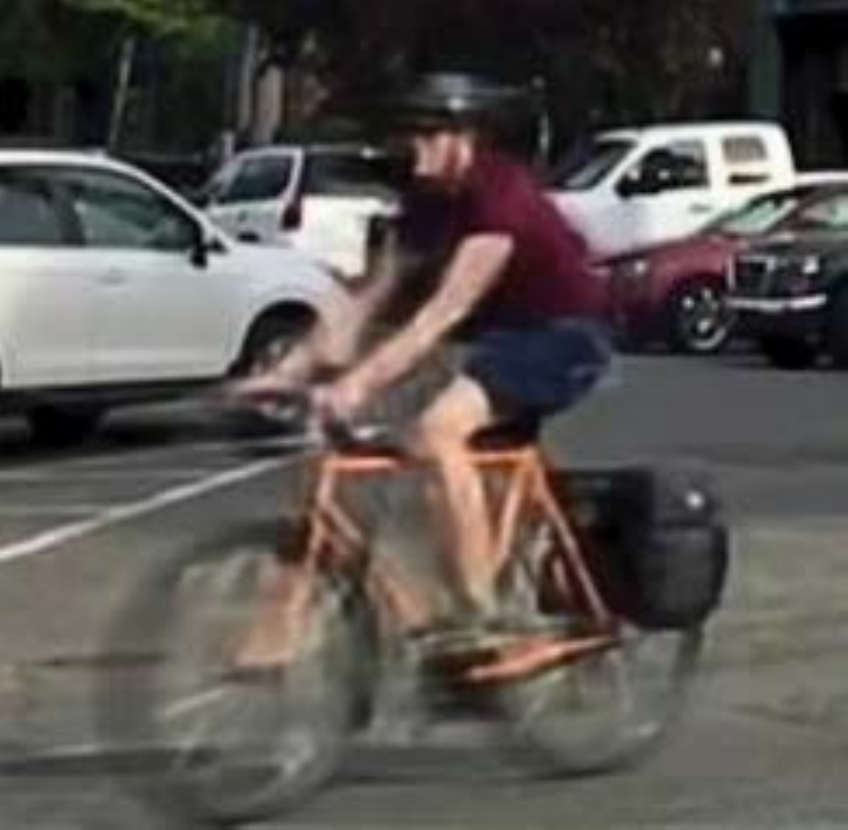} &
	\includegraphics[width=0.235\linewidth]{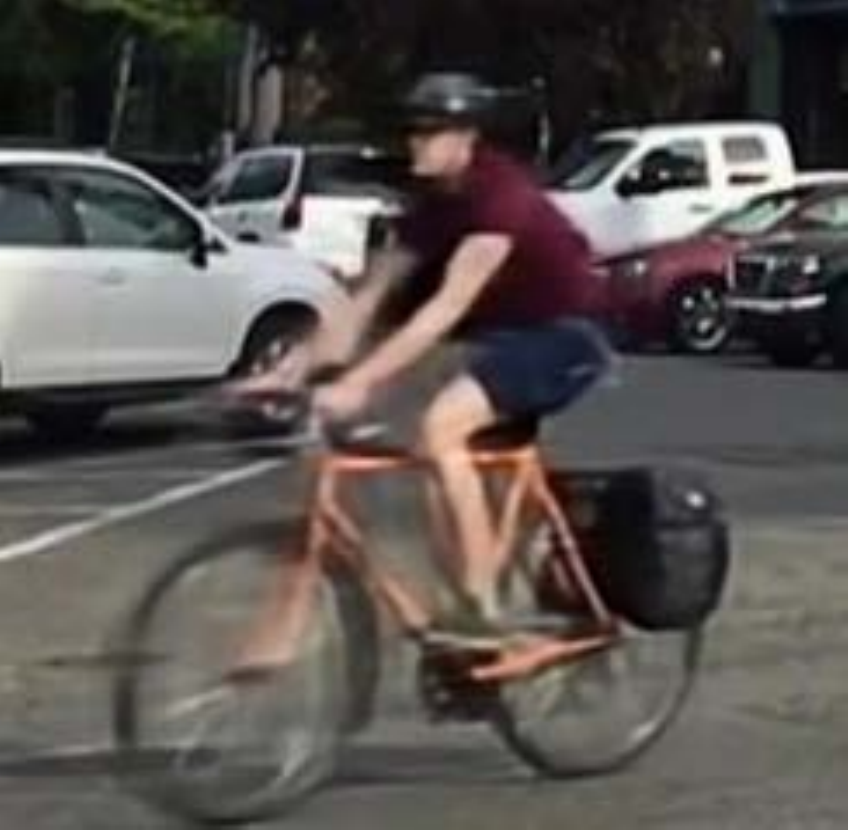} &
	\includegraphics[width=0.235\linewidth]{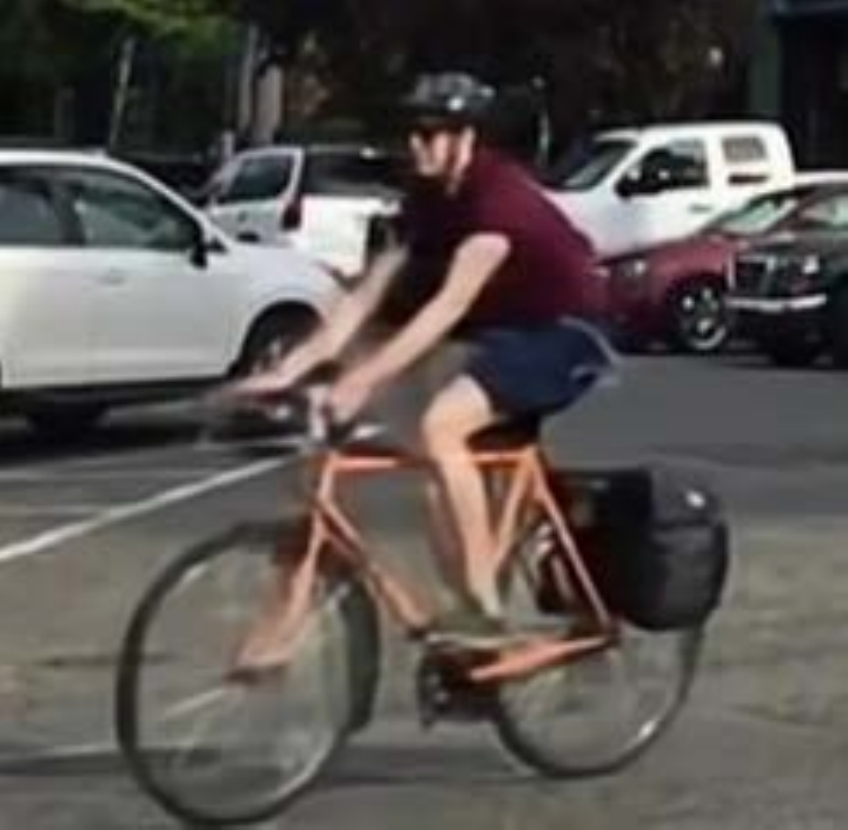} &
	\includegraphics[width=0.235\linewidth]{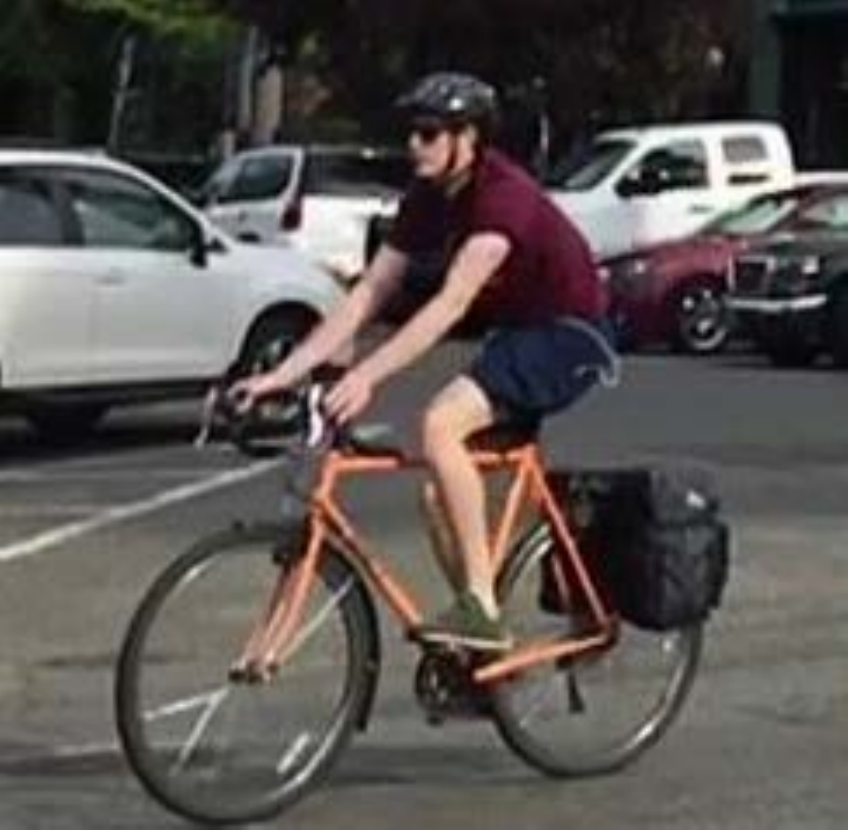}\\
	(e) DVD~\cite{su2017deep} & (f) w/o FAC & (g) Ours & (h) Ground truth\\
\end{tabular}
\caption{
	One challenging example of video deblurring.
	Due to the large motion and spatially variant blur, the existing image (b)~\cite{tao2018scale} and video deblurring (c, d, e)~\cite{hyun2015generalized,hyun2017online,su2017deep} methods are less effective.
	By using the proposed filter adaptive convolutional (FAC) layer for frame alignment and deblurring, our method generates a much clearer image.
	When the FAC layers are removed (f), our method cannot perform well anymore.}
\label{fig:head_pic}
\end{figure}

Unlike single-image deblurring, video deblurring methods can exploit additional information that exists across neighboring frames.
Significant progress has been made due to the use of sharper regions from neighboring frames~\cite{matsushita2006full, cho2012video} or the optical flow from consecutive frames~\cite{hyun2015generalized,ren2017video}.
However, directly utilizing sharp regions of surrounding frames usually generates significant artifacts because the neighboring frames are not fully aligned. 
Although using the motion field from two adjacent frames, such as optical flow, is able to overcome the alignment problem or approximate the non-uniform blur kernels, 
the estimation of motion field from blurry adjacent frames is quite challenging.

Motivated by the success of the deep neural networks in low-level vision, several algorithms have been proposed to solve video deblurring~\cite{hyun2017online,su2017deep}. 
Kim~\etal~\cite{hyun2017online} concatenate the multi-frame features to restore the current image by a deep recurrent network. However, this method fails to make full use of the information of neighboring frames without explicitly considering alignment,
and cannot perform well when the videos contain large motion.
Su \textit{et al.}~\cite{su2017deep} align the consecutive frames to the reference frame. 
It shows that this method performs well when the input frames are not too blurry but are less effective for the frames containing severe blur.
We also empirically find that both alignment and deblurring are crucial for deep networks to restore sharper frames from blurry videos.

Another group of methods~\cite{gong2017motion, sun2015learning, hyun2014segmentation, hyun2015generalized} use single or multiple images to estimate optical flow which is treated as the approximation of non-uniform blur kernels.
With the estimated optical flow, these methods usually use the existing non-blind deblurring algorithms (\eg,~\cite{zoran2011learning}) to reconstruct the sharp images.
However, these methods highly depend on the accuracy of the optical flow field.
In addition, these methods can only predict line-shaped blur kernel which is inaccurate under some scenarios.
To handle non-uniform blur in dynamic scenes, Zhang \text{et al.}~\cite{zhang2018dynamic} develop the spatially variant recurrent neural network (RNN)~\cite{liu2016learning} for image deblurring, whose pixel-wise weights are learned from a convolutional neural network (CNN). This algorithm does not need additional non-blind deblurring algorithms.
However, it is limited to single image deblurring and cannot be directly extended to video deblurring.

To overcome the above limitations, we propose a Spatio-Temporal Filter Adaptive Network (STFAN) for video deblurring.
Motivated by dynamic filter networks~\cite{jia2016dynamic, niklaus2017video, mildenhall2018burst} which apply the generated filters to the input images,
we propose the element-wise filter adaptive convolutional (FAC) layer.
Compared with~\cite{jia2016dynamic, niklaus2017video, mildenhall2018burst}, FAC layer applies the generated spatially variant filters on down-sampled features, which allows it to obtain a larger receptive field using a smaller filter size.
It also has stronger capability and flexibility due to different filters are dynamically estimated for different channels of the features.
The proposed method formulates the alignment and deblurring as two element-wise filter adaptive convolution processes in a unified network.
Specifically, given both blurry and restored images of the previous frame and blurry image of the current frame,
STFAN dynamically generates corresponding alignment and deblurring filters for feature transformation.
In contrast with estimating non-uniform blur kernels from a single blurry image~\cite{zhang2018dynamic, gong2017motion, sun2015learning, hyun2014segmentation} or two adjacent blurry images~\cite{hyun2015generalized}, 
our method estimates the deblurring filters from a richer inputs: three images and the motion information of two adjacent frames obtained from alignment filters.
By using FAC layer, STFAN adaptively aligns the features obtained at different time steps, without explicitly estimating optical flow and warping images, 
thereby leading to a tolerance of alignment accuracy.
In addition, the FAC layers allow our network handle spatially variant blur better, with deblurring in the feature domain.
An example in Figure~\ref{fig:head_pic} shows that our method generates a much sharper image (Figure~\ref{fig:head_pic}(g)) than our baseline without FAC layers (Figure~\ref{fig:head_pic}(f)) as well as the competing methods.

The main contributions are summarized as follows:
\vspace{-3pt}
\begin{itemize}
\setlength{\itemsep}{3pt}
\setlength{\parsep}{0pt}
\setlength{\parskip}{3pt}
\item We propose a filter adaptive convolutional (FAC) layer that applies the generated element-wise filters to feature transformation, which is utilized for two spatially variant tasks, i.e. alignment and deblurring in the feature domain.
\item We propose a novel spatio-temporal filter adaptive network (STFAN) for video deblurring. It integrates the frame alignment and deblurring into a unified framework without explicit motion estimation and formulates them as two spatially variant convolution process based on the FAC layers.
\item We quantitatively and qualitatively evaluate our network on benchmark dataset and show that it performs favorably against state-of-the-art algorithms in terms of accuracy, speed as well as model size.
\end{itemize}
\section{Related Work}
Our work formulates the neighboring frame alignment and non-uniform blur removal in video deblurring task as two element-wise filter adaptive convolution processes.
The following is a review of relevant works on single-image deblurring, multi-image deblurring, and kernel prediction network, respectively.

\noindent \textbf{Single-Image Deblurring.}
Numerous methods have been proposed for single-image deblurring. Early researchers assume a uniform blur kernel and design some natural image priors, such as $L_0$-regularized prior~\cite{xu2013unnatural}, dark channel prior~\cite{pan2016blind}, to compensate for the ill-posed blur removal process.
However, it is hard for these methods to model spatially-varying blur under dynamic scenes.
To model the non-uniform blur, the method \cite{hyun2013dynamic} and \cite{pan2016soft} estimate different blur kernels for different segmented the image patches.
Other works~\cite{gong2017motion, sun2015learning, hyun2014segmentation} estimate a dense motion field and a pixel-wise blur kernel.
\begin{figure*}[!]
\centering
\resizebox{0.97\linewidth}{!} {
	\includegraphics{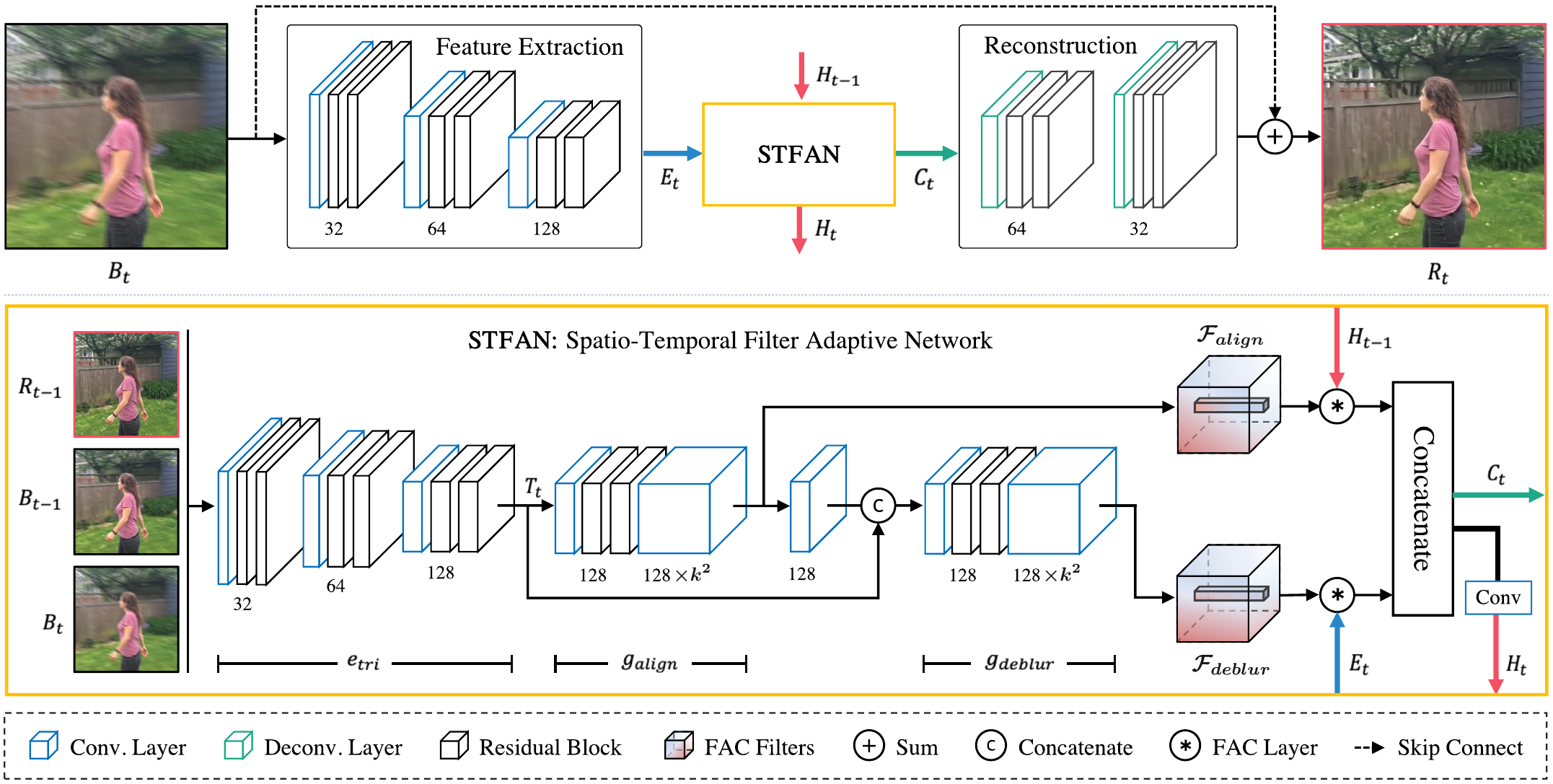}
}
\vspace{0mm}
\caption{Proposed network structure. It contains three sub-networks: spatio-temporal filter adaptive network (STFAN), feature extraction network, and reconstruction network.
	Given the triplet images (blurry $B_{t-1}$ and restored $R_{t-1}$ image of the previous frame, and current input image $B_t$),
	the sub-network STFAN generates the alignment filters $\mathcal{F}_{align}$ and deblurring filters $\mathcal{F}_{deblur}$ in order.
	Then, using the proposed FAC layer $\circledast$, STFAN aligns deblurred features $H_{t-1}$ of the previous time step with the current time step and removes blur from the features $E_t$ extracted from the current blurry image by the feature extraction network.
	At last, the reconstruction network is utilized to restore the sharp image from the fused features $C_t$.
	$k$ denotes the filter size of FAC layer.}
\label{fig:networks}
\vspace{-1mm}
\end{figure*}

With the development of deep learning, many CNN-based methods have been proposed to solve dynamic scene deblurring.
Method \cite{sun2015learning} and \cite{gong2017motion} utilize CNNs to estimate the non-uniform blur kernels. However, the predicted kernels are line-shaped which are inaccurate in some scenarios, 
and time-consuming conventional non-blind deblurring~\cite{zoran2011learning} is generally required to restore the sharp image.
More recently, many end-to-end CNN models~\cite{tao2018scale, zhang2018dynamic, kupyn2018deblurgan, nah2017deep, noroozi2017motion} have also been proposed for image deblurring.
To obtain a large receptive field for handling the large blur, the multi-scale strategy is used in~\cite{tao2018scale,nah2017deep}.
In order to deal with dynamic scene blur, Zhang \textit{et al.}~\cite{zhang2018dynamic} use spatially variant RNNs~\cite{liu2016learning} to remove blur in feature space with a generated RNN weights by a neural network.
However, compared with the video-based method, the accuracy of RNN weights is highly limited to having only a single blurry image as input.
To reduce the difficulty of restoration and ensures color consistency, Noroozi \textit{et al.}~\cite{noroozi2017motion} build skip connections between the input and output.
The adversarial loss is used in~\cite{nah2017deep, kupyn2018deblurgan} to generate sharper images with more details.

\noindent \textbf{Multi-Image Deblurring.}
Many methods utilize multiple images to solve dynamic scene deblurring from video, burst or stereo images.
The algorithms by \cite{wulff2014modeling} and \cite{ren2017video} use the predicted optical flow to segment layers with different blur and estimate the blur layer-by-layer.
In addition, Kim \textit{et al.}~\cite{hyun2015generalized} treat optical flow as a line-shaped approximation of blur kernels, which optimize optical flow and blur kernels iteratively.
The stereo-based methods~\cite{xu2012depth, sellent2016stereo, pan2017simultaneous} estimate depth from stereo images, which is used to predict the pixel-wise blur kernels.
Zhou \textit{et al.}~\cite{zhou2019davanet} propose a stereo deblurring network with depth awareness and view aggregation.
To improve the generalization ability, Chen \textit{et al.}~\cite{chen2018reblur2deblur} propose an optical flow based reblurring step to reconstruct the blurry input, which is employed to fine-tune deblurring network via self-supervised learning.
Recently, several end-to-end CNN methods~\cite{su2017deep, hyun2017online, kim2018spatio} have been proposed for video deblurring.
After image alignment using optical flow, \cite{su2017deep} and \cite{kim2018spatio} aggregate information across the neighboring frames to restore the sharp images.
Kim \textit{et al.}~\cite{hyun2017online} apply a temporal recurrent network to propagate the features from the previous time step into those of the current one.
Despite the fact that motion can be the useful guidance for blur estimation, Aittala \textit{et al.}~\cite{aittala2018burst} propose a burst deblurring network in an order-independent manner by repeatedly exchanging the information between the features of the burst images.

\noindent \textbf{Kernel Prediction Network.}
Kernel (filter) prediction network (KPN) has recently witnessed rapid progress in low-level vision tasks.
Jia \textit{et al.}~\cite{jia2016dynamic} first propose the dynamic filter network, which consists of a filter prediction network that predicts kernels conditioned on an input image,
and a dynamic filtering layer that applies the generated kernels to another input.
Their method shows the effectiveness on video and stereo prediction tasks.
Niklaus \textit{et al.}~\cite{niklaus2017video} apply kernel prediction network to video frame interpolation, which merges optical flow estimation and frame synthesis into a unified framework.
To alleviate the demand for memories, they subsequently propose separable convolution~\cite{niklaus2017sep} which estimates two separable 1D kernels to approximate 2D kernels.
In \cite{mildenhall2018burst}, they utilize KPN for both burst frame alignment  and denoising, using the same predicted kernels.
\cite{jo2018deep} reconstructs high-resolution image from low-resolution input using generated dynamic upsampling filters.
However, all the above methods directly apply the predicted kernels (filters) in the image domain.
In addition, Wang \textit{et al.}~\cite{wang2018recovering} propose a spatial feature transform (SFT) layer for image super-resolution.
It generates transformation parameters for pixel-wise feature modulation, which can be considered as the KPN with a kernel size of $1\times 1$ in the feature domain.
\section{Proposed Algorithm}
In this section, we first give an overview of our algorithm in Sec.~\ref{sec:overview}.
Then we introduce the proposed filter adaptive convolutional (FAC) layer in Sec.~\ref{sec:kac}.
Upon this layer, we show the structure of the proposed networks in Sec.~\ref{sec:arch}.
Finally, we present the loss functions that are used to constrain the network training in Sec.~\ref{sec:loss}.
\subsection{Overview}\label{sec:overview} 
Different from the standard CNN-based video deblurring methods \cite{su2017deep, hyun2017online, kim2018spatio} that take five or three consecutive blurry frames as input to restore the sharp mid-frame,
we propose a frame-recurrent method, which requires information of the previous frame and the current input.
Due to the recurrent property, the proposed method is able to explore and utilize the information from a large number of previous frames without increasing the computational demands.
As shown in Figure~\ref{fig:networks}, the proposed STFAN generates the filters for alignment and deblurring from the triplet images (blurry and restored image of the previous time step $t-1$, and current input blurry image).
Then, using FAC layers, STFAN aligns the deblurred features from the previous time step with the current one and removes blur from the features  extracted from the current blurry image.
Finally, a reconstruction network is applied to restore the sharp image by fusing the above two transformed features.
\subsection{Filter Adaptive Convolutional Layer}\label{sec:kac}
\begin{figure}
\centering
\resizebox{0.9\linewidth}{!} {
	\includegraphics{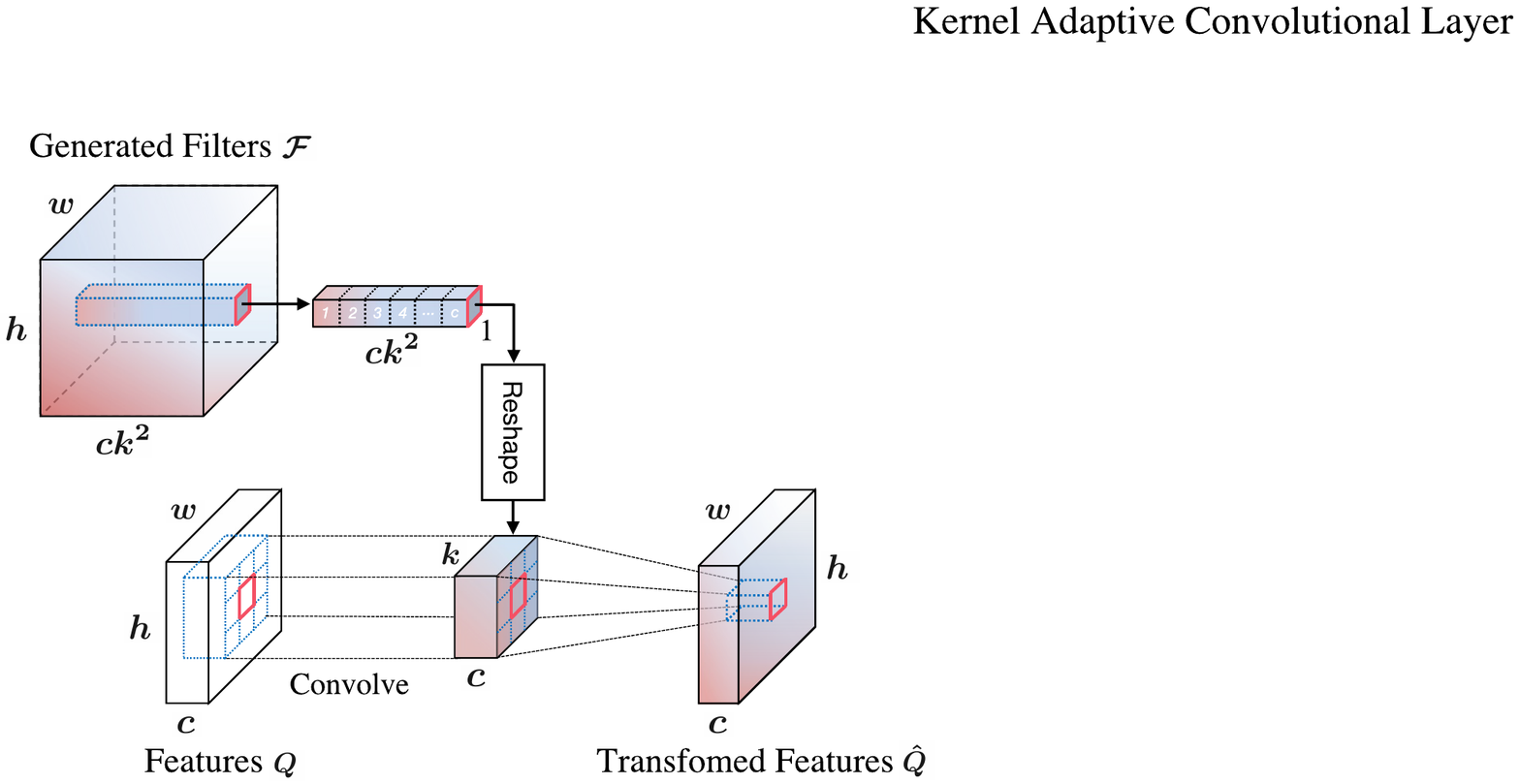}
}
\caption{Filter Adaptive Convolutional Layer}
\vspace{-8px}
\label{fig:kac}
\end{figure}

Motivated by the Kernel Prediction Network (KPN)~\cite{jia2016dynamic, niklaus2017video, mildenhall2018burst}, which applies the generated spatially variant filters to the input image,
we propose the filter adaptive convolutional (FAC) layer which applies generated element-wise convolutional filters to the features, as shown in Figure~\ref{fig:kac}.
The filters predicted in~\cite{jia2016dynamic, niklaus2017video, mildenhall2018burst} are the same for RGB channels of each position.
To be more capable and flexible for spatially variant tasks, the generated filters for FAC layer are different for each channel.
Limited by large memory demand, we only consider the convolution within channels.
In theory, the element-wise adaptive filters is five-dimensional ($h \times w \times c \times k \times k$).
In practice, the dimension of the generated filter $\mathcal{F}$ is $h\times w\times ck^2$ and we reshape it into the five-dimensional filter.
For each position $(x, y, c_i)$ of input feature $Q\in\mathds{R}^{h\times w\times c}$, a specific local filter $\mathcal{F}_{x, y, c_i}\in\mathds{R}^{k\times k}$ (reshape from $1\times 1\times k^2$) is applied to the region centered around $Q_{x, y, c_i}$ as follows:
\begin{align}\label{eq:kac}
\hat{Q}(x, y, c_i) &= \mathcal{F}_{x, y, c_i} \ast Q_{x, y, c_i}  \nonumber \\
&= \sum_{n=-r}^{r} \sum_{m=-r}^{r} \mathcal{F}(x, y, k^2c_i+kn+m) \nonumber \\
&\times Q(x-n, y-m, c_i),
\end{align}
where $r=\frac{k-1}{2}$, $\ast$ donates convolution operation, $\mathcal{F}$ is the generated filter, $Q(x, y, c_i)$ and $\hat{Q}(x, y, c_i)$ denote the input features and transformed features, respectively.
The proposed FAC layer is trainable and efficient, which is implemented and accelerated by CUDA.

A large receptive field is essential to handle large motions and blurs.
The standard KPN methods~\cite{jia2016dynamic, niklaus2017video, mildenhall2018burst} have to predict the filter much larger in size than motion blur for each pixel of the input image, which requires large computational cost and memory.
In contrast, the proposed network does not require a large filter size due to the use of FAC layer on down-sampled features. 
The experimental results in Table~\ref{tab:filter_size} show a small filter size (e.g. 5) on intermediate feature layer is sufficient for deblurring.
\subsection{Network Architecture}\label{sec:arch}
As shown in Figure~\ref{fig:networks}, our network is composed of a spatio-temporal filter adaptive network (STFAN), a feature extraction network, and a reconstruction network.

\noindent \textbf{Feature Extraction Network.}
This network extracts features $E_t$ from the blurry image $B_t$,
which consists of three convolutional blocks and each of them has one convolutional layer with stride 2 and two residual blocks~\cite{he2016deep} with LeakyReLU (negative slope $\lambda = 0.1$) as the activation functions.
The extracted features are feed into STFAN for deblurring using FAC layer.

\noindent \textbf{Spatio-Temporal Filter Adaptive Network.}
The proposed STFAN consists of three modules: encoder $e_{tri}$ of triplet images, alignment filter generator $g_{align}$, and deblurring filter generator $g_{deblur}$.

Given the triplet input: the blurry image $B_{t-1}$ and restored image $R_{t-1}$ of the previous frame and the current blurry image $B_{t}$, STFAN extracts features $T_t$ by the encoder $e_{tri}$.
The encoder consists of three convolutional blocks (kernel size 3) and each of them is composed of one convolutional layer with stride 2 and two residual blocks.
The alignment filter generator $g_{align}$ takes the extracted features $T_t$ of triplet images as input to predict the adaptive filters for alignment, denoted as $\mathcal{F}_{align}\in\mathds{R}^{h\times w\times ck^2}$:
\begin{equation}
\label{eq:align}
\mathcal{F}_{align} =  g_{align}(e_{tri}(B_{t-1}, R_{t-1}, B_{t})),
\end{equation}
where generated $\mathcal{F}_{align}$ contains rich motion information, which is helpful to model the non-uniform blur in the dynamic scene.
To make full use of it, the deblurring filter generator $g_{deblur}$ takes alignment filters $\mathcal{F}_{align}$ as well as the features $T$ of triplet images to generate the spatially variant filters for deblurring, denoted as $\mathcal{F}_{deblur}\in\mathds{R}^{h\times w\times ck^2}$:
\begin{equation}
\label{eq:deblur}
\mathcal{F}_{deblur} =  g_{deblur}(e_{tri}(B_{t-1}, R_{t-1}, B_{t}), \mathcal{F}_{align}),
\end{equation}
Both filter generators consist of one convolution layer and two residual blocks with kernel size $3\times 3$, followed by a $1\times 1$ convolution layer to expand the channels of output to $ck^2$.

With the two generated filters, two FAC layers are utilized to align the deblurred features $H_{t-1}$ from the previous time step with the current frame and remove the blur from the extracted features $E_t$ of current blurry frame in the feature domain.
After that, we concatenate these two transformed features as $C_t$ and restore the sharp image by the reconstruction network.
To propagate the deblurred information $H_{t}$ to the next time step, we pass the features $C_t$ to the next iteration through a convolutional layer.

It is worth noting that both the blurry $B_{t-1}, B_t$ and restored $R_{t-1}$ are required to learn the filters for alignment and deblurring, and thus are taken as the triplet input to STFAN.
On the one hand, $B_{t-1}$ and $B_t$ are crucial to capture the motion information across frames and thus benefit alignment. 
On the other hand, the inclusion of $B_{t-1}$ and $R_{t-1}$ makes it possible to implicitly exploit the blur kernel at frame $t-1$ for improving the deblurring at frame $t$.
Moreover, deblurring is assumed to be more difficult but can be benefited by alignment. Thus we stack $g_{deblur}$ upon $g_{align}$ in our implementation.
We will analyze the effect of taking triplet images $B_{t-1}, R_{t-1}, B_{t}$ as input in Sec.~\ref{sec: Effectiveness of the Triplet Input of STFAN}.

\noindent \textbf{Reconstruction Network.}
The reconstruction network is used to restore the sharp images by taking the fusion features from STFAN as input. 
It consists of scale convolutional blocks, each of which has one deconvolutional layer and two residual blocks as shown in Figure~\ref{fig:networks}.
\subsection{Loss Function}\label{sec:loss}
To effectively train the proposed network, we consider two kinds of loss functions.
The first loss is the mean squared error (MSE) loss that measures the differences between the restored frame $R$ and its corresponding sharp ground truth $S$:
\begin{equation}
\label{eq:mse_loss}
\mathcal{L}_{mse} =  \frac{1}{CHW} || R- S ||^2,
\end{equation}
where $C, H, W$ are dimensions of image, respectively; $R$ and $S$ respectively denote the restored image and the corresponding ground truth.

To generate more realistic images, we further use the perceptual loss proposed in~\cite{johnson2016perceptual}, which is defined as the Euclidean distance between the VGG-19~\cite{simonyan2015very} features of restored frame $R$ and ground truth $S$:
\begin{equation}
\label{eq:percept_loss}
\mathcal{L}_{perceptual} = \frac{1}{\mathcal{C}_j\mathcal{H}_j\mathcal{W}_j} ||\Phi_j(R) - \Phi_j(S)||^2,
\end{equation}
where $\Phi_j(\cdot)$ denotes the features from the $j$-th convolutional layer of the pretrained VGG-19 network and $\mathcal{C}_j, \mathcal{H}_j, \mathcal{W}_j$ are dimensions of features. In this paper, we use the features of conv3-3 ($j=15$).
The final loss function for the proposed network is defined as:
\begin{equation}
\label{eq:deblur_loss}
\mathcal{L}_{deblur} = \mathcal{L}_{mse} + \lambda\mathcal{L}_{perceptual},
\end{equation}
where the weight $\lambda$ is set as $0.01$ in our experiments.
\section{Experiments}
\subsection{Implementation Details}
In our experiments, we train the proposed network using the video deblurring dataset from~\cite{su2017deep}.
It contains 71 videos (6,708 blurry-sharp pairs), splitting into 61 training videos (5,708 pairs) and 10 testing videos (1,000 pairs).

\noindent \textbf{Data Augmentation.}
We perform several data augmentations for training.
We first divide each video into several sequences with length 20.
To add motion diversity into the training data, we reverse the order of sequence randomly.
For each sequence, we perform the same image transformations.
It consists of chromatic transformations such as brightness, contrast as well as saturation, which are uniformly sampled from [0.8,~1.2] and geometric transformations including randomly flipping horizontally and vertically and randomly cropping to $256\times256$ patches.
To make our network robust in real-world scenarios, a Gaussian random noise from $\mathcal{N}(0, 0.01)$ is added to the input images.
\begin{table*}
\centering
\caption{Quantitative evaluation on the video deblurring dataset~\cite{su2017deep}, in terms of PSNR, SSIM,
	running time (sec) and parameter numbers ($\times 10^6$) of different networks.
	All existing methods are evaluated using their publicly available code.
	`-' indicates that it is not available.}
\vspace{-1mm}
\resizebox{\linewidth}{!} {
	\begin{tabular}{lccccccc|cccc}
		\toprule
		Method
		& Whyte~\cite{whyte2012non}
		& Sun~\cite{sun2015learning}
		& Gong~\cite{gong2017motion}
		& Nah~\cite{nah2017deep}
		& Kupyn~\cite{kupyn2018deblurgan}
		& Zhang~\cite{zhang2018dynamic}
		& Tao~\cite{tao2018scale}
		& Kim~\cite{hyun2015generalized}
		& Kim~\cite{hyun2017online}
		& Su~\cite{su2017deep}
		& Ours\\
		\midrule
		Frame\#  & 1           & 1           & 1           & 1           & 1           & 1            & 1             & 3             & 5          & 5          & 2  \\
		PSNR     & 25.29      & 27.24      & 28.22      & 29.51      & 26.78     & 30.05  & 29.97  &  27.01  & 29.95  & 30.05 & \bf{31.24}\\
		SSIM      & 0.832      & 0.878      & 0.894      & 0.912    & 0.848      & 0.922   & 0.919   &  0.861 & 0.911 &  0.920  &  \bf{0.934}\\
		\midrule
		Time (sec)     & 700      & 1200      & 1500       & 4.78           & 0.22       & 1.40     & 2.52     &  880 &  0.13  &  6.88  & 0.15\\
		Params (M)    & -          & 7.26        & 10.29      & 11.71     & 11.38      & 9.22      & 8.06    &  - &  0.92  & 16.67  & 5.37\\
		\bottomrule
	\end{tabular}
}
\label{tab:psnr_time_size}
\end{table*}
\begin{figure*}[h]\footnotesize
\centering
\renewcommand{\tabcolsep}{1pt}
\renewcommand{\arraystretch}{1}
\begin{center}
	\begin{tabular}{ccccc}
		\includegraphics[width=0.192\linewidth]{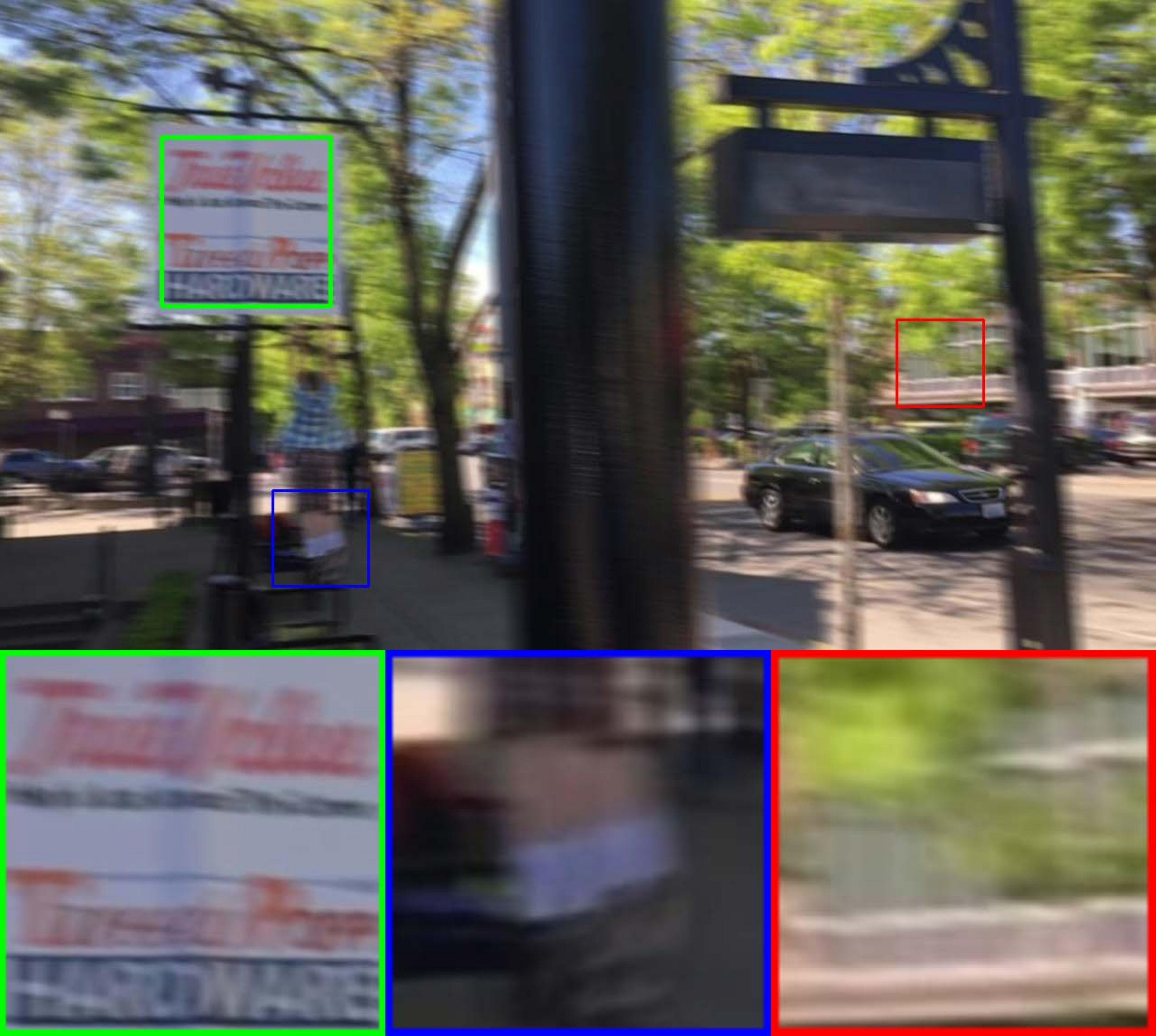} &
		\includegraphics[width=0.192\linewidth]{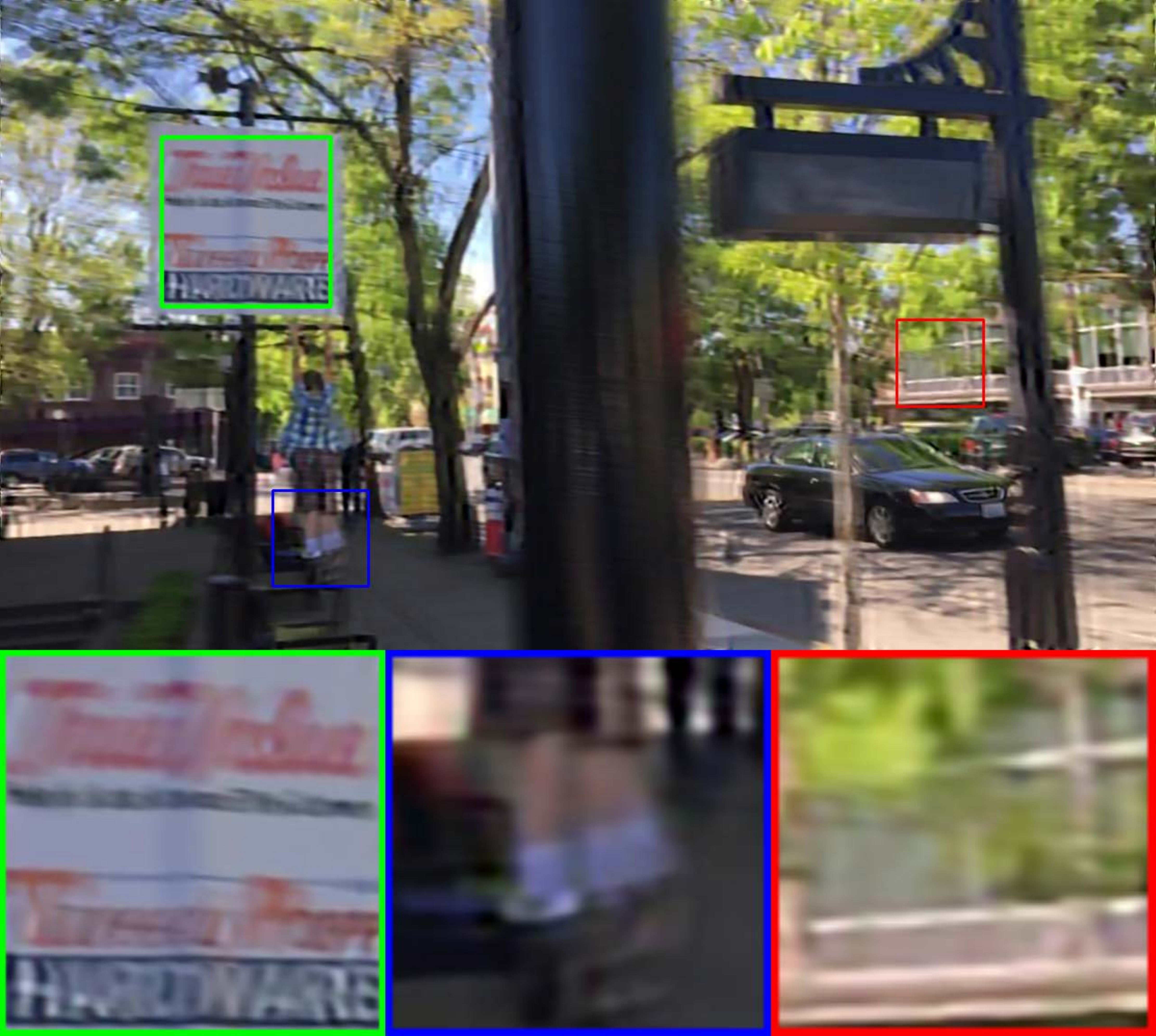} &
		\includegraphics[width=0.192\linewidth]{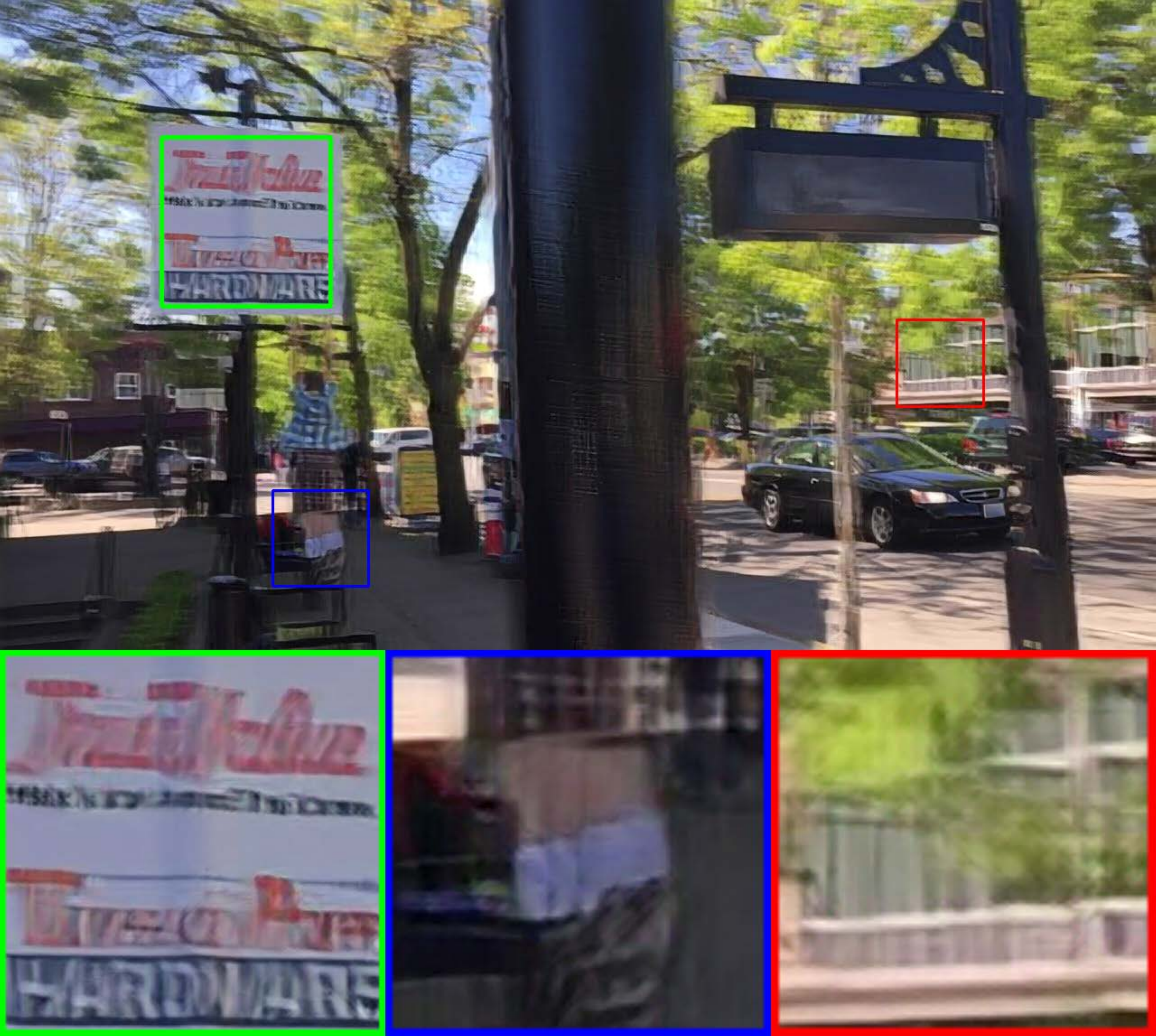} &
		\includegraphics[width=0.192\linewidth]{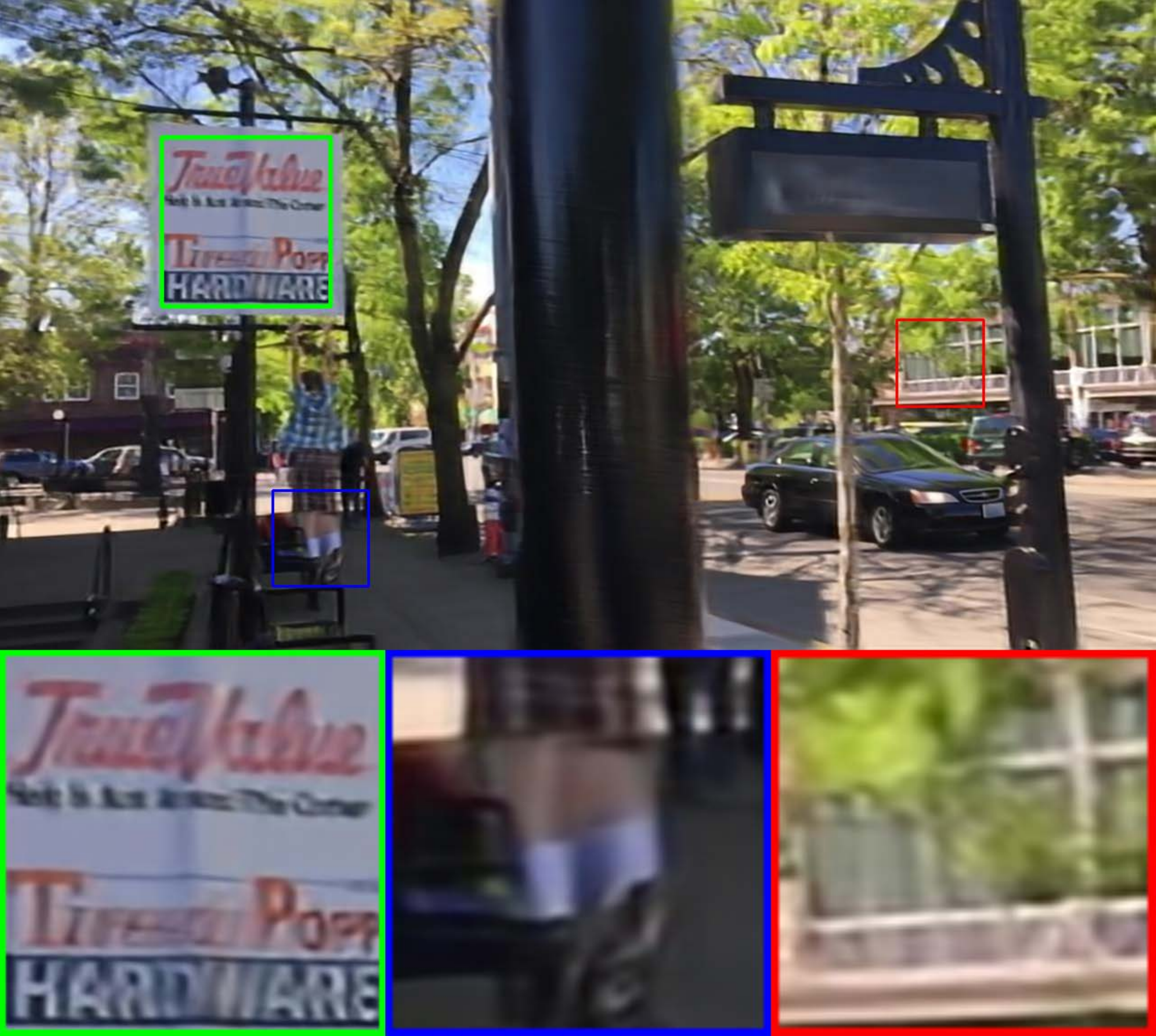} &	
		\includegraphics[width=0.192\linewidth]{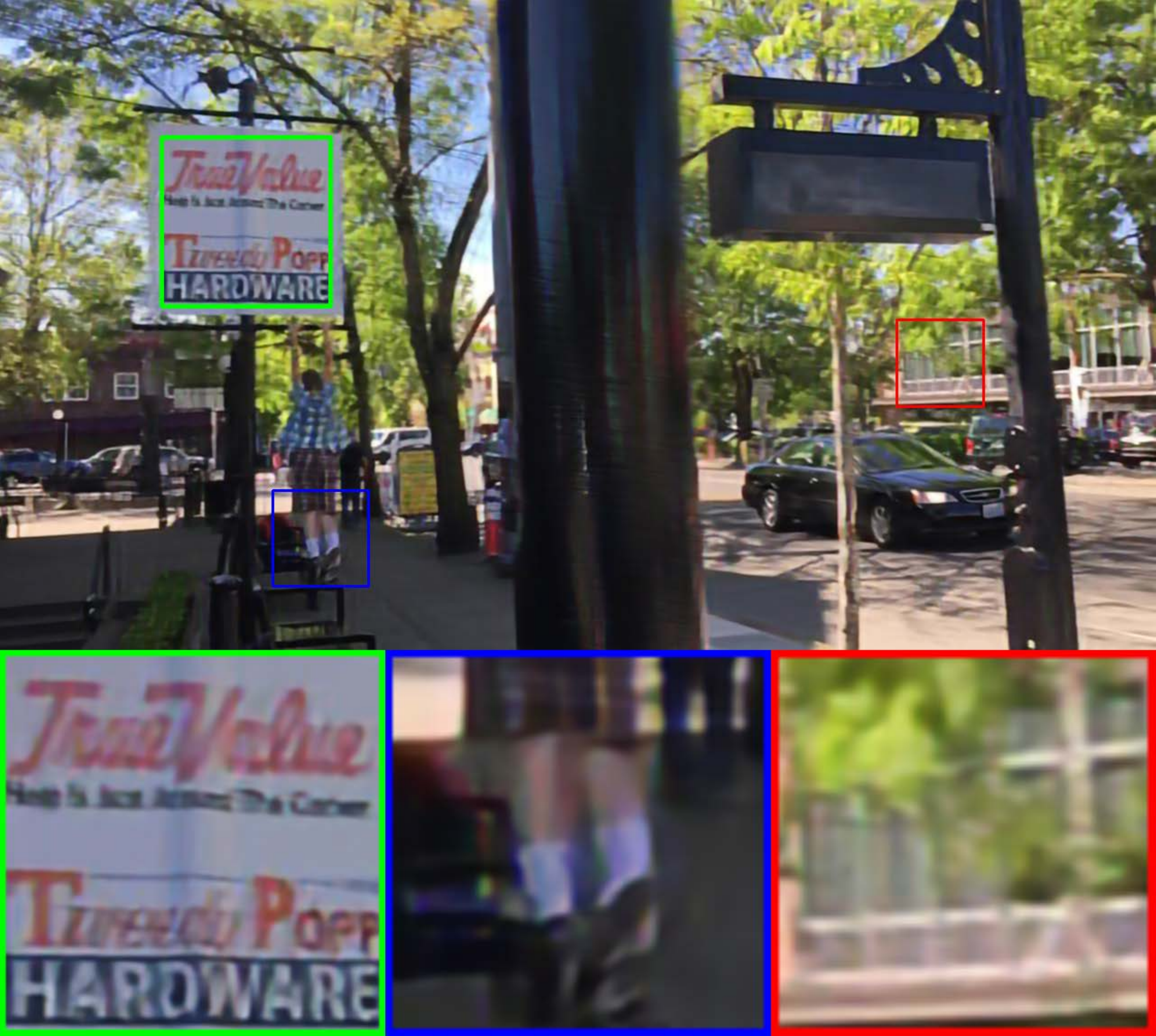} \\		
		(a) Blurry image & (b) Gong \textit{et al.}~\cite{gong2017motion} & (c) Kupyn \textit{et al.}~\cite{kupyn2018deblurgan} & (d) Zhang \textit{et al.}~\cite{zhang2018dynamic}& (e) Tao \textit{et al.}~\cite{tao2018scale} \\
		PSNR / SSIM  &  22.72 / 0.7911  & 21.22 / 0.7189  &  23.92 / 0.8321 & 25.29 / 0.8533
		\vspace{1.5pt}\\
		\includegraphics[width=0.192\linewidth]{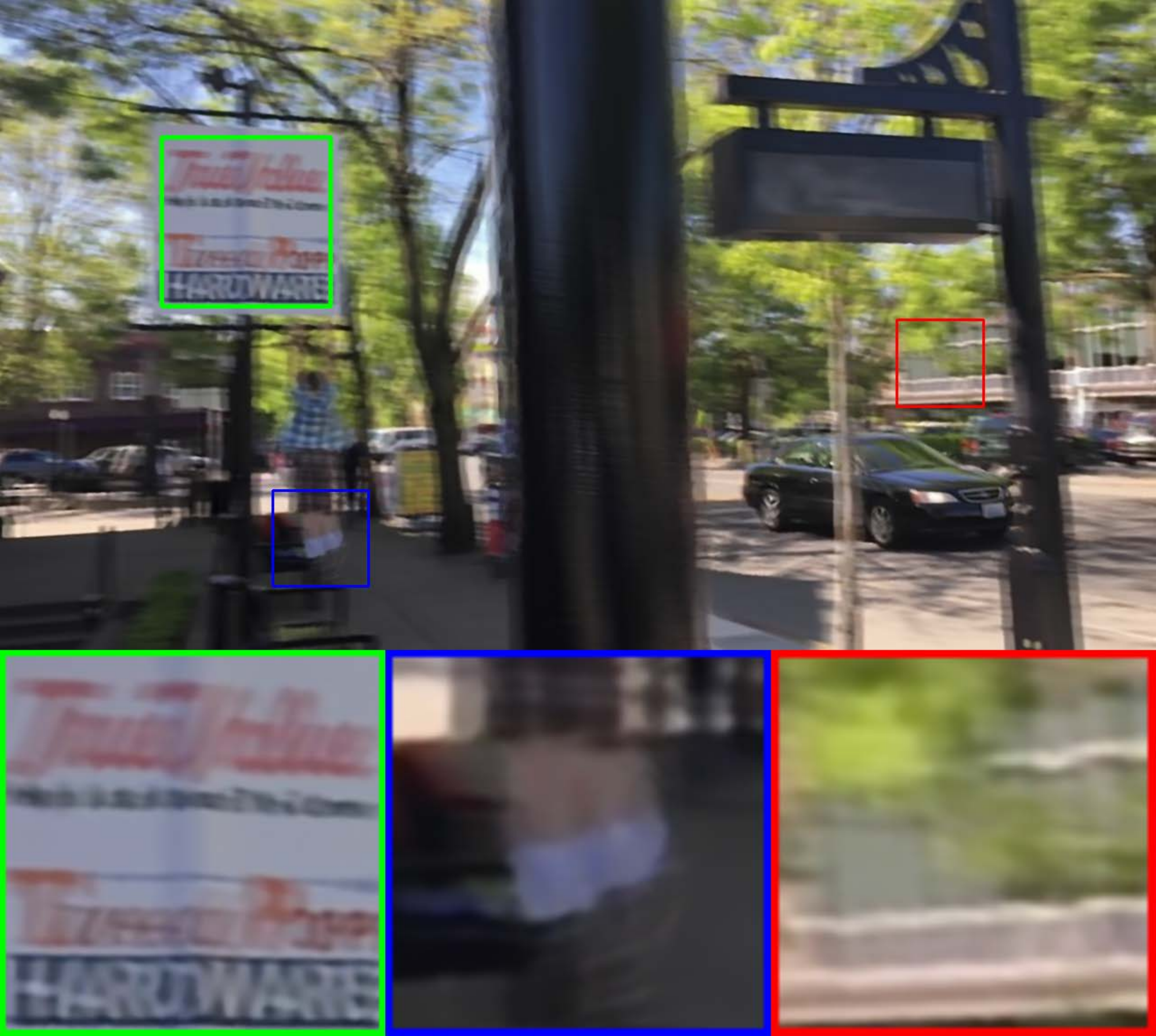} &
		\includegraphics[width=0.192\linewidth]{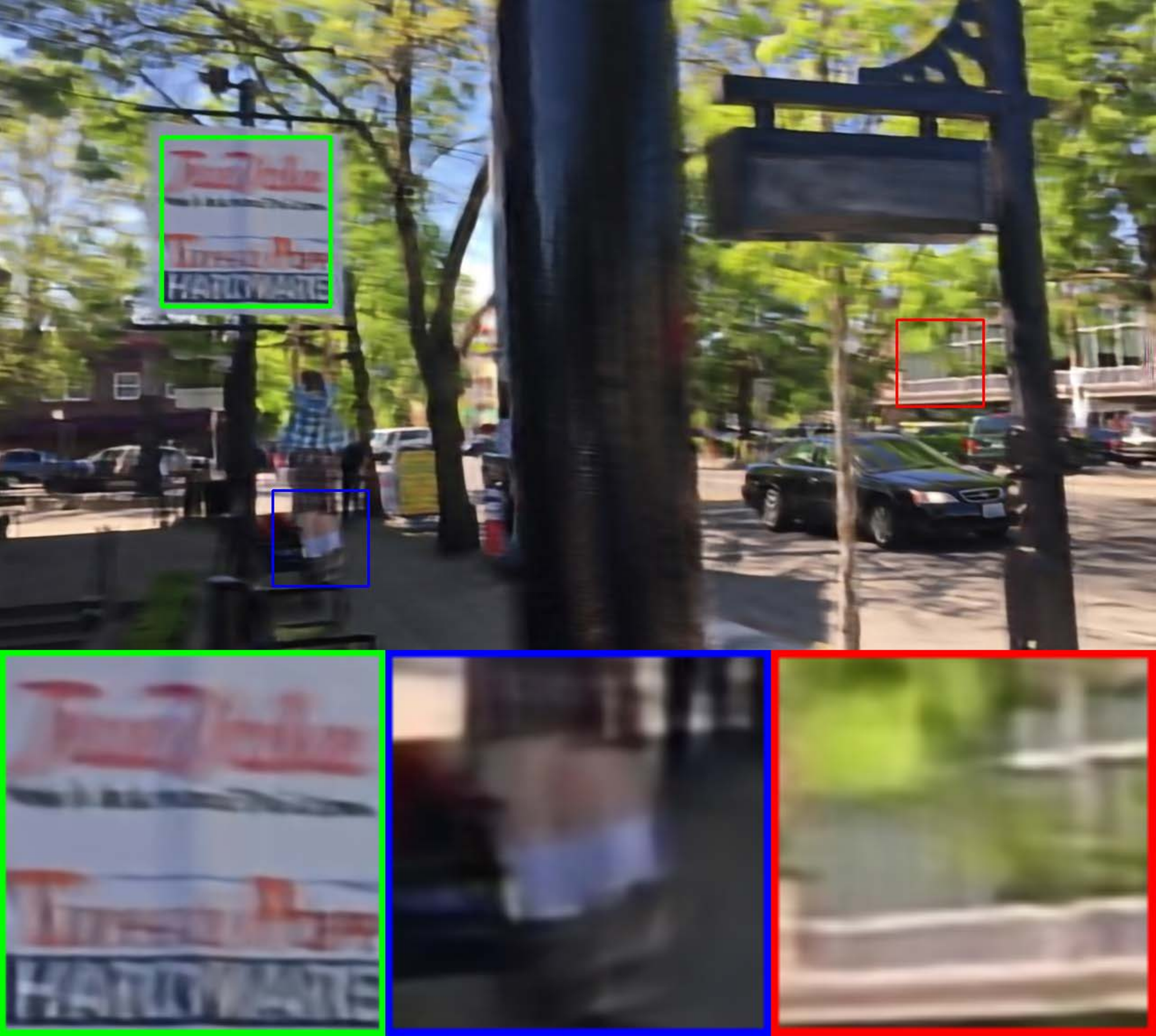} &
		\includegraphics[width=0.192\linewidth]{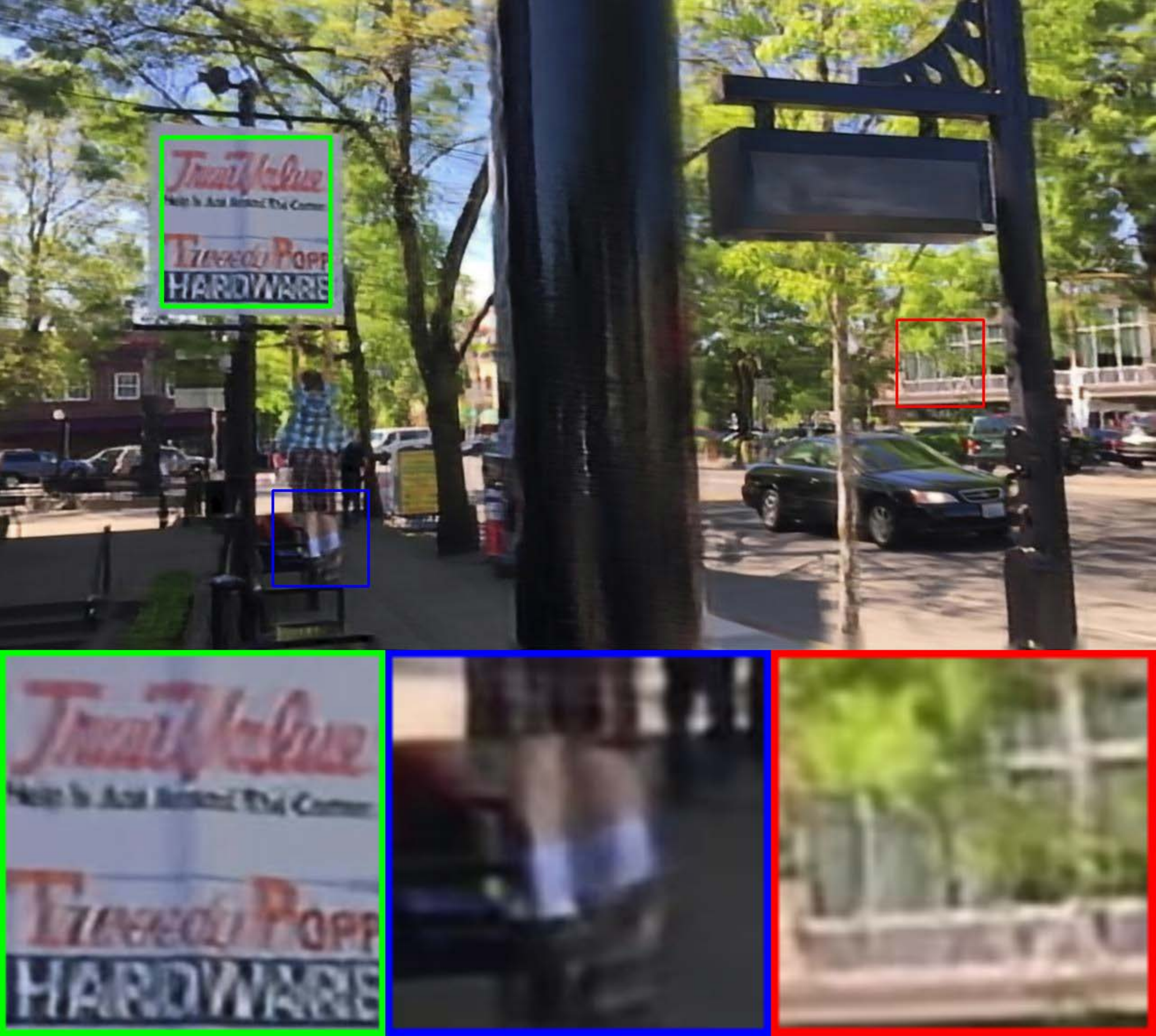} &
		\includegraphics[width=0.192\linewidth]{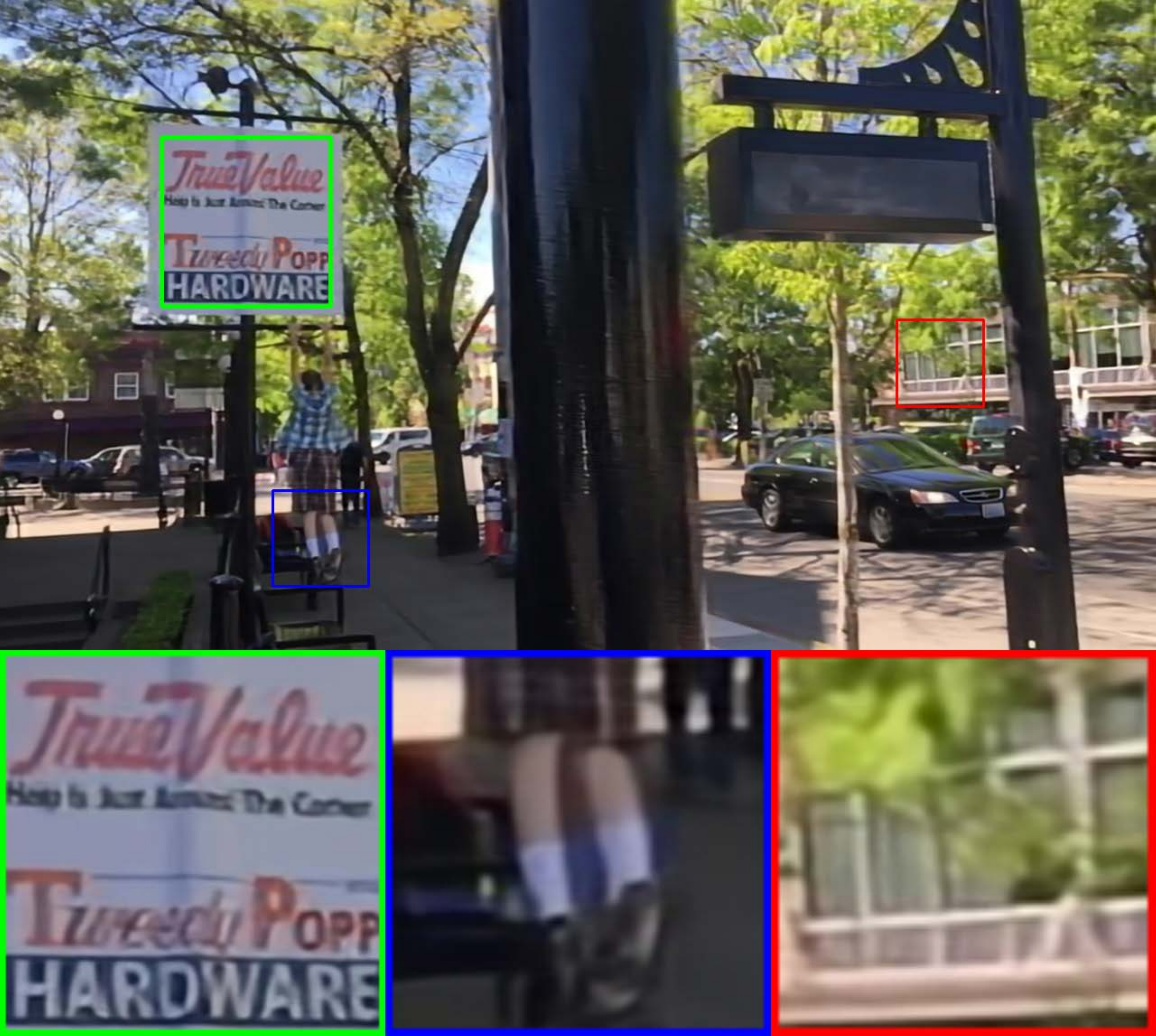} &	
		\includegraphics[width=0.192\linewidth]{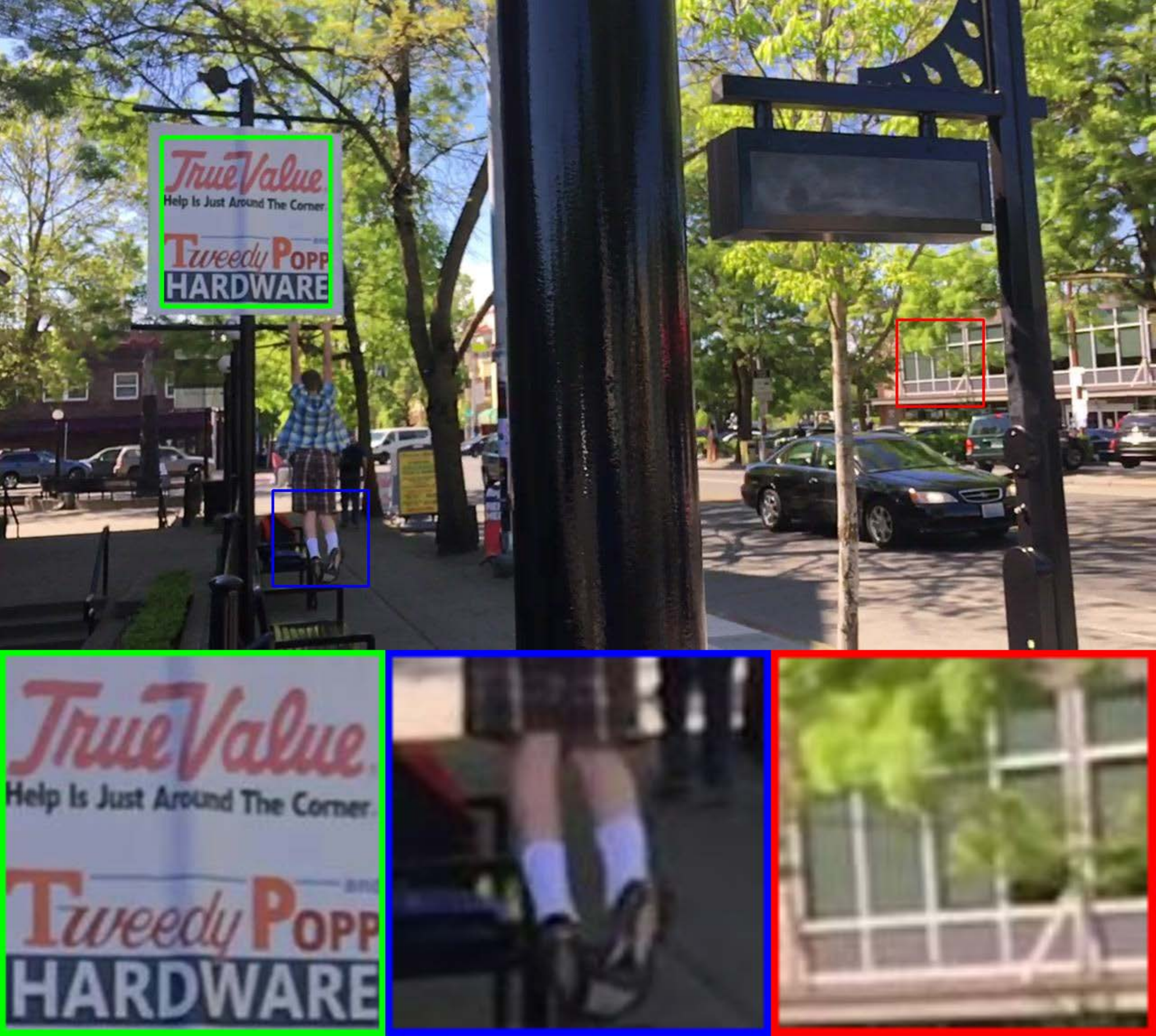} \\		
		(f) Kim and Lee~\cite{hyun2015generalized} & (g) Kim \textit{et al.}~\cite{hyun2017online} & (h) Su \textit{et al.}~\cite{su2017deep} & (i) Ours & (j) Ground truth \\
		20.97 / 0.7235  &  23.21 / 0.8023  & 23.98 / 0.8291  & \textbf{26.50 / 0.8820} & $+\infty$ / 1.0 \\
	\end{tabular}
\end{center}
\vspace{-3mm}
\caption{Qualitative evaluations on Video Deblurring Dataset~\cite{su2017deep}. The proposed method generates much sharper images with higher PSNR and SSIM.}
\label{fig:show_test}
\vspace{-2mm}
\end{figure*}
\begin{figure*}[t]\footnotesize
	\centering
	\renewcommand{\tabcolsep}{1pt}
	\renewcommand{\arraystretch}{1}
	\begin{center}
		\begin{tabular}{ccccc}
			\includegraphics[width=0.192\linewidth]{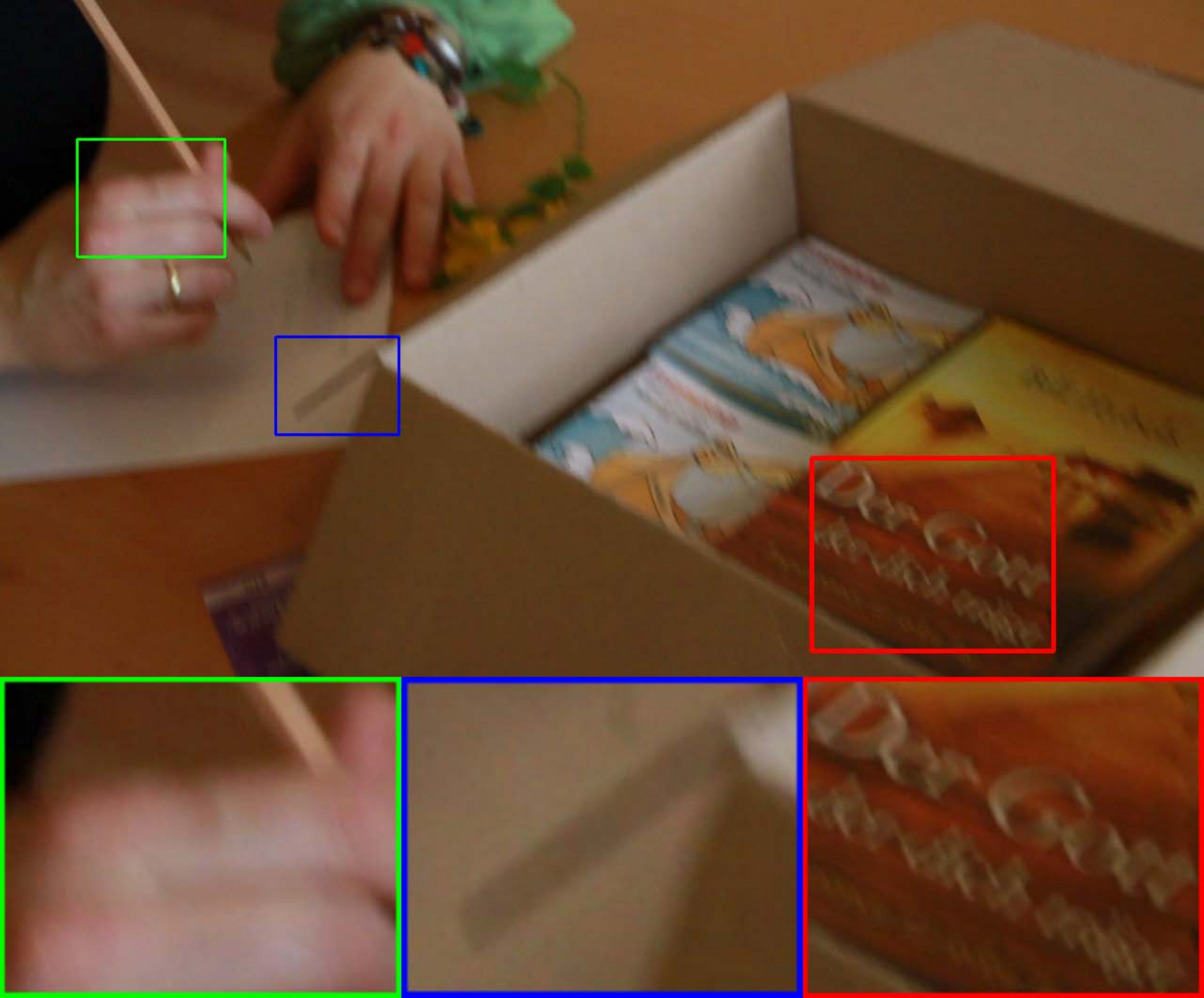} &
			\includegraphics[width=0.192\linewidth]{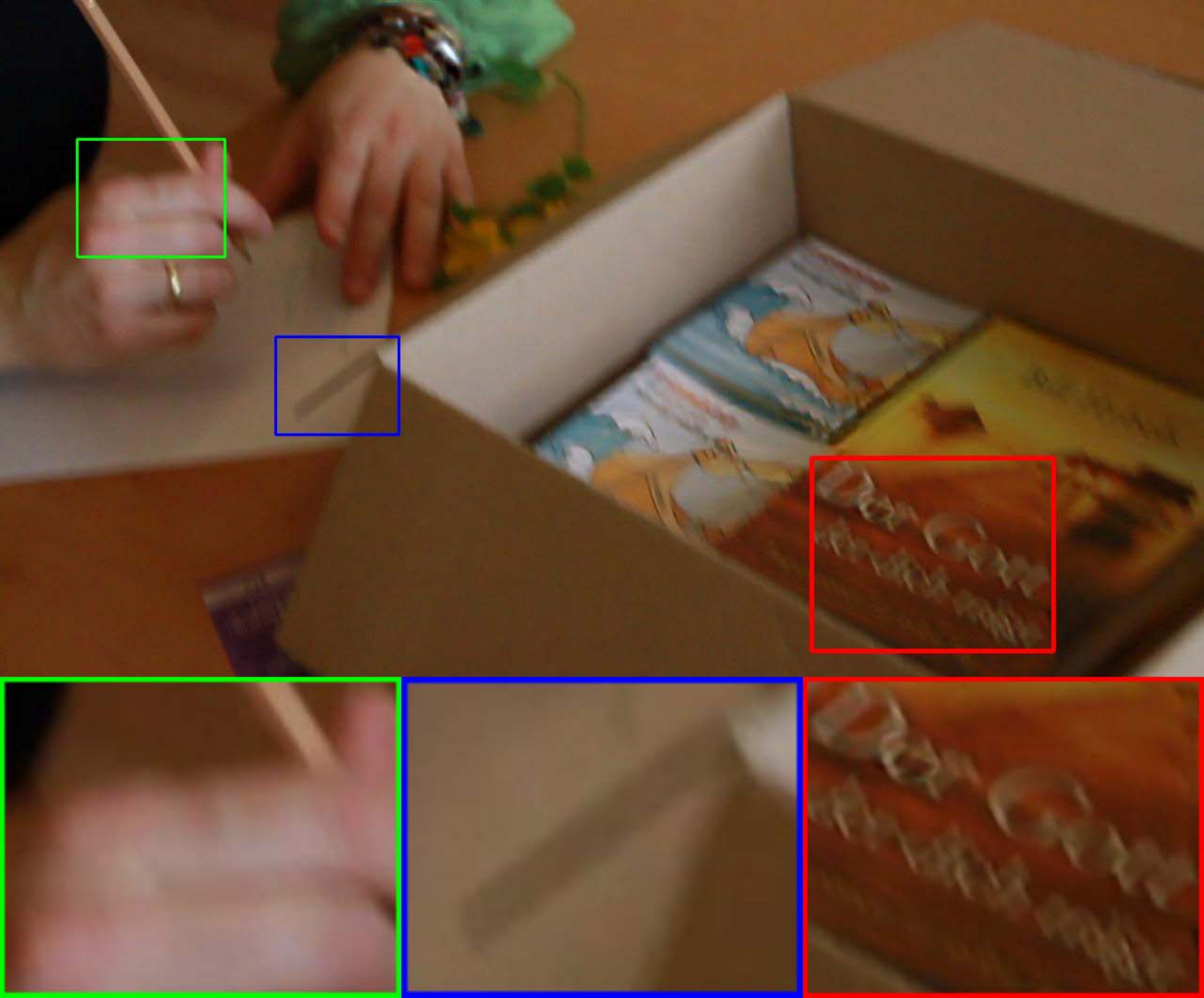} &
			\includegraphics[width=0.192\linewidth]{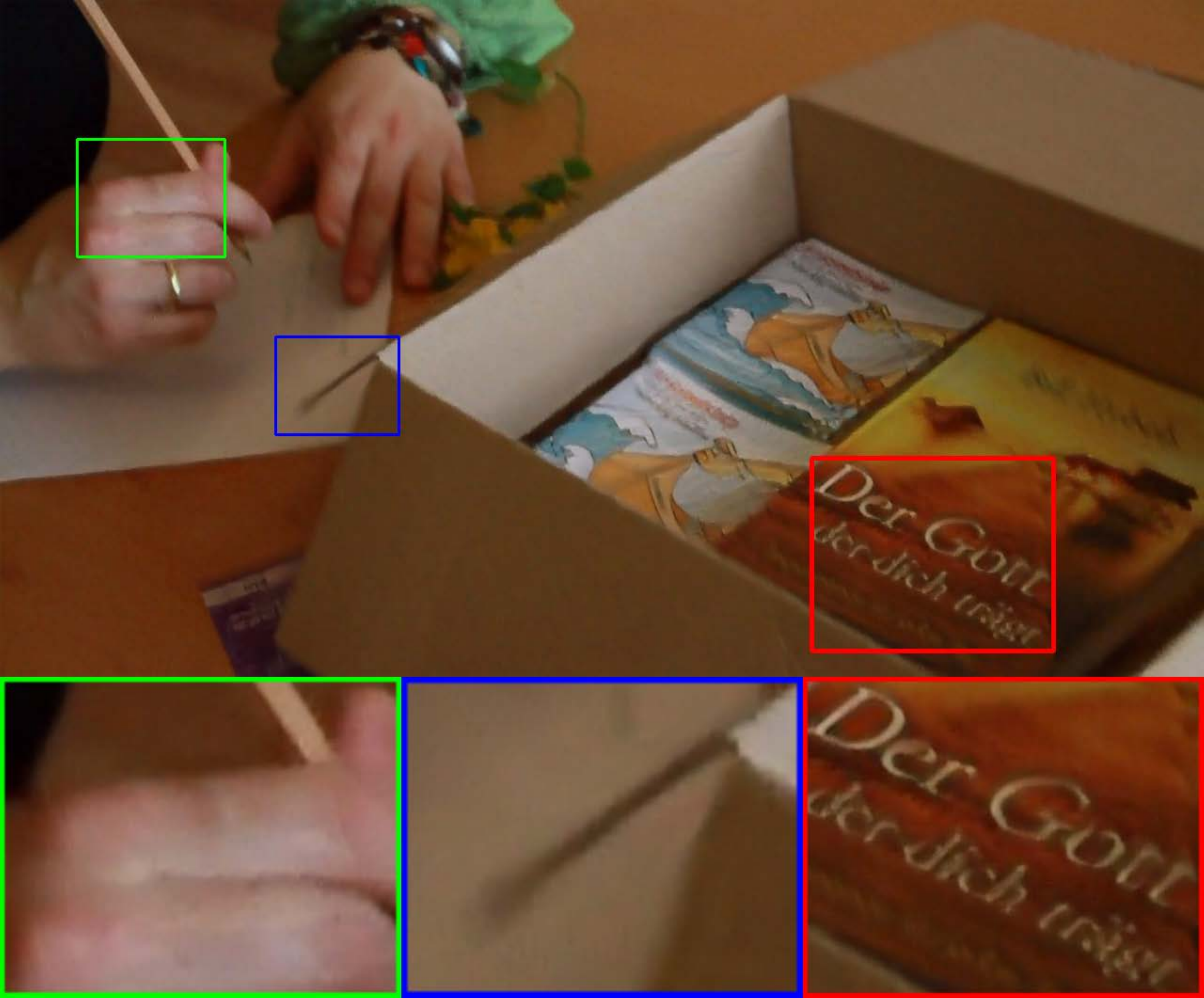} &
			\includegraphics[width=0.192\linewidth]{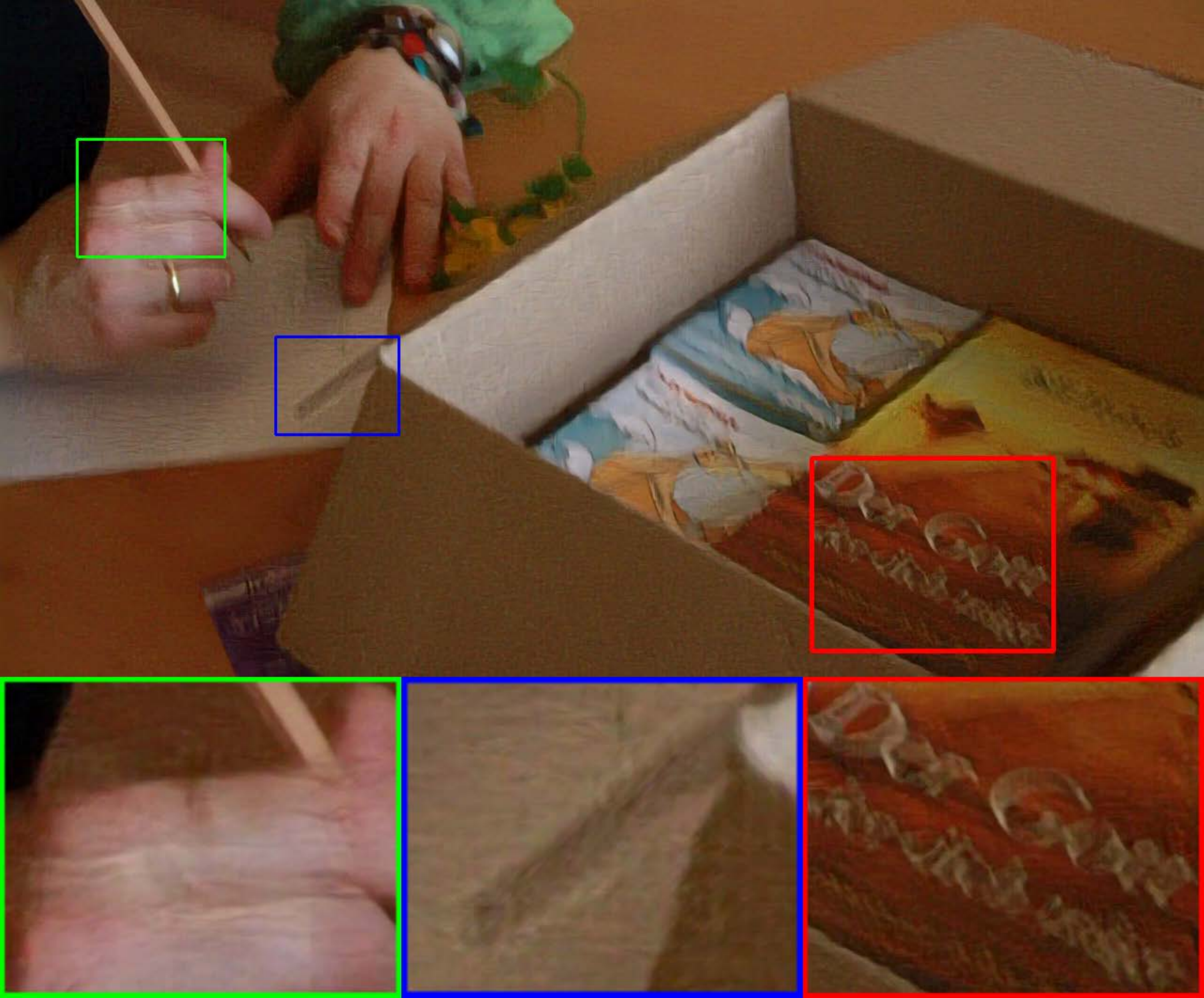} &
			\includegraphics[width=0.192\linewidth]{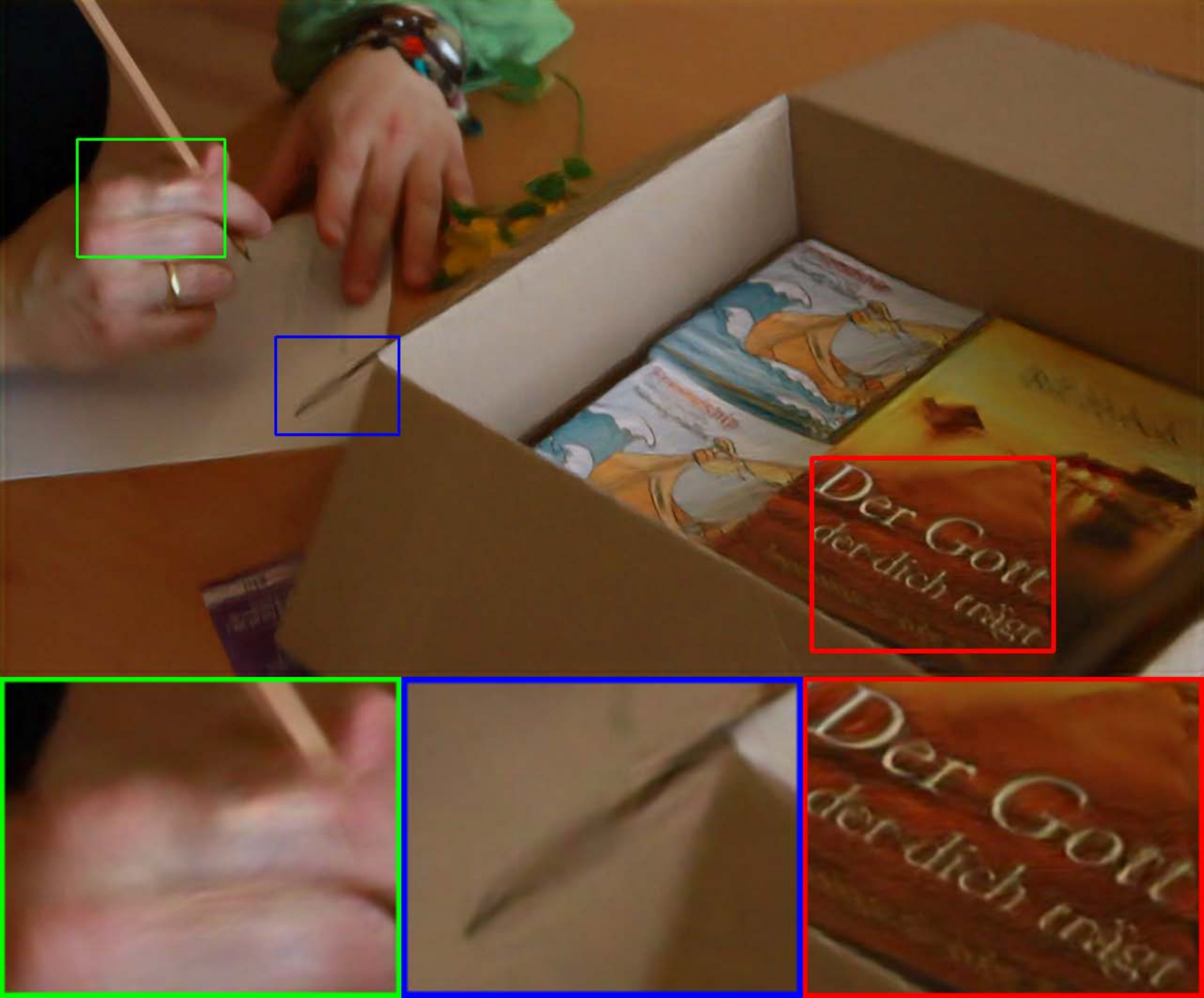} \\
			(a) Blurry image & (b) Gong \textit{et al.}~\cite{gong2017motion} & (c) Nah \textit{et al.}~\cite{nah2017deep}  & (d) Kupyn \textit{et al.}~\cite{kupyn2018deblurgan}& (e) Zhang \textit{et al.}~\cite{zhang2018dynamic}
			\vspace{1.5pt}\\
			\includegraphics[width=0.192\linewidth]{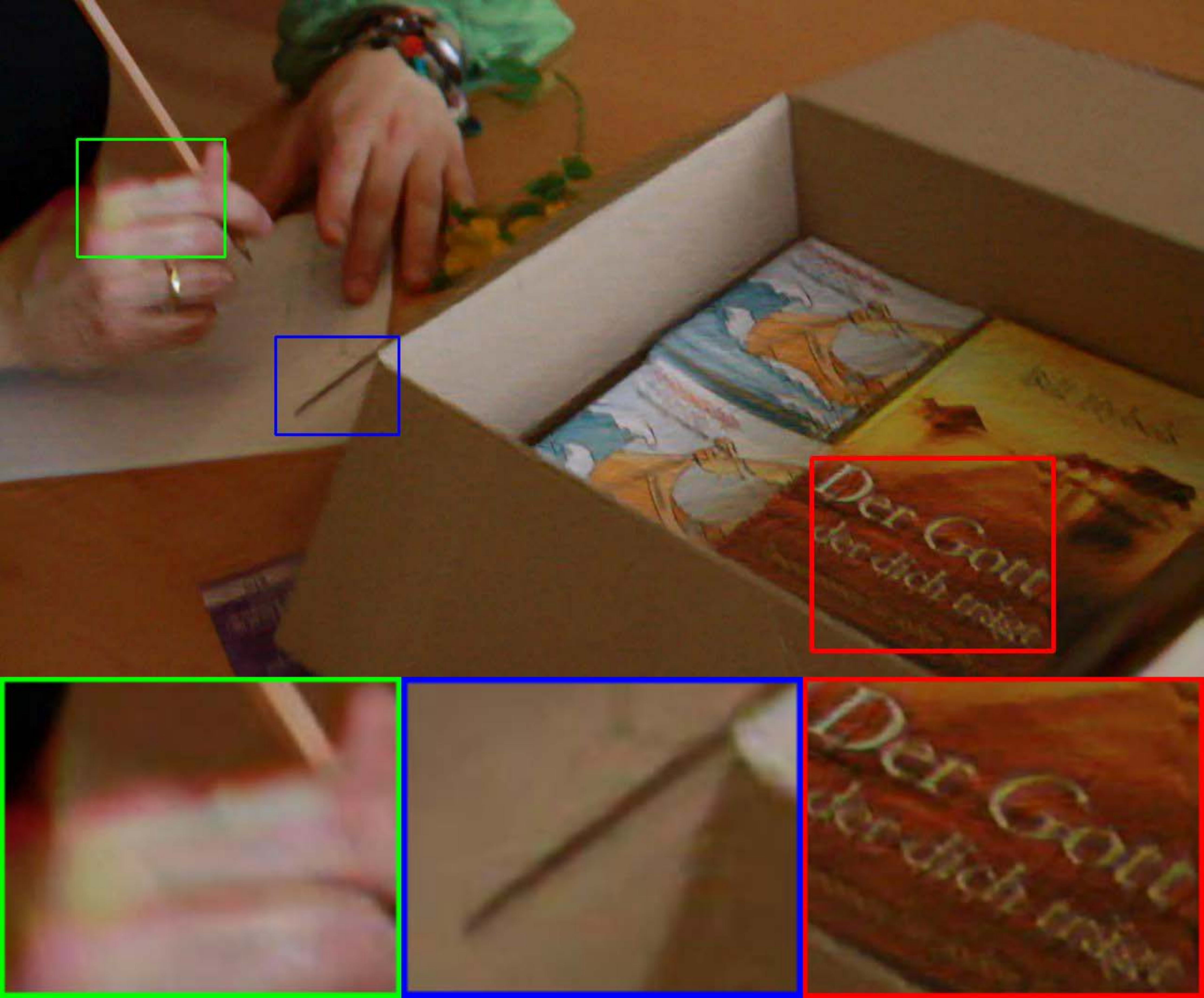}&
			\includegraphics[width=0.192\linewidth]{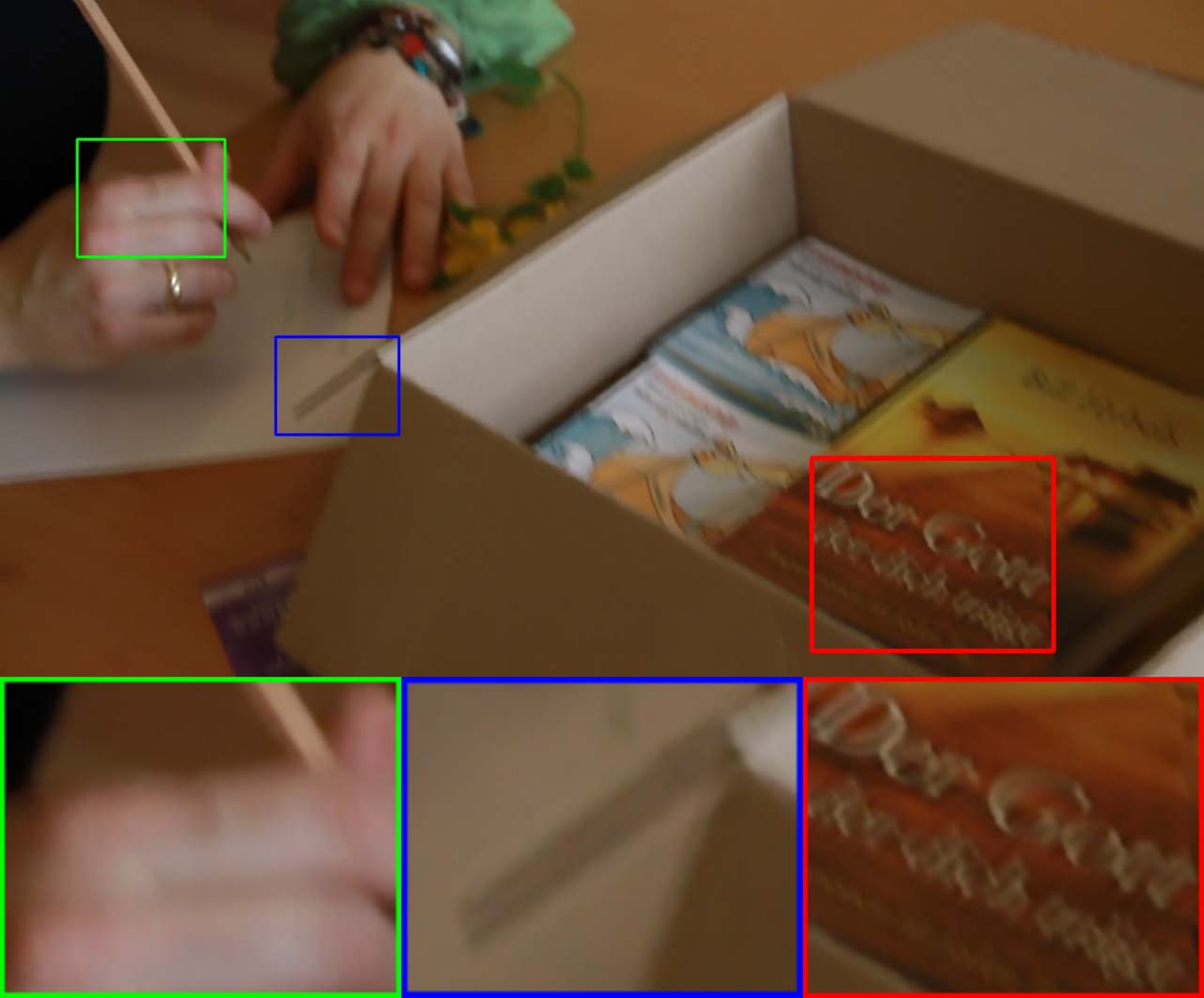} &
			\includegraphics[width=0.192\linewidth]{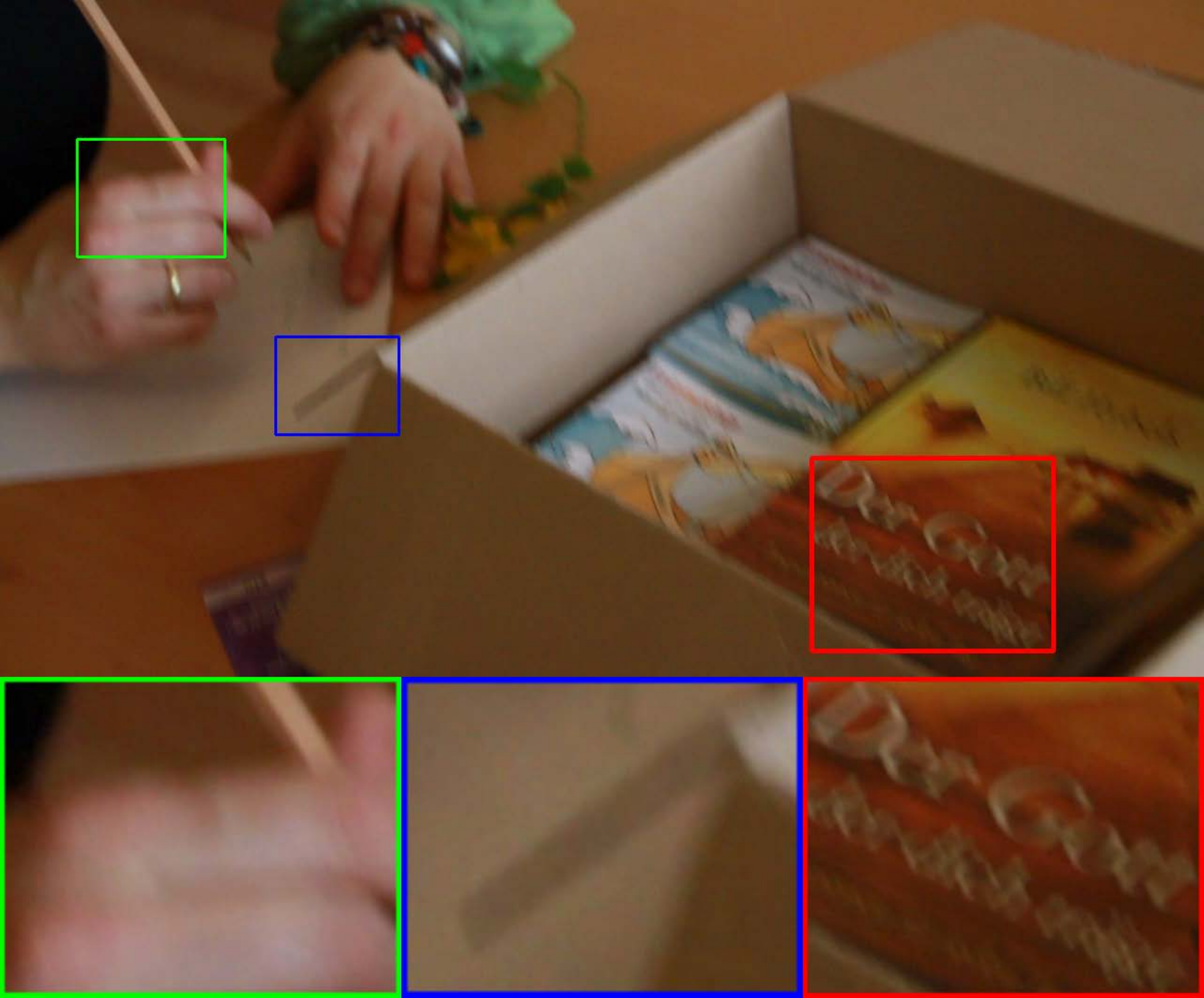} &
			\includegraphics[width=0.192\linewidth]{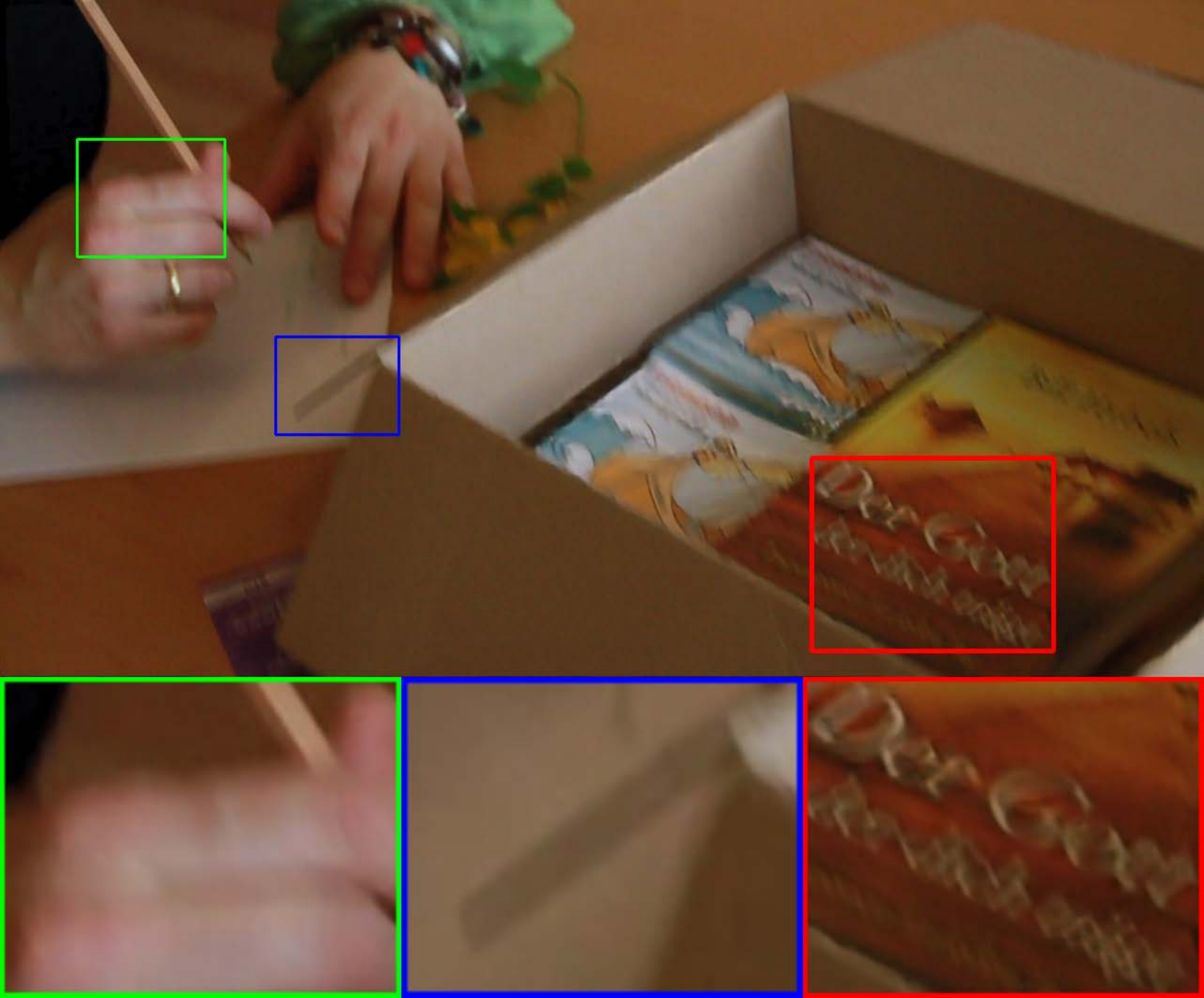} &
			\includegraphics[width=0.192\linewidth]{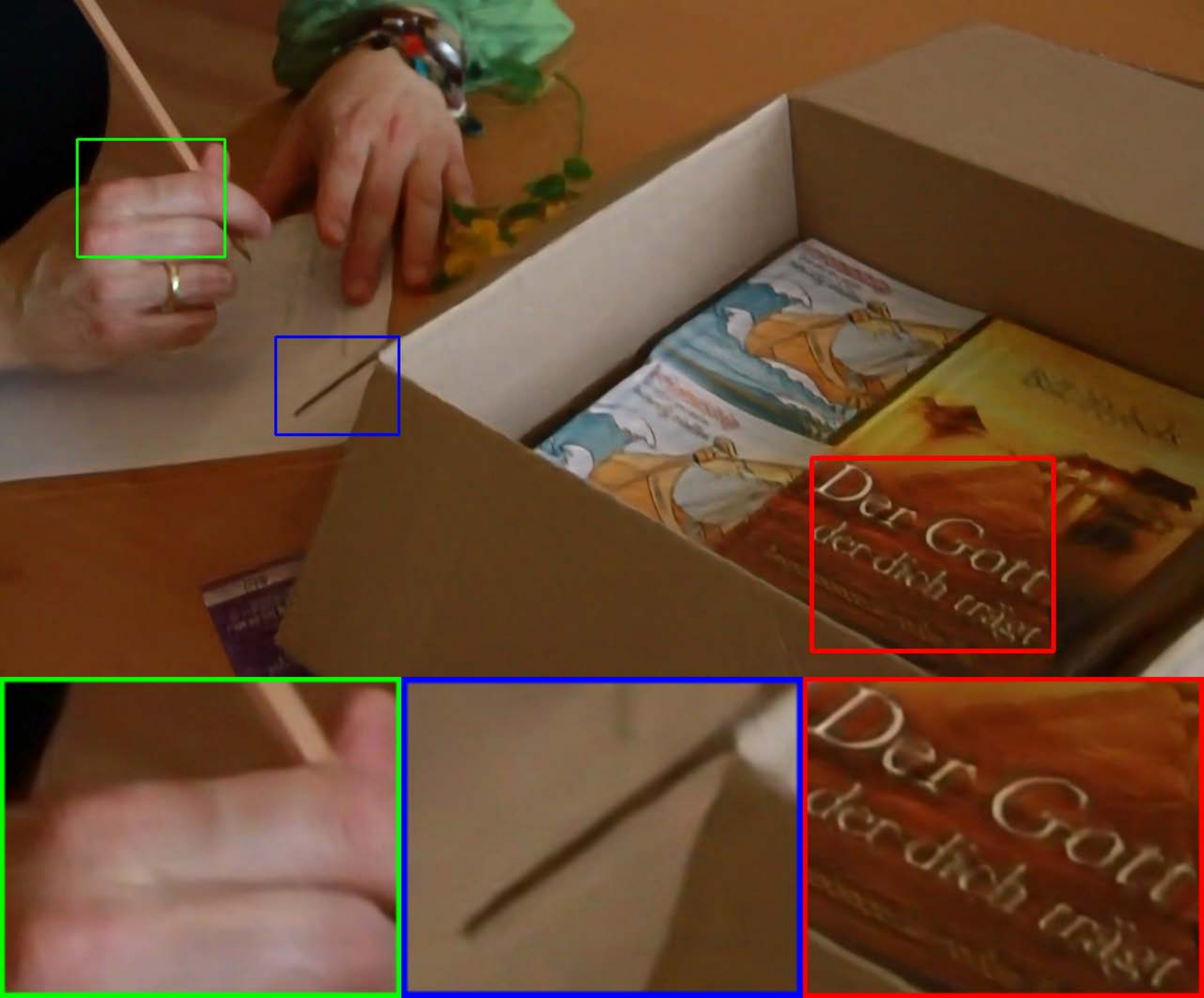} \\
			(f) Tao \textit{et al.}~\cite{tao2018scale} & (g) Kim and Lee~\cite{hyun2015generalized} & (h) Kim \textit{et al.}~\cite{hyun2017online}  & (i) Su \textit{et al.}~\cite{su2017deep} & (j) Ours \\
		\end{tabular}
	\end{center}
	\vspace{-3mm}
	\caption{Qualitative evaluations on the real blurry videos~\cite{su2017deep}. The proposed method generates much clearer images.}
	\label{fig:show_real}
	\vspace{-3mm}
\end{figure*}

\vspace{-4mm}
\noindent\textbf{Experimental Settings.}
\label{sec:training}
We initialize our neural network using the initialization method in~\cite{he2015delving},
and train it using Adam~\cite{kingma2015adam} optimizer with $\beta_1 = 0.9$ and $\beta_2 = 0.999$.
We set the initial learning rate to $10^{-4}$ and decayed by 0.1 every 400k iterations.
The proposed network converges after 900k iterations. 
We quantitatively and qualitatively evaluate the proposed method on the video deblurring dataset \cite{su2017deep}.
For a fair comparison with the most related deep learning-based algorithms~\cite{nah2017deep, kupyn2018deblurgan, zhang2018dynamic, tao2018scale},
we finetune all these methods by the corresponding publicly released implementations on the video deblurring dataset~\cite{su2017deep}.
In our experiments, we use both PSNR and SSIM as quantitative evaluation metrics for synthetic testing set.
The training code, test model, and experimental results will be available to the public.
\subsection{Experimental Results}
\noindent\textbf{Quantitative Evaluations.} We compare the proposed algorithm with the state-of-the-art video deblurring methods including conventional optical flow-based pixel-wise kernel estimation~\cite{hyun2015generalized} and CNN based methods~\cite{su2017deep,hyun2017online}.
We also compare it with the state-of-the-art image deblurring methods including conventional non-uniform deblurring~\cite{whyte2012non}, CNN based spatially variant blur kernel estimation~\cite{sun2015learning, gong2017motion}, and end-to-end CNN methods~\cite{nah2017deep, kupyn2018deblurgan, zhang2018dynamic, tao2018scale}.

Table~\ref{tab:psnr_time_size} shows that the proposed method performs favorably against the state-of-the-art algorithms on the testing set of dynamic scene video deblurring dataset~\cite{su2017deep}.

Figure~\ref{fig:show_test} shows some examples in the testing set from~\cite{su2017deep}.
It shows that the existing methods cannot keep sharp details and remove the non-uniform blur well.
With temporal alignment and spatially variant deblurring, our network performs the best and restores much clearer images with more details.

\noindent\textbf{Qualitative Evaluations.}
To further validate the generalization ability of the proposed method,
we also qualitatively compare the proposed network with other algorithms on real blurry images from \cite{su2017deep}.
As illustrated in Figure~\ref{fig:show_real}, the proposed method can restore shaper images with more image details than the state-of-the-art image and video deblurring methods.
The comparison results show that our STFAN can robustly handle unknown real blur in dynamic scenes, which further demonstrates the superiority of the proposed framework.
\subsection{Running Time and Model Size}
We implement the proposed network using PyTorch platform~\cite{pytorch}.
To speed up, we implement the proposed FAC layer with CUDA.
We evaluate the proposed method and state-of-the-art image or video deblurring methods on the same server with an Intel Xeon E5 CPU and an NVIDIA Titan Xp GPU.
The traditional algorithms~\cite{whyte2012non, hyun2015generalized} are time-consuming due to a complex optimization process.
Therefore, \cite{sun2015learning} and \cite{gong2017motion} utilize the CNN to estimate non-uniform blur kernels based on motion flow.
However, they are still time-consuming since the traditional non-blind deblurring algorithm~\cite{zoran2011learning} is used to restore the sharp images.
DVD~\cite{su2017deep} uses CNN to restore sharp images from neighboring multiple blurry frames, but they use a traditional optical flow method~\cite{perez2013tv} to align these input frames and is computationally expensive.
With GPU implementation, the end-to-end CNN-based methods~\cite{nah2017deep, kupyn2018deblurgan, zhang2018dynamic, tao2018scale, hyun2017online} are relatively efficient.
To enlarge the receptive field, the networks in \cite{nah2017deep, kupyn2018deblurgan, zhang2018dynamic, tao2018scale} are very deep, which lead to a large model size as well as a long processing time.
Even though spatially variant RNNs are used in \cite{zhang2018dynamic} to enlarge the receptive field, they need a deep network to estimate the RNN weights and RNNs are also time-consuming.
Our network uses the aligned deblurred features of the previous frame, which reduces the difficulty for the network to restore the sharp image of the current frame.
In addition, the FAC layer is effective for spatially variant alignment and deblurring.
Benefited from the above two merits, our networks are designed to be small and efficient.
As shown in Table~\ref{tab:psnr_time_size}, the proposed network has less running time and smaller model size than the existing end-to-end CNN methods.
Even though \cite{hyun2017online} runs slightly faster and has smaller model size, the proposed method performs better with the frame alignment and deblurring in the feature domain.
\subsection{Temporal consistency}
\begin{figure}[h]
	\centering
	\vspace{1mm}
	\renewcommand{\tabcolsep}{0.5pt}
	\renewcommand{\arraystretch}{0.8}
	\scriptsize
	\def \k {0.13} 
	\begin{center}
		\centering 
		\begin{tabular}{lcccccccc}
			\begin{sideways}{\tiny\ \ \ \ Input}\end{sideways}&
			\includegraphics[width=\k\linewidth]{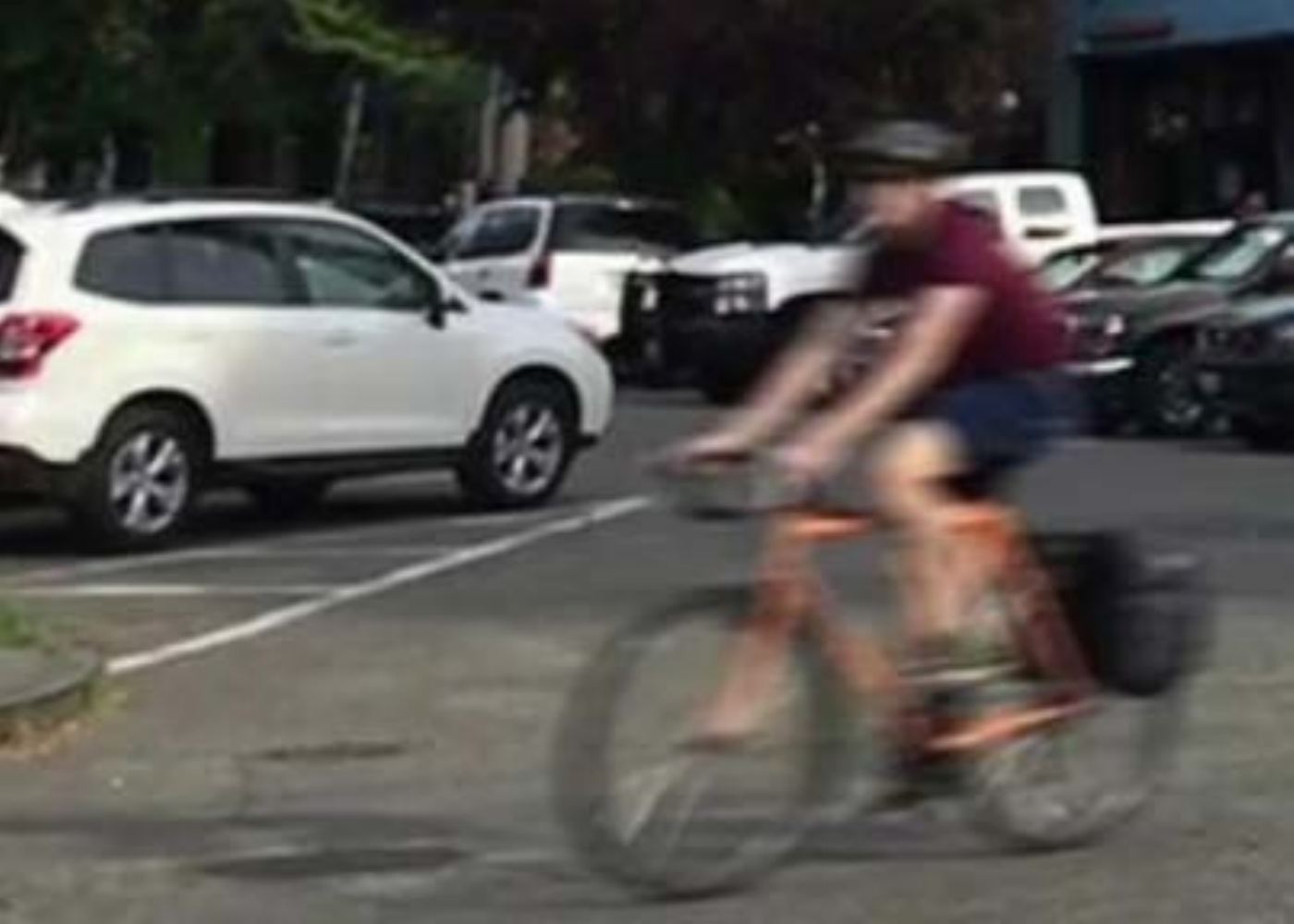} &
			\includegraphics[width=\k\linewidth]{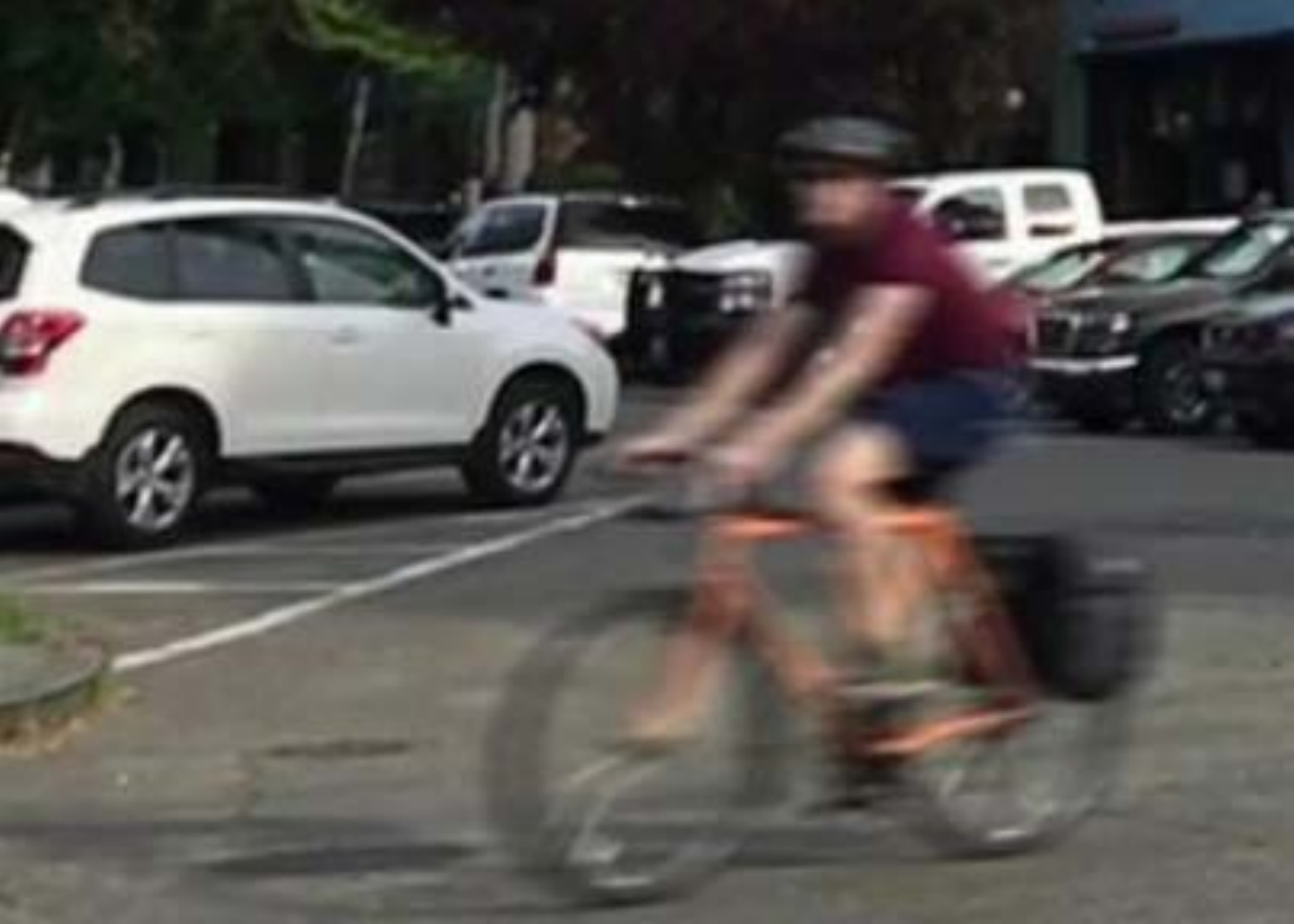} &
			\includegraphics[width=\k\linewidth]{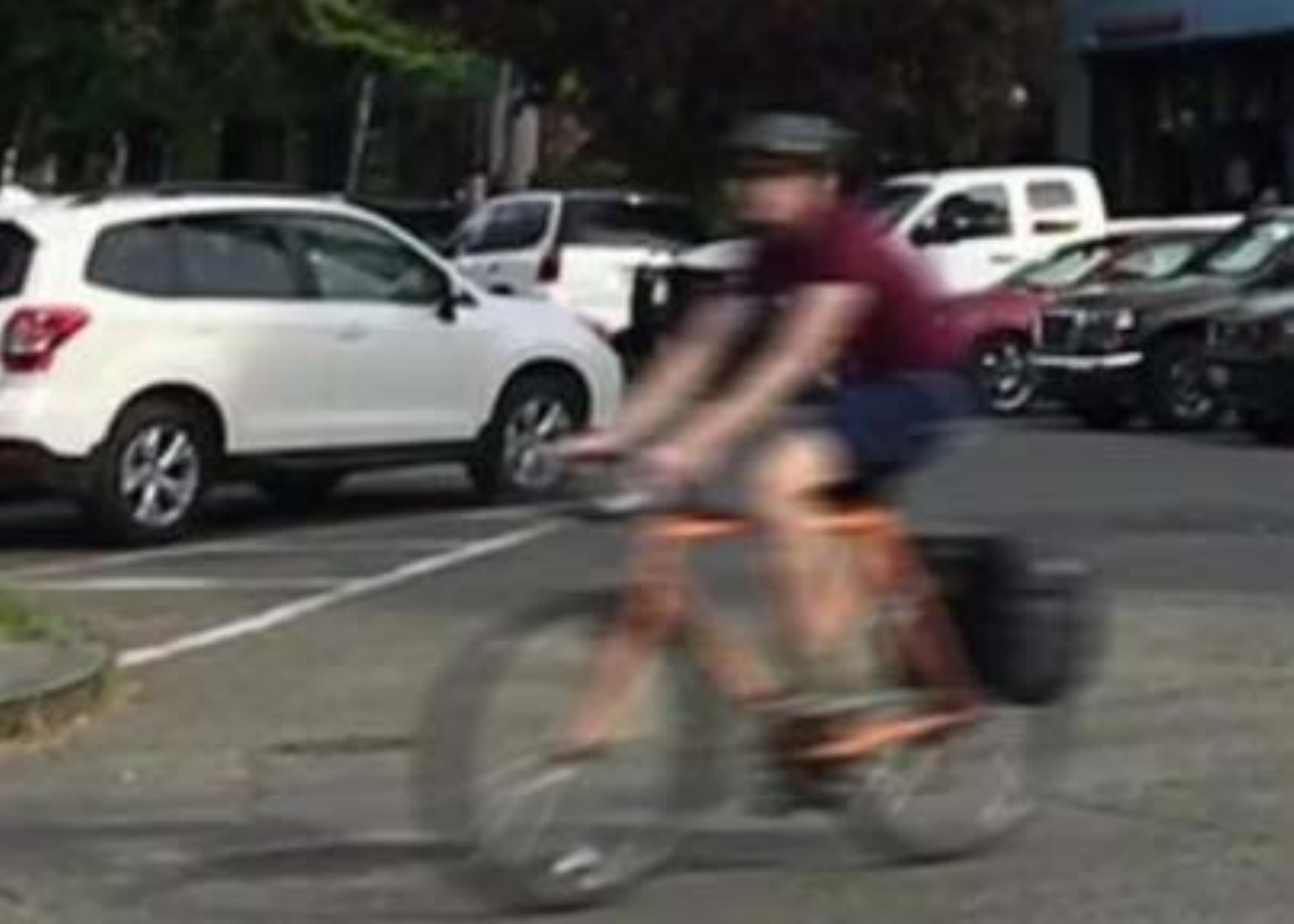} &
			\includegraphics[width=\k\linewidth]{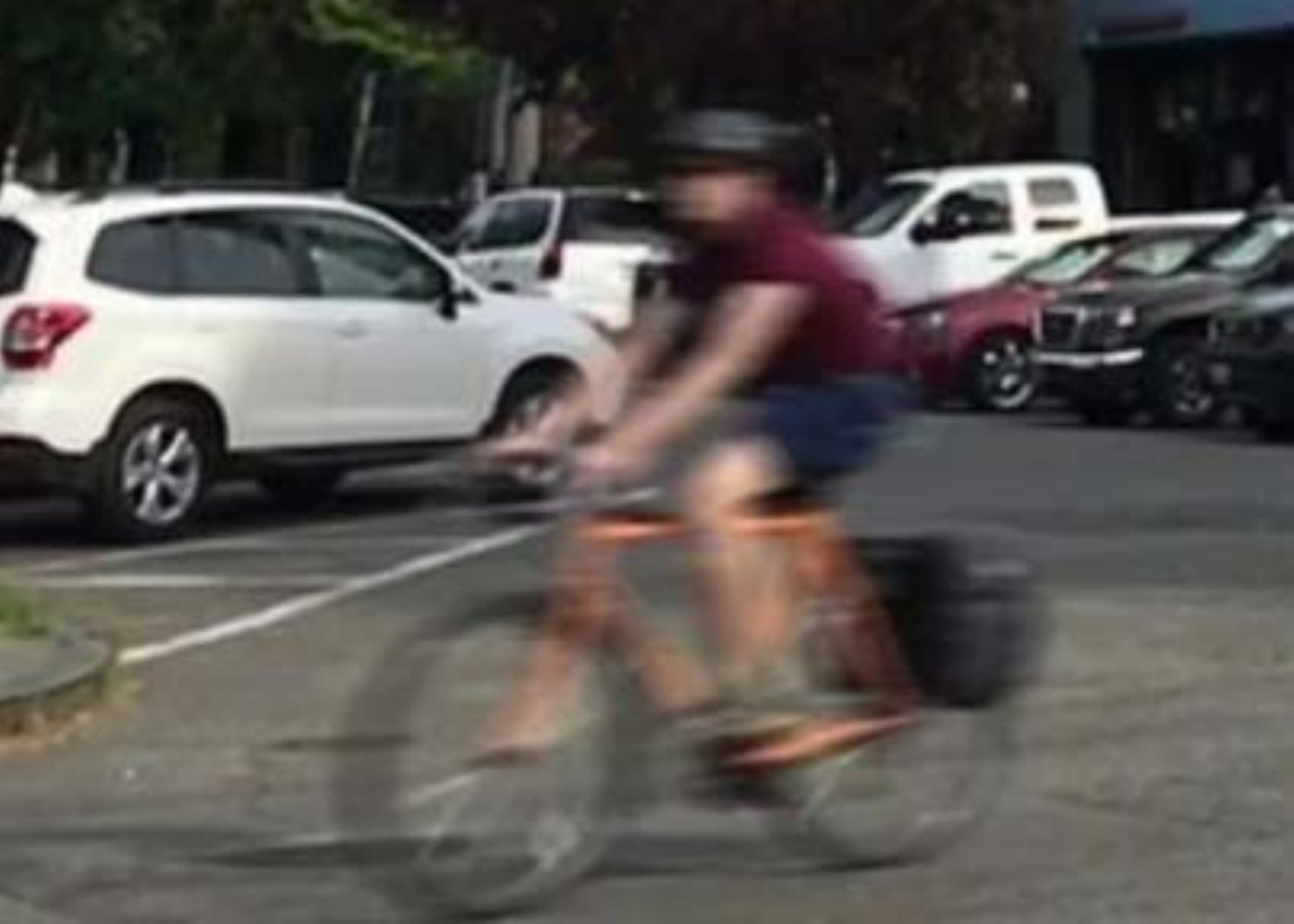} &
			\includegraphics[width=\k\linewidth]{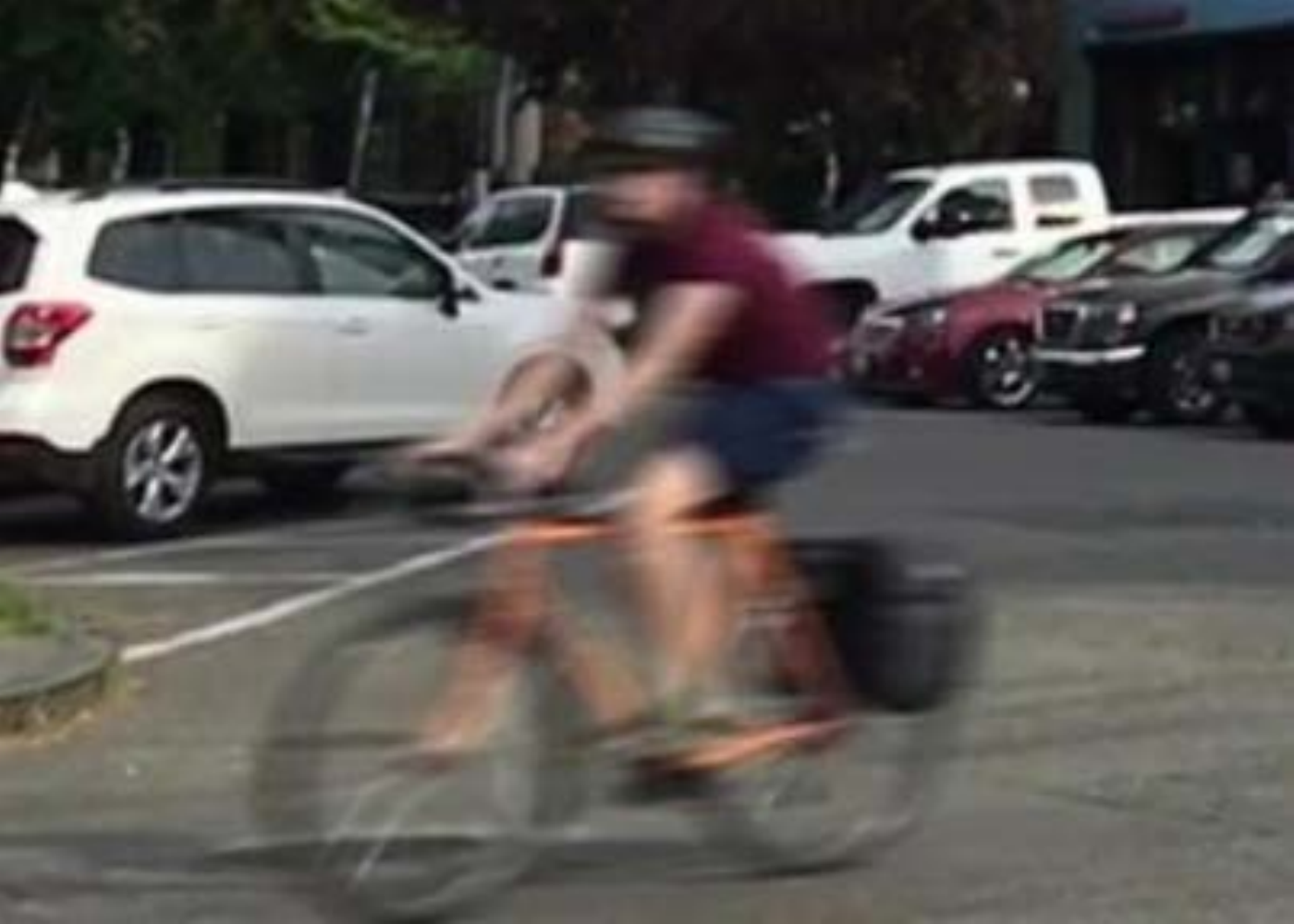} &
			\includegraphics[width=\k\linewidth]{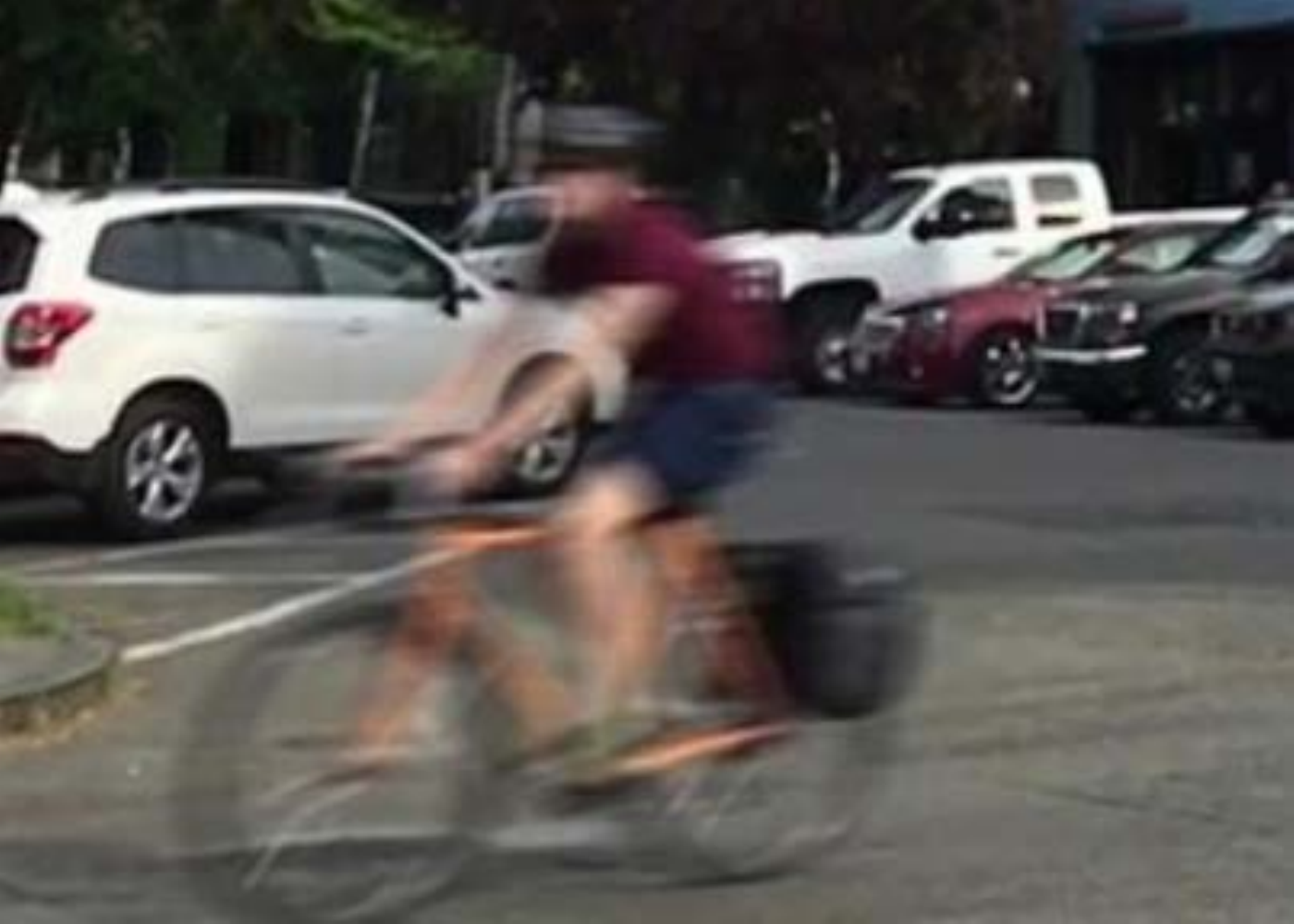} &
			\includegraphics[width=\k\linewidth]{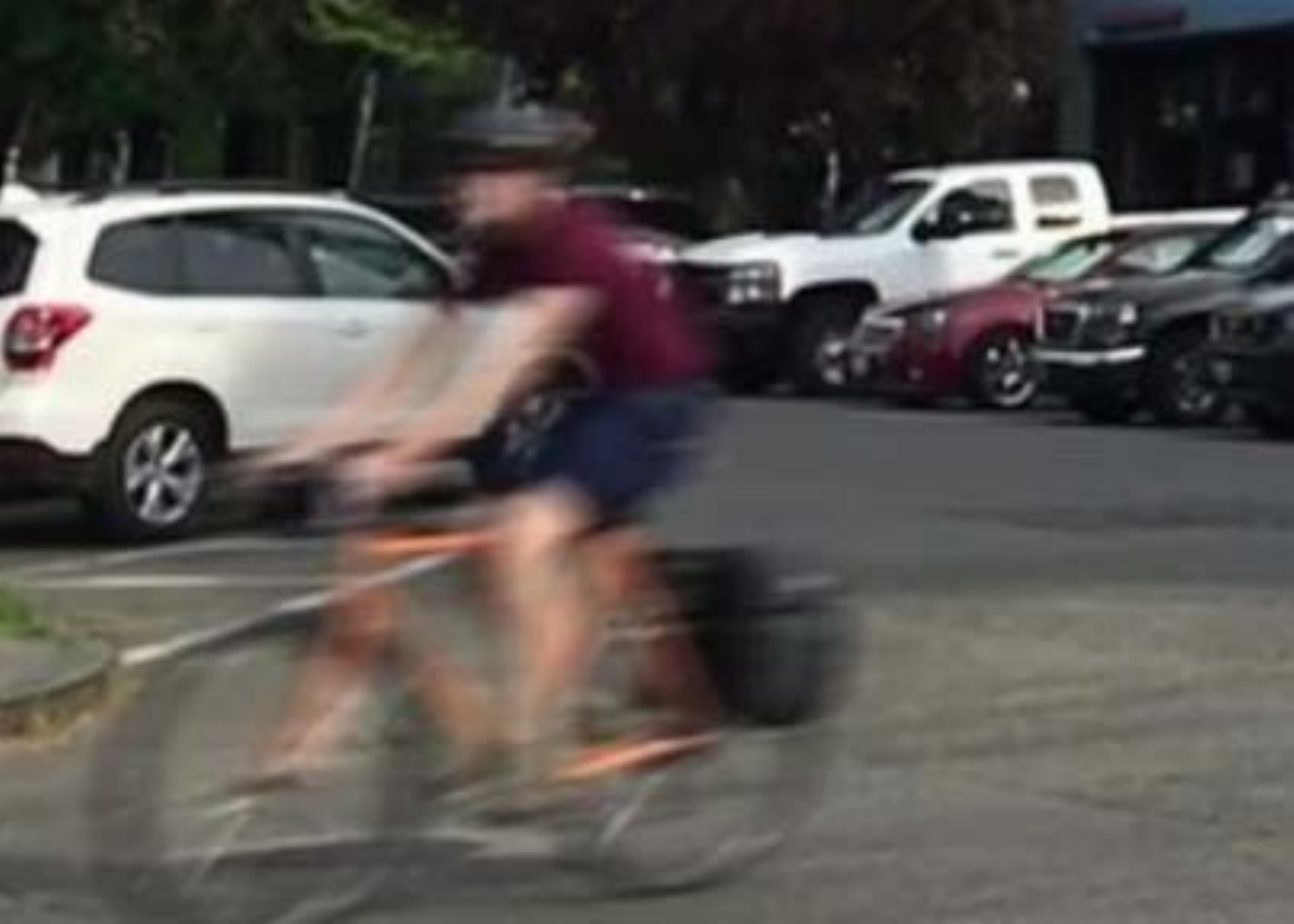} \\
			\begin{sideways}{\tiny \ OVD~\cite{hyun2017online}}\end{sideways}&
			\includegraphics[width=\k\linewidth]{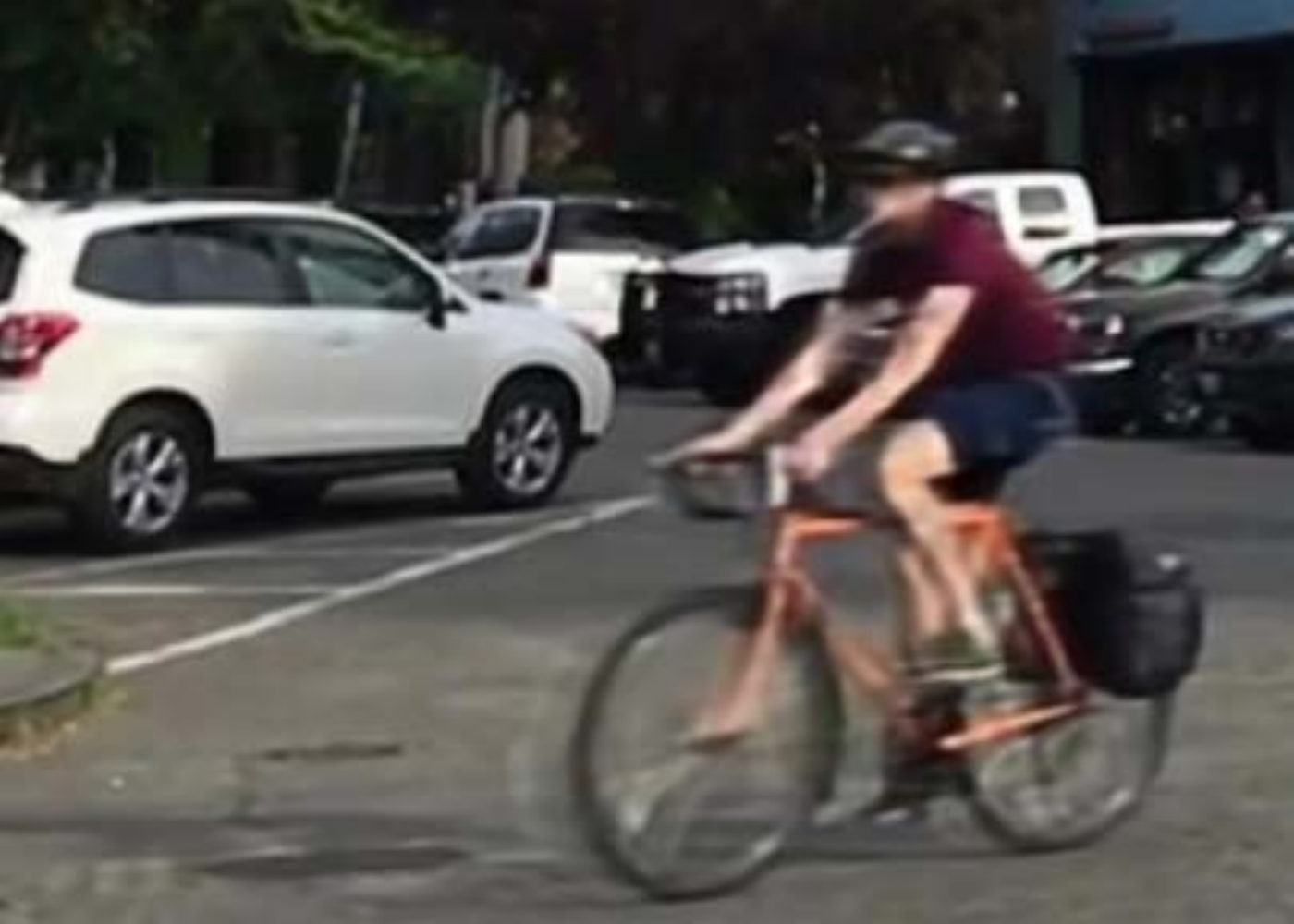} &
			\includegraphics[width=\k\linewidth]{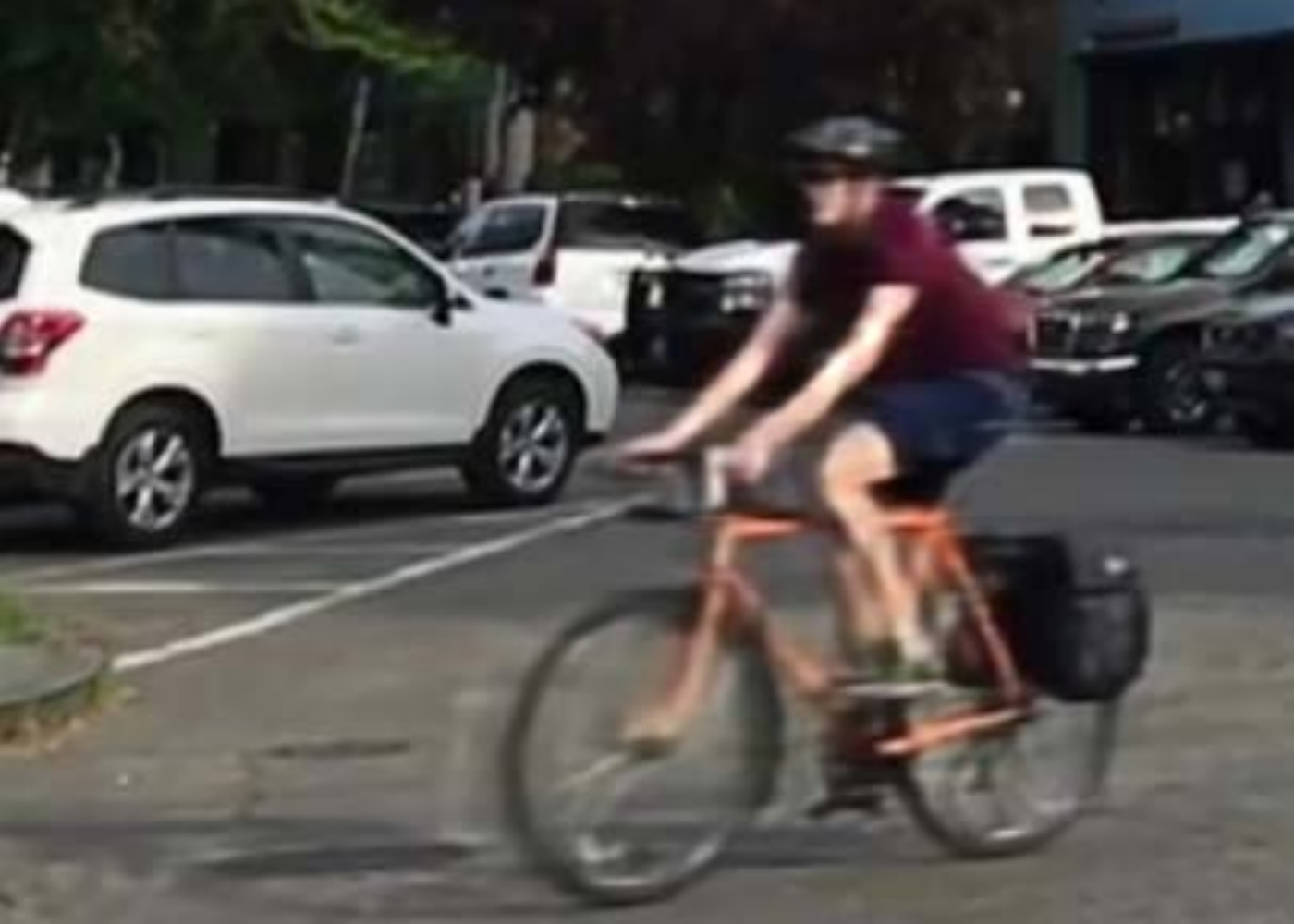} &
			\includegraphics[width=\k\linewidth]{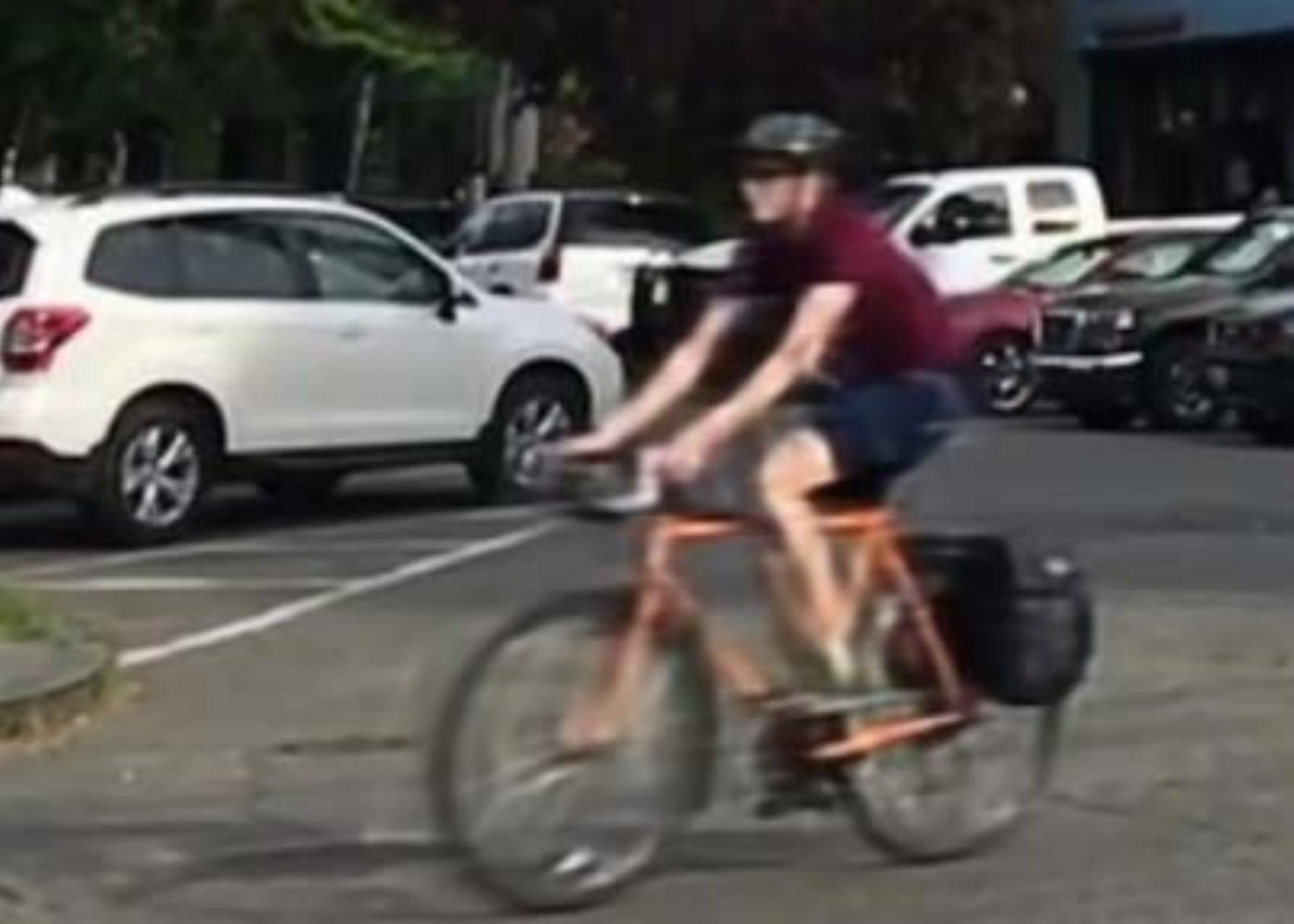} &
			\includegraphics[width=\k\linewidth]{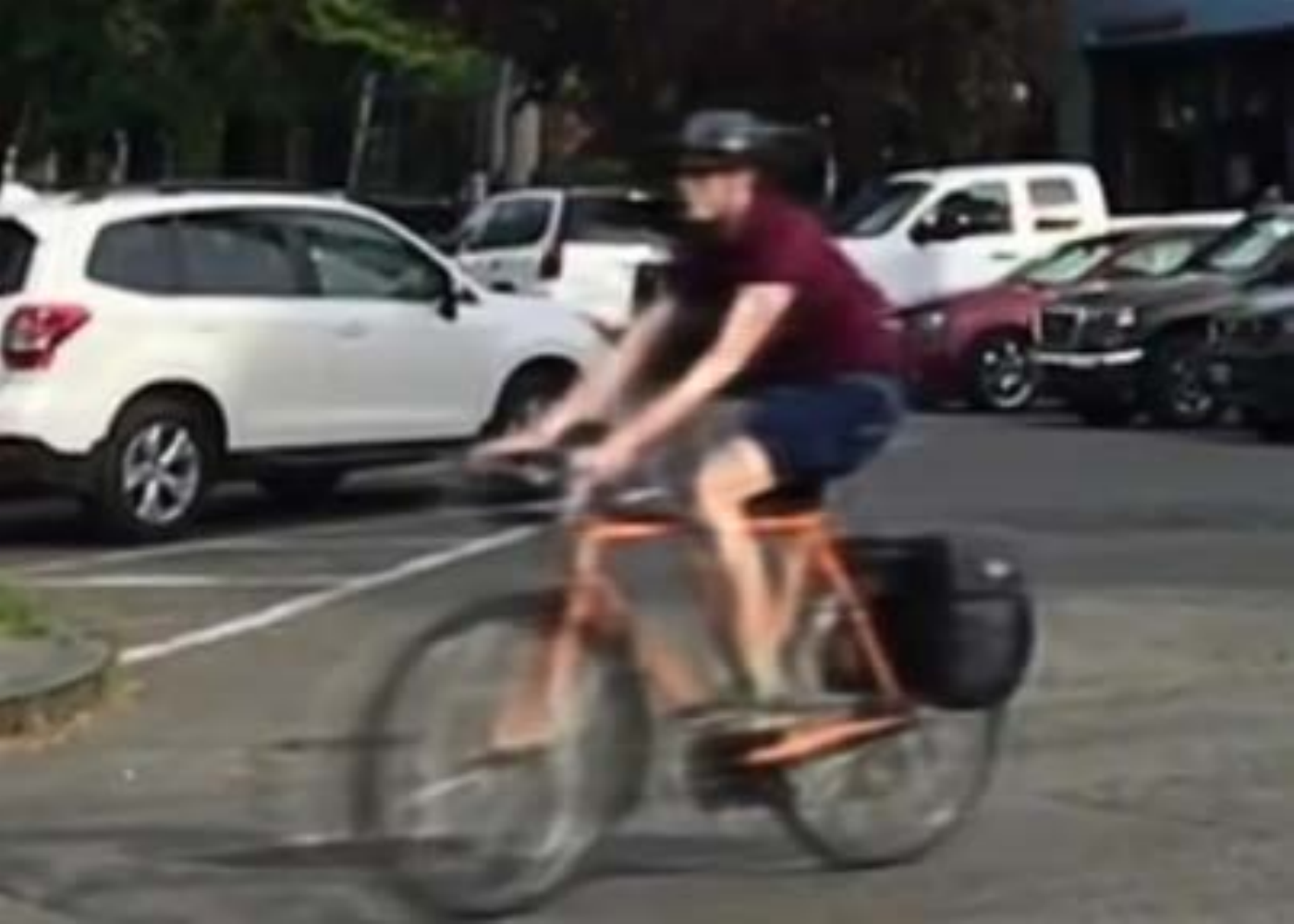} &
			\includegraphics[width=\k\linewidth]{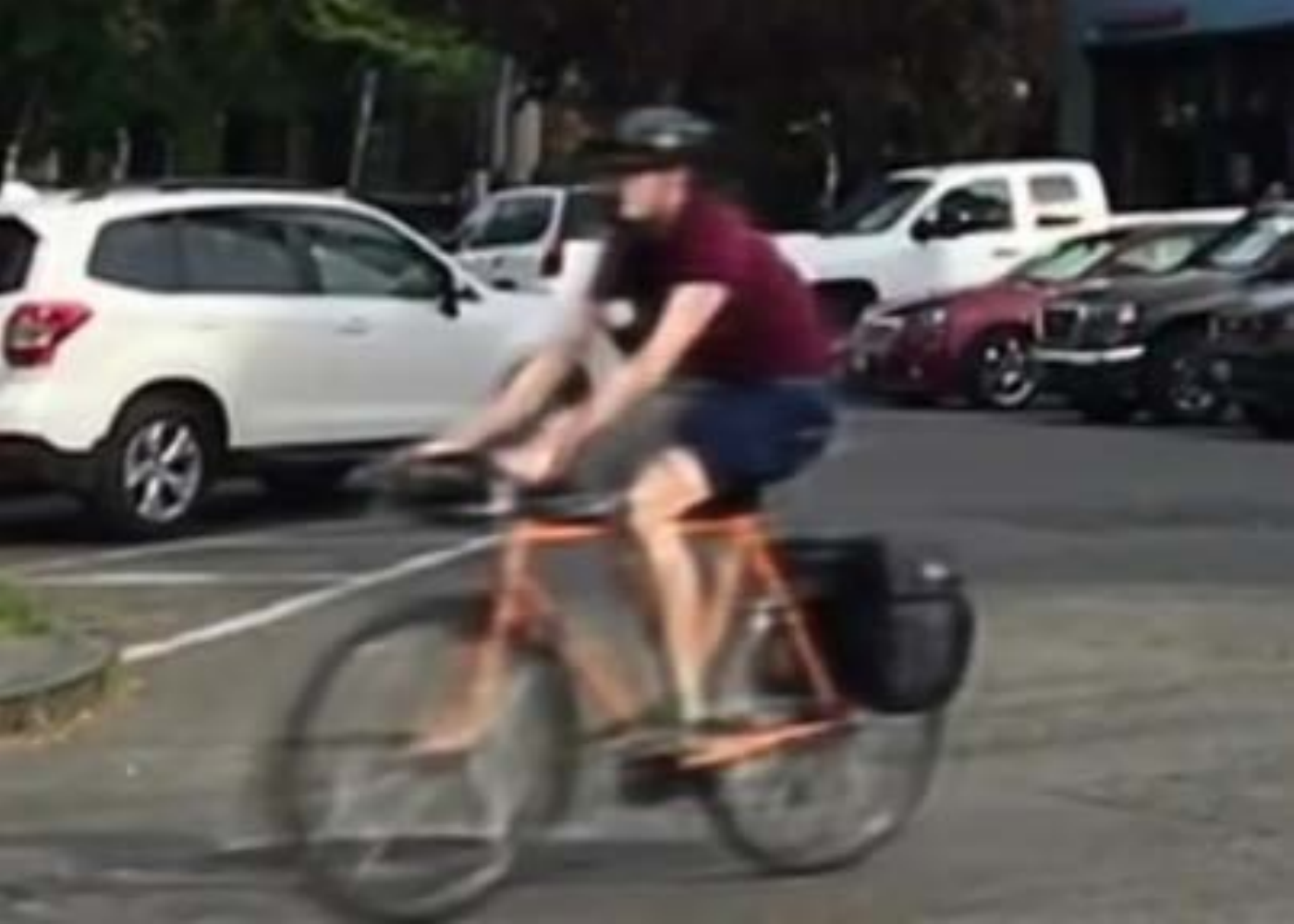} &
			\includegraphics[width=\k\linewidth]{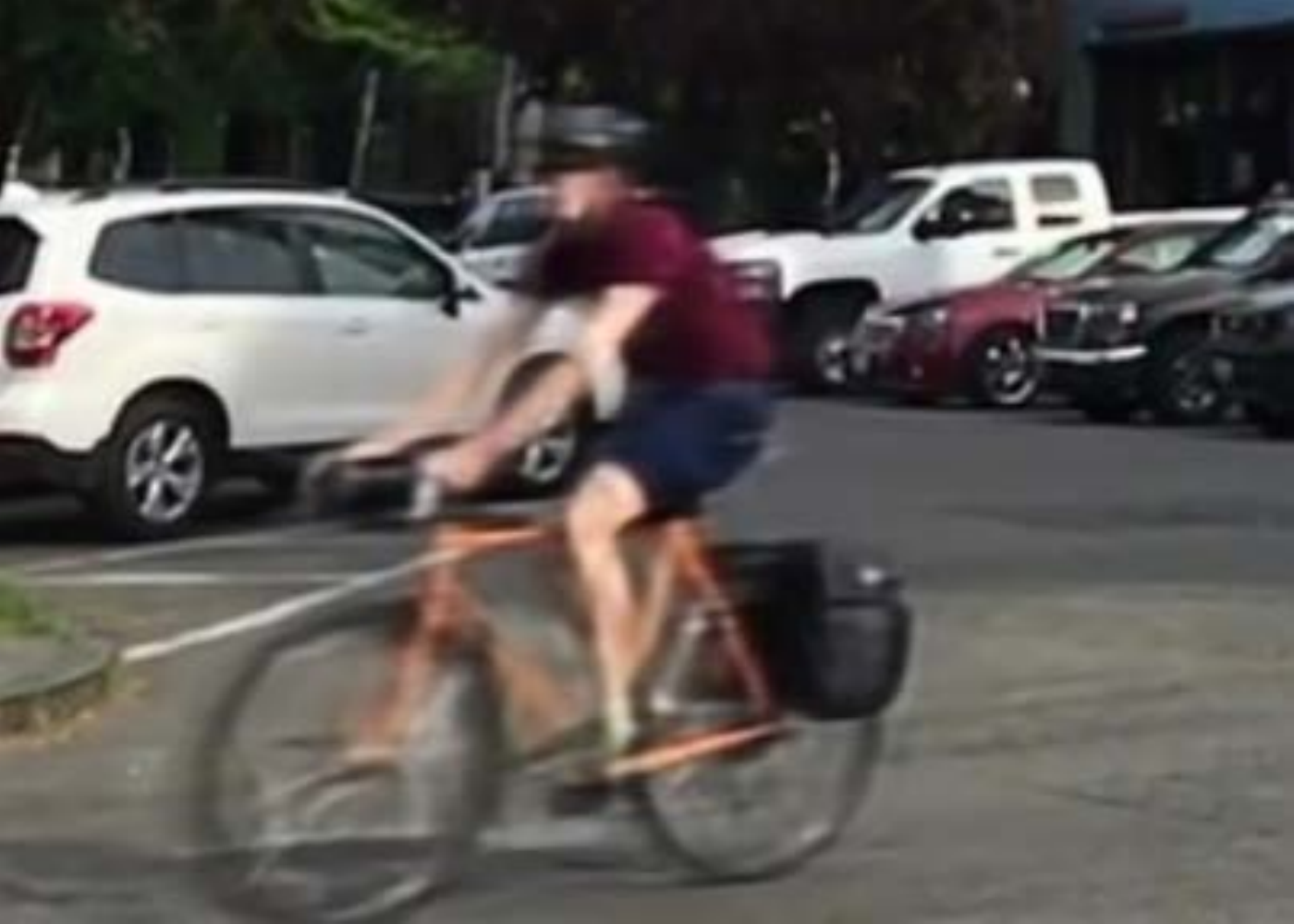} &
			\includegraphics[width=\k\linewidth]{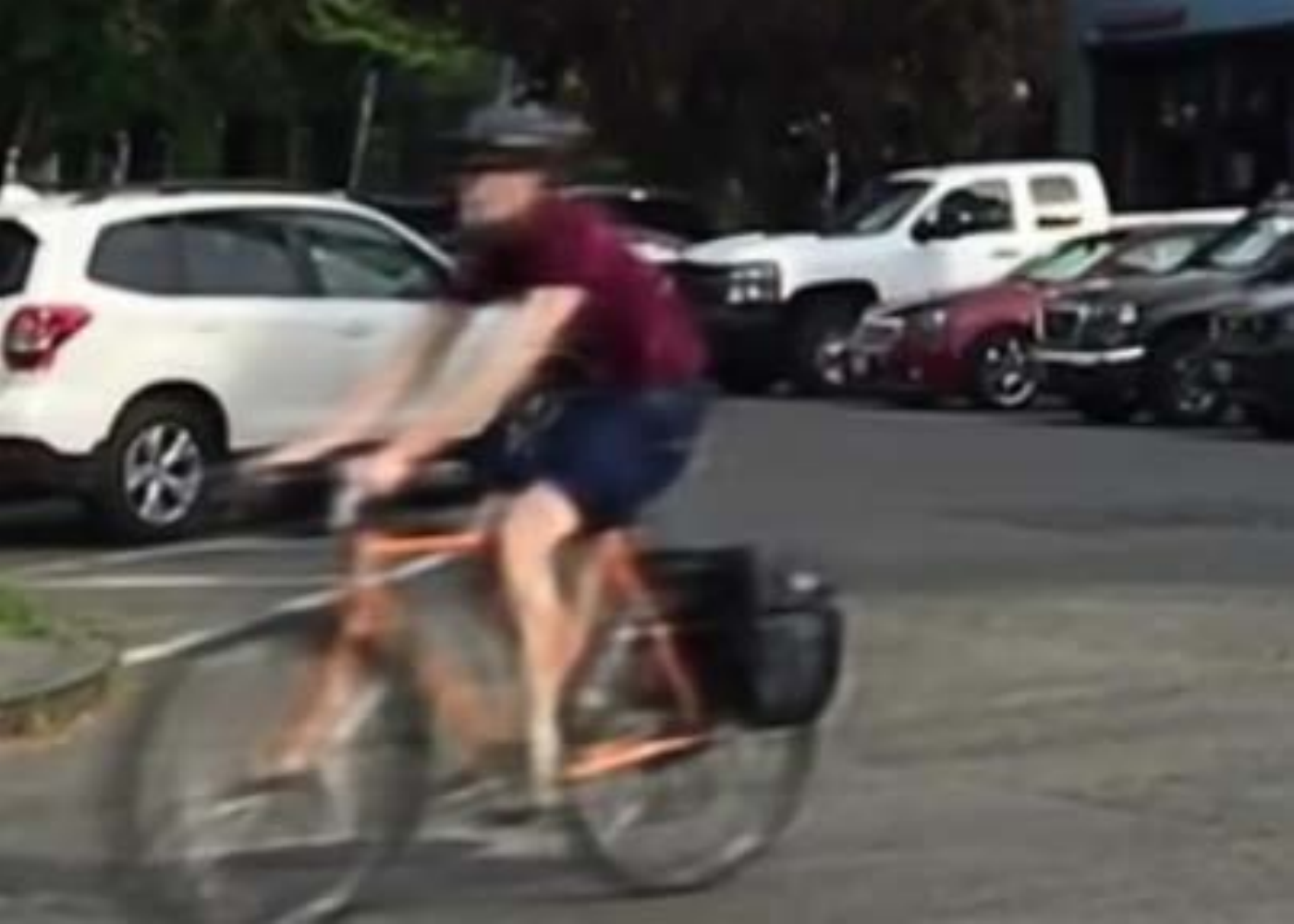} \\
			\begin{sideways}{\tiny \ DVD~\cite{su2017deep}}\end{sideways}&
			\includegraphics[width=\k\linewidth]{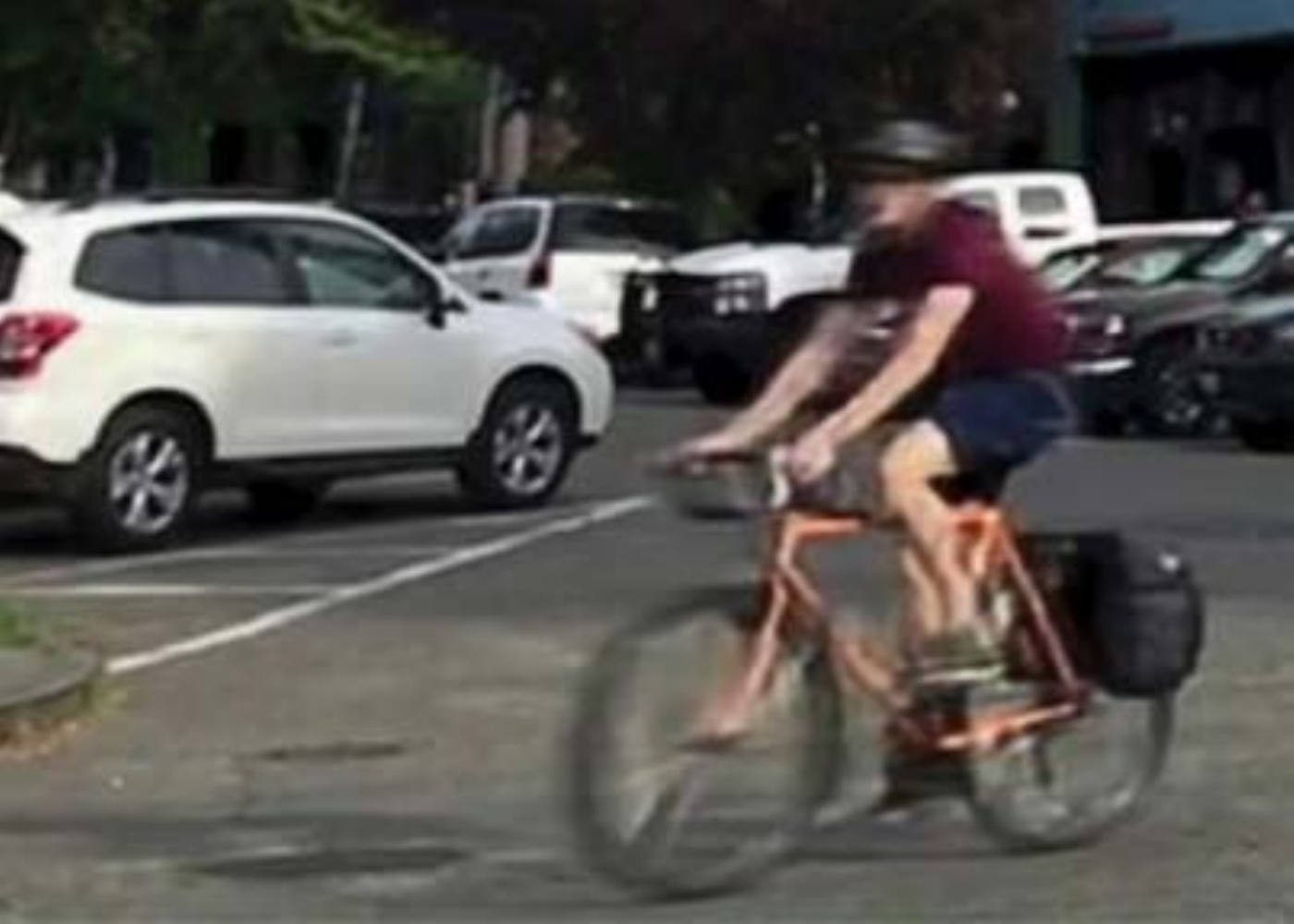} &
			\includegraphics[width=\k\linewidth]{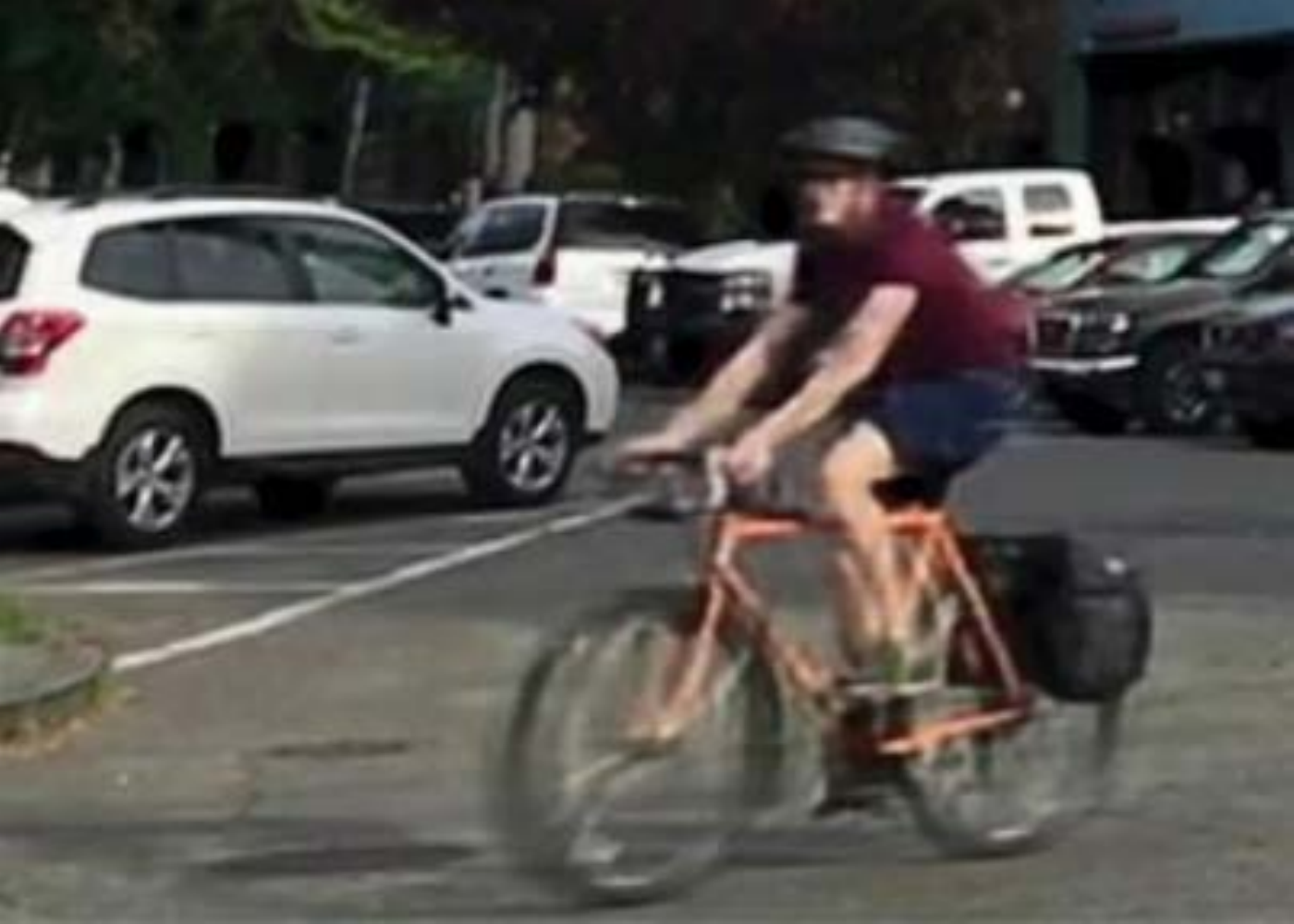} &
			\includegraphics[width=\k\linewidth]{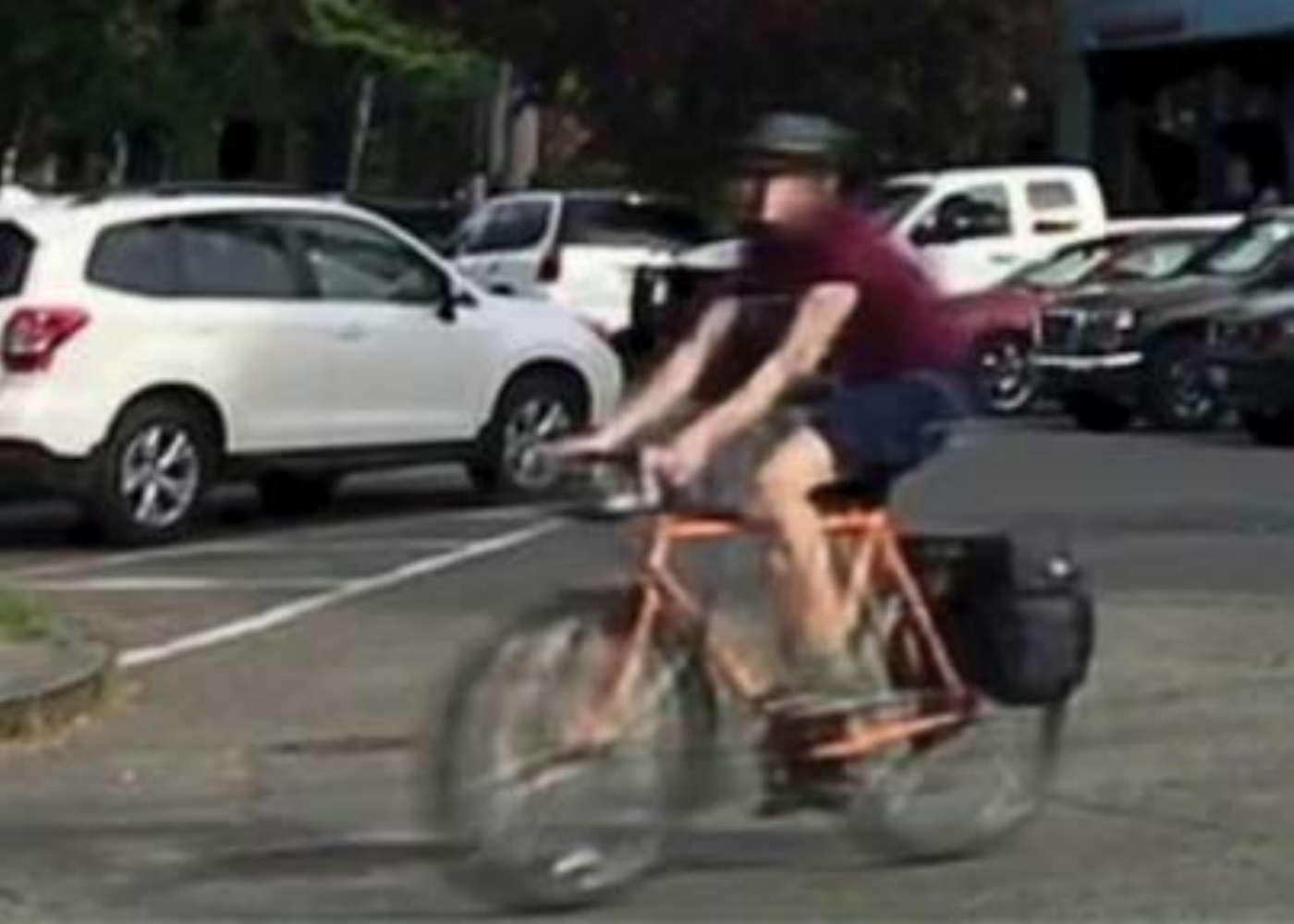} &
			\includegraphics[width=\k\linewidth]{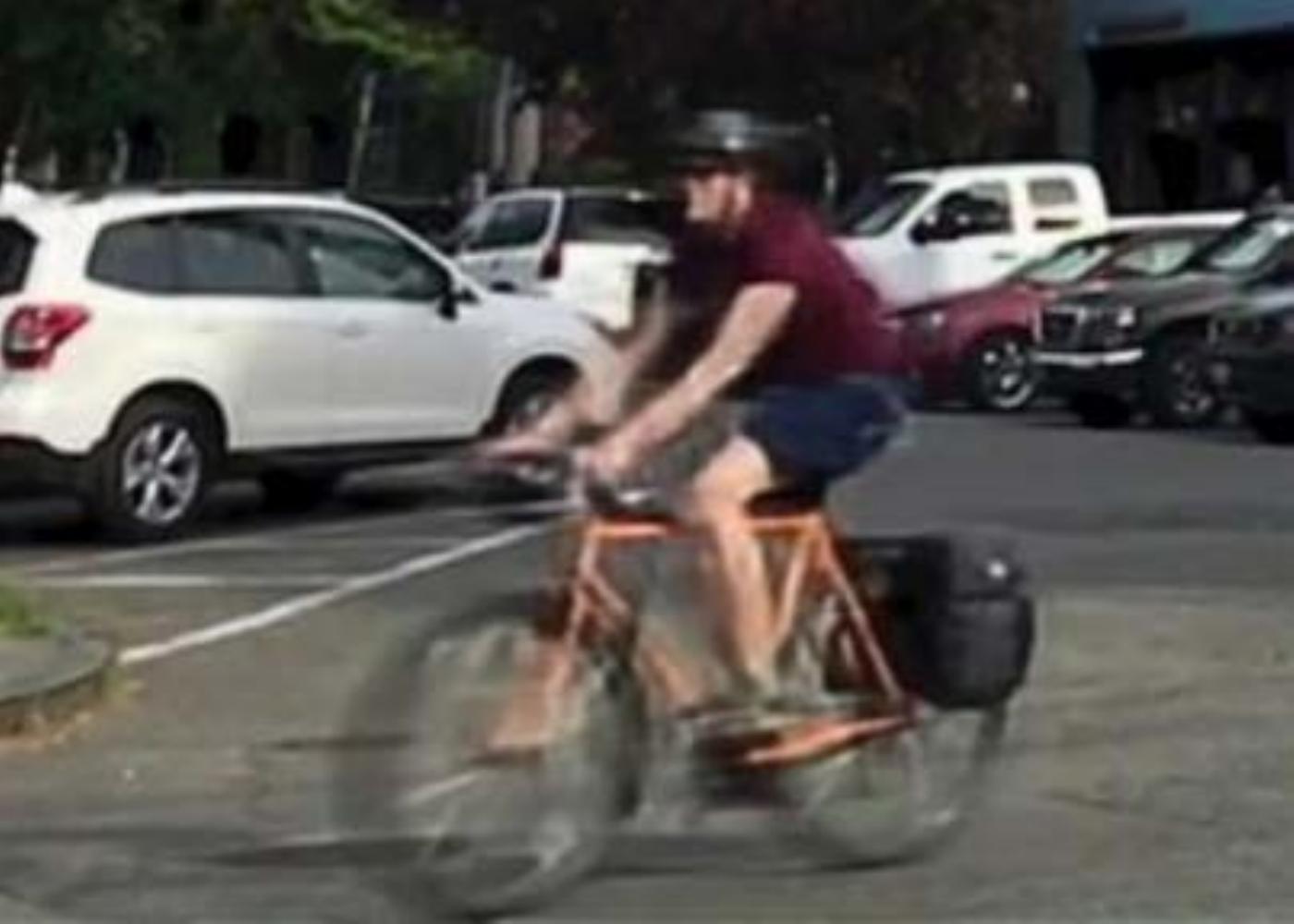} &
			\includegraphics[width=\k\linewidth]{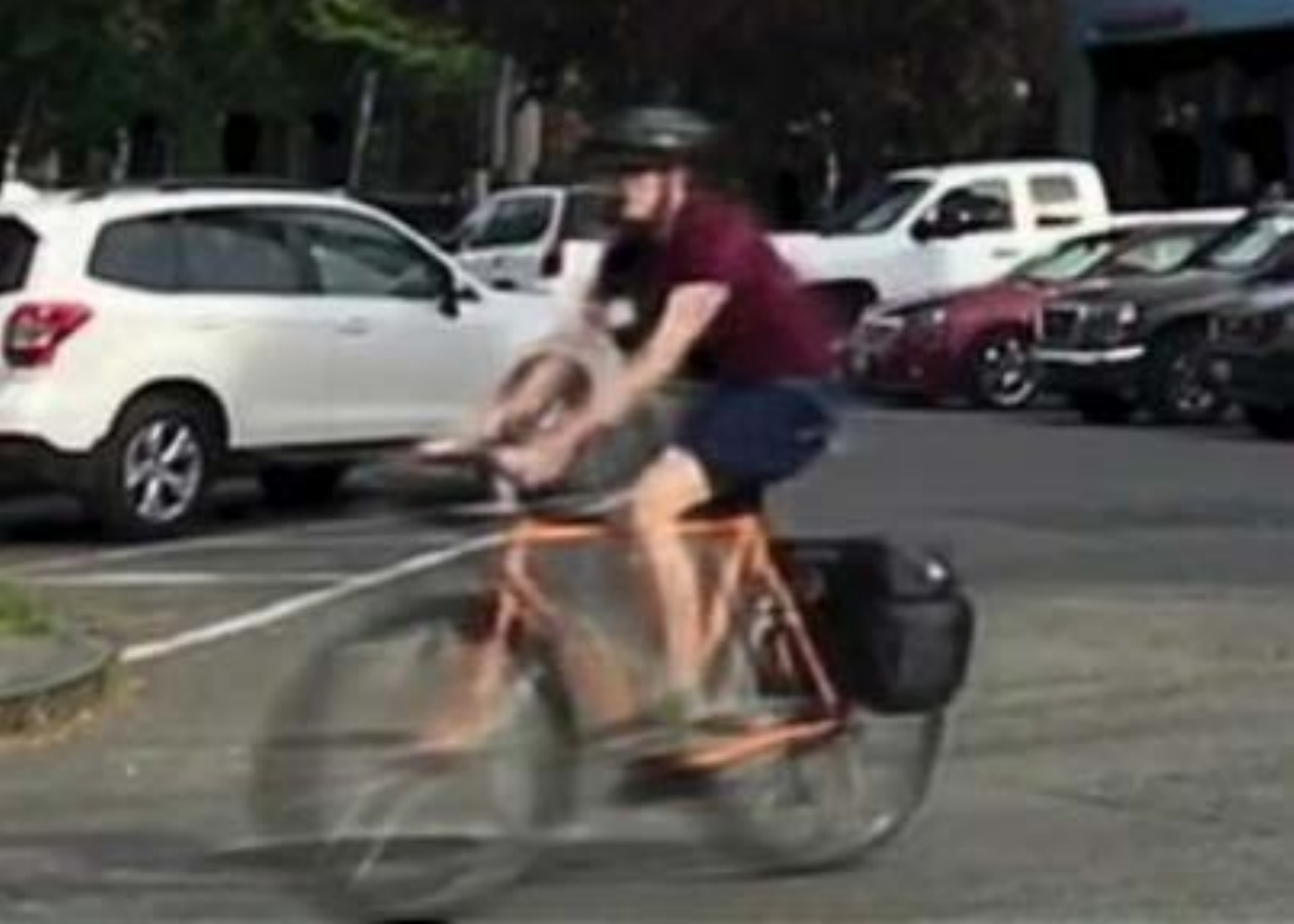} &
			\includegraphics[width=\k\linewidth]{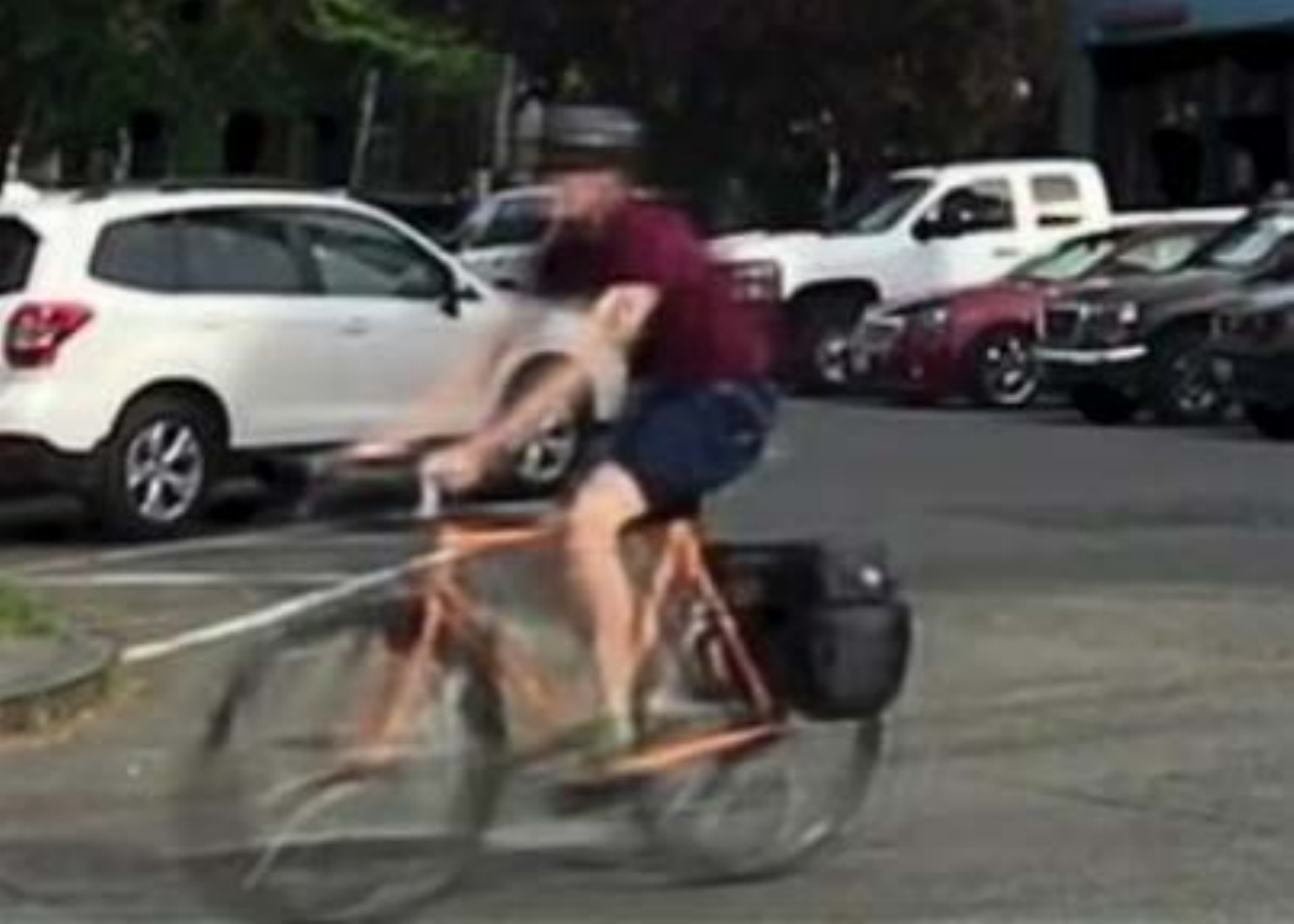} &
			\includegraphics[width=\k\linewidth]{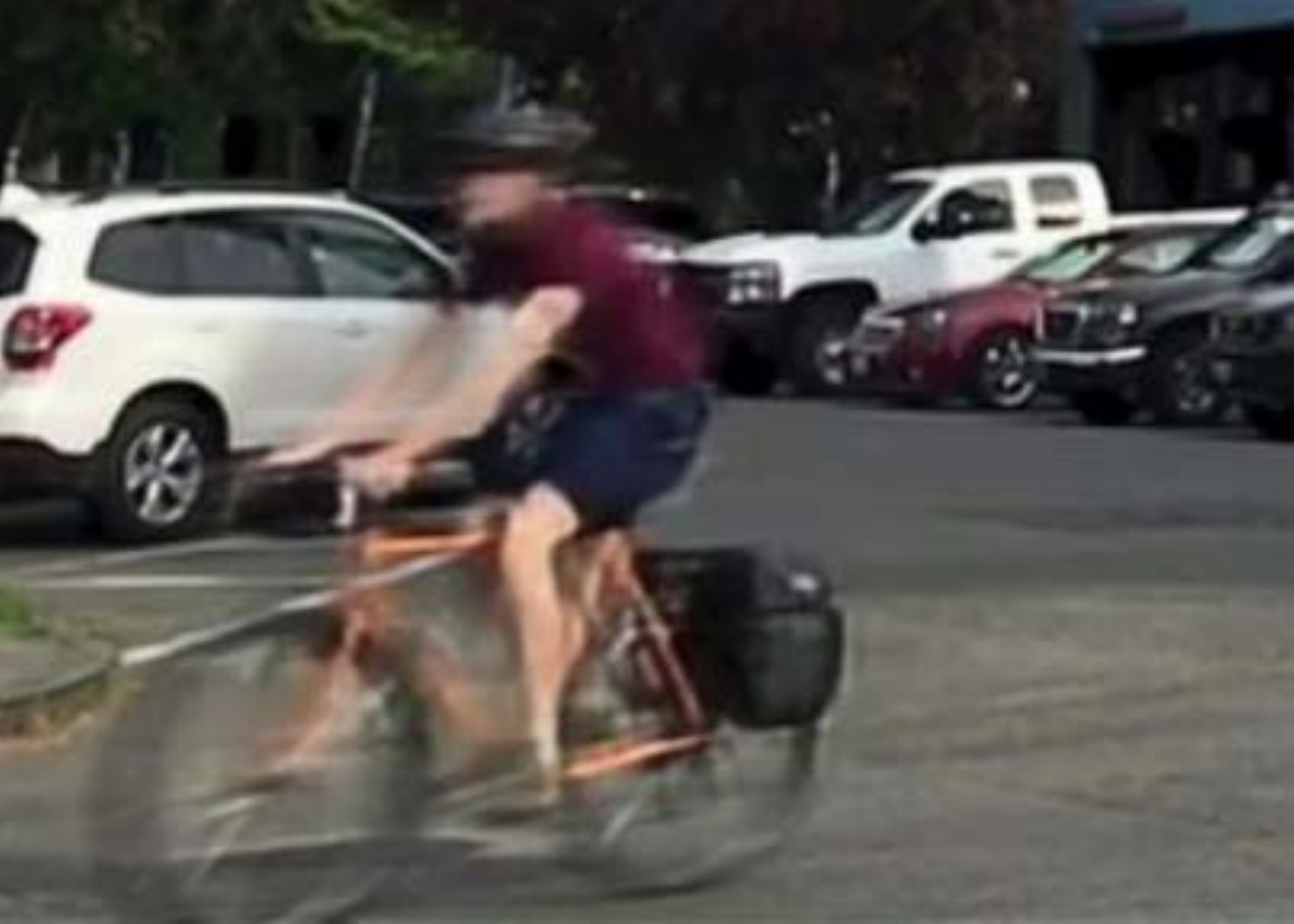} \\
			\begin{sideways}{\tiny\ \ \ \ Ours}\end{sideways}&
			\includegraphics[width=\k\linewidth]{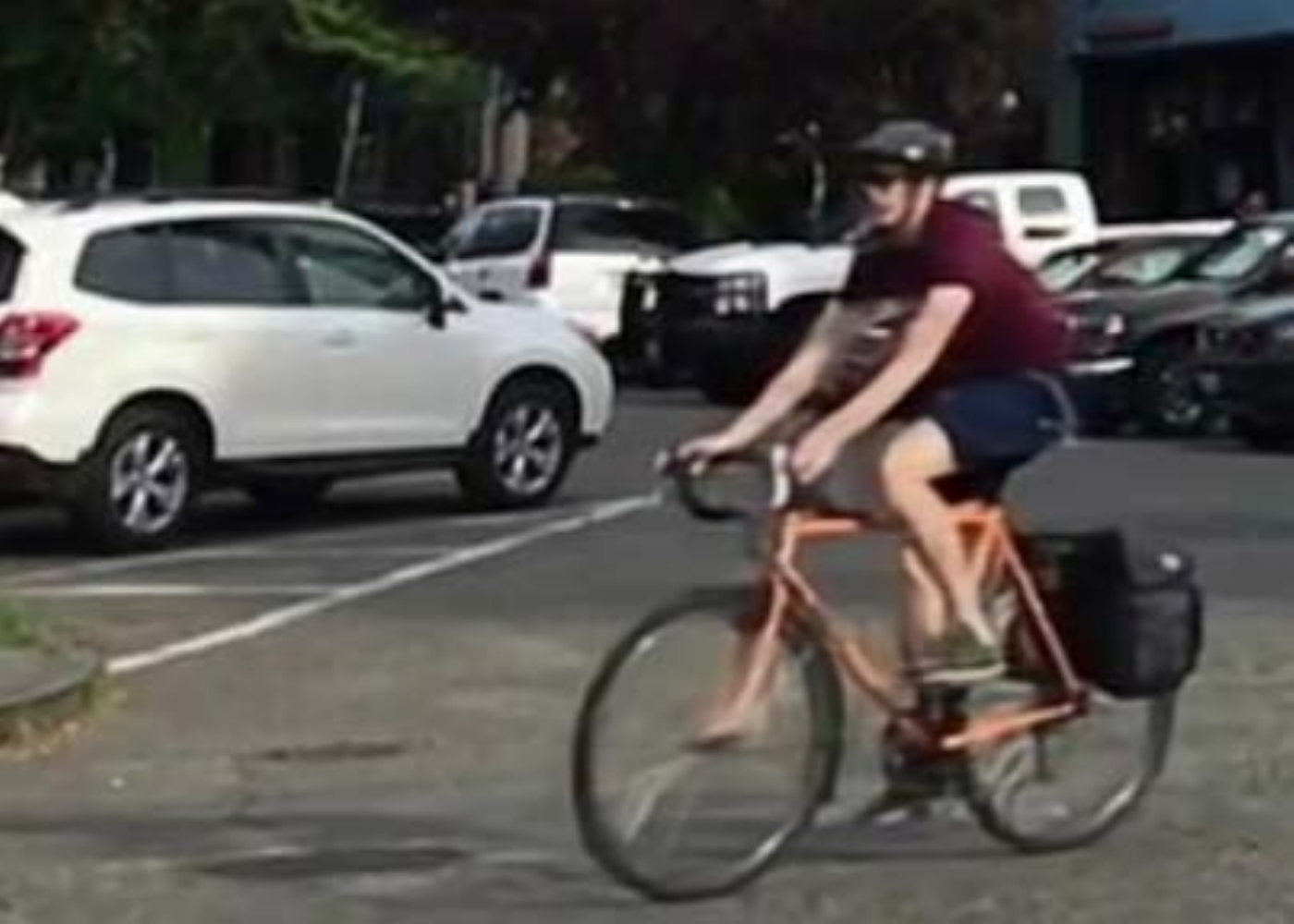} &
			\includegraphics[width=\k\linewidth]{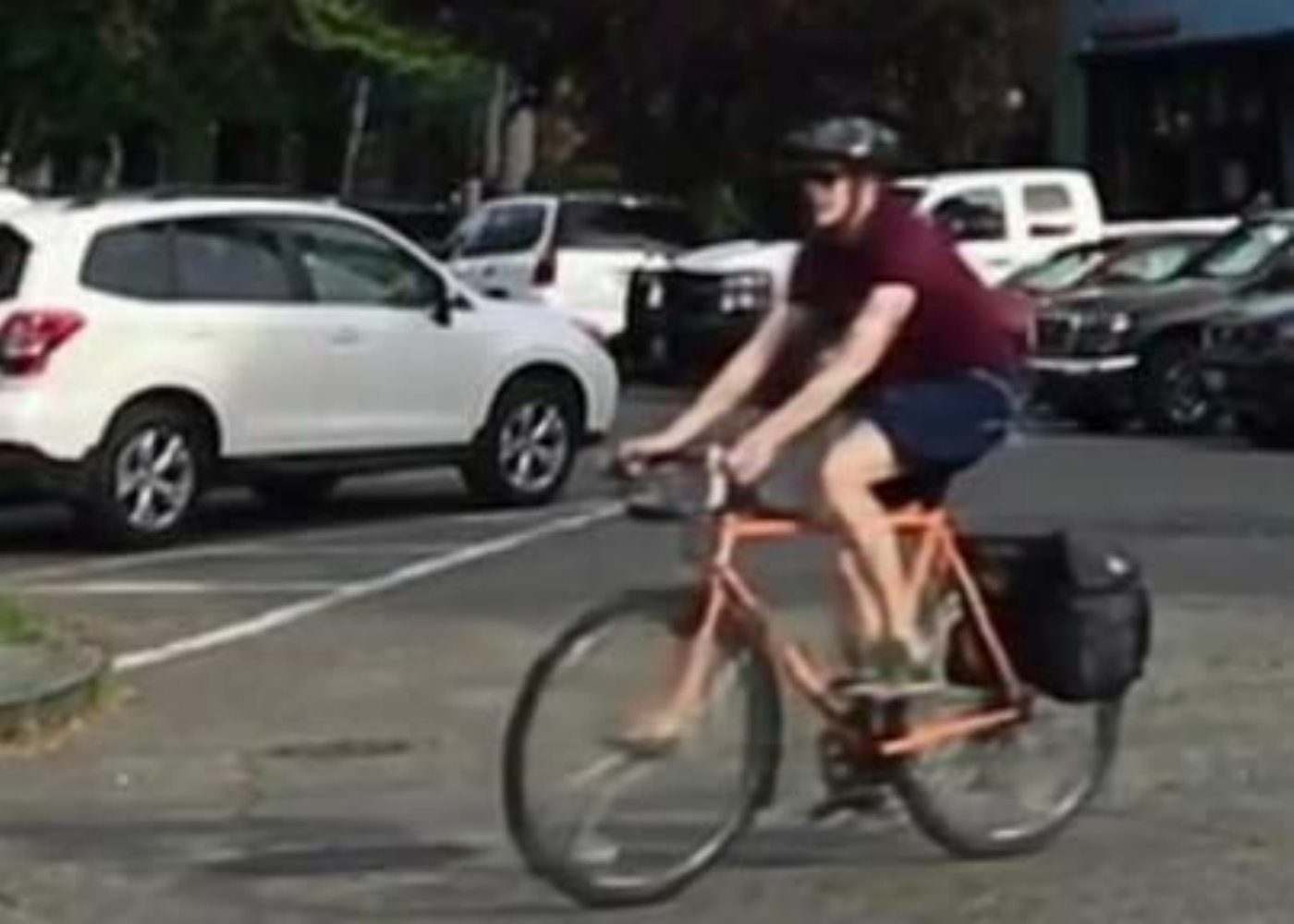} &
			\includegraphics[width=\k\linewidth]{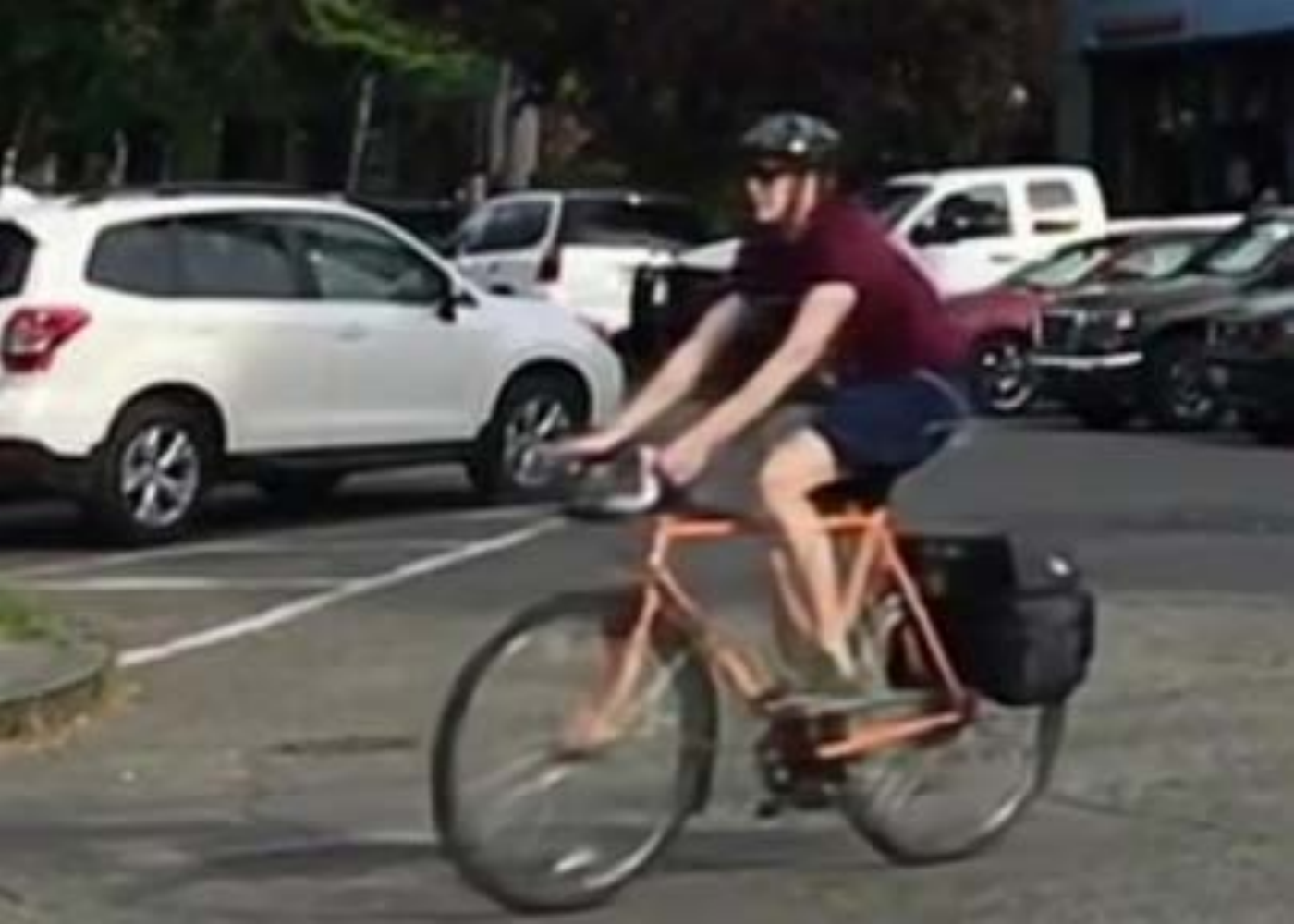} &
			\includegraphics[width=\k\linewidth]{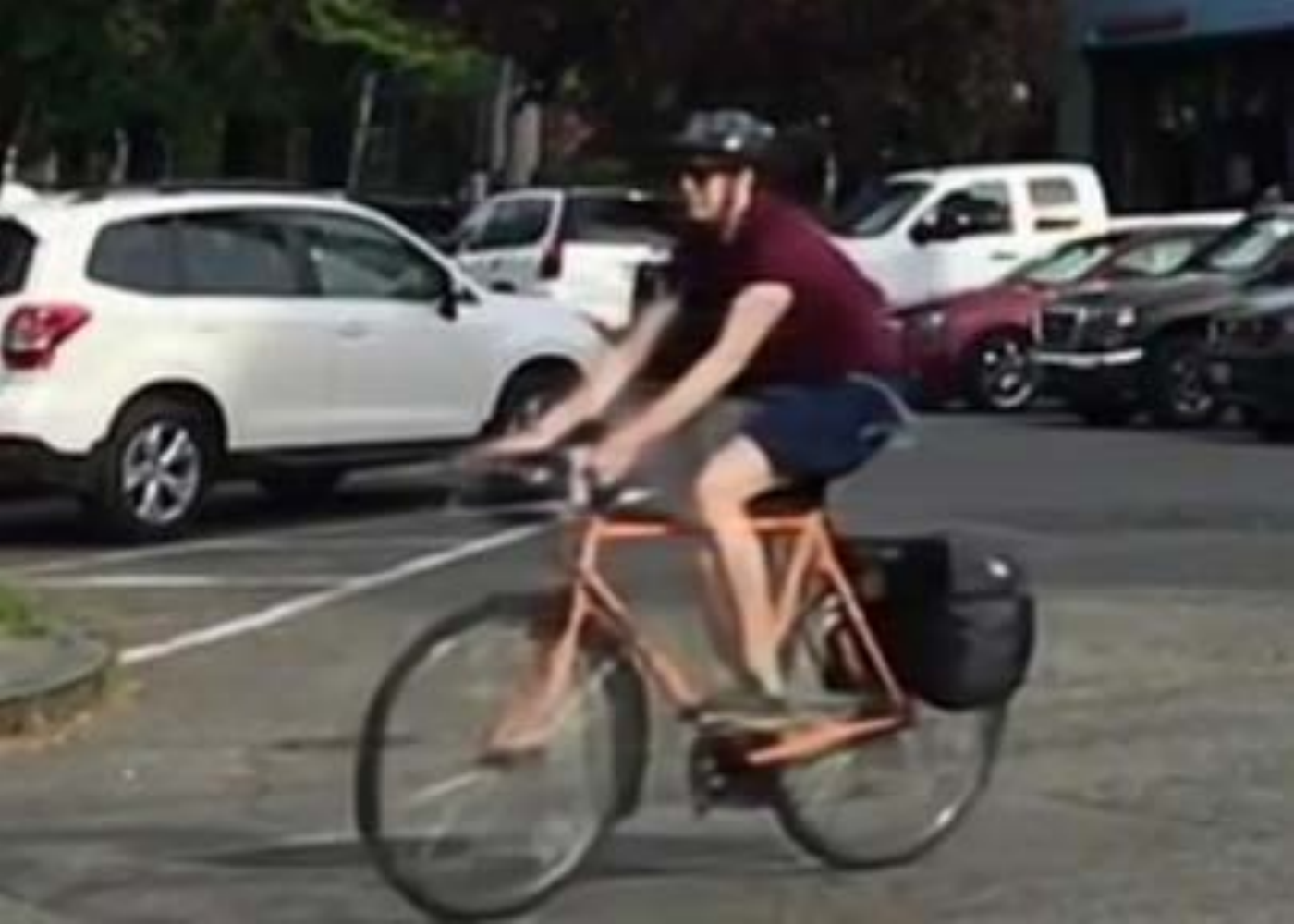} &
			\includegraphics[width=\k\linewidth]{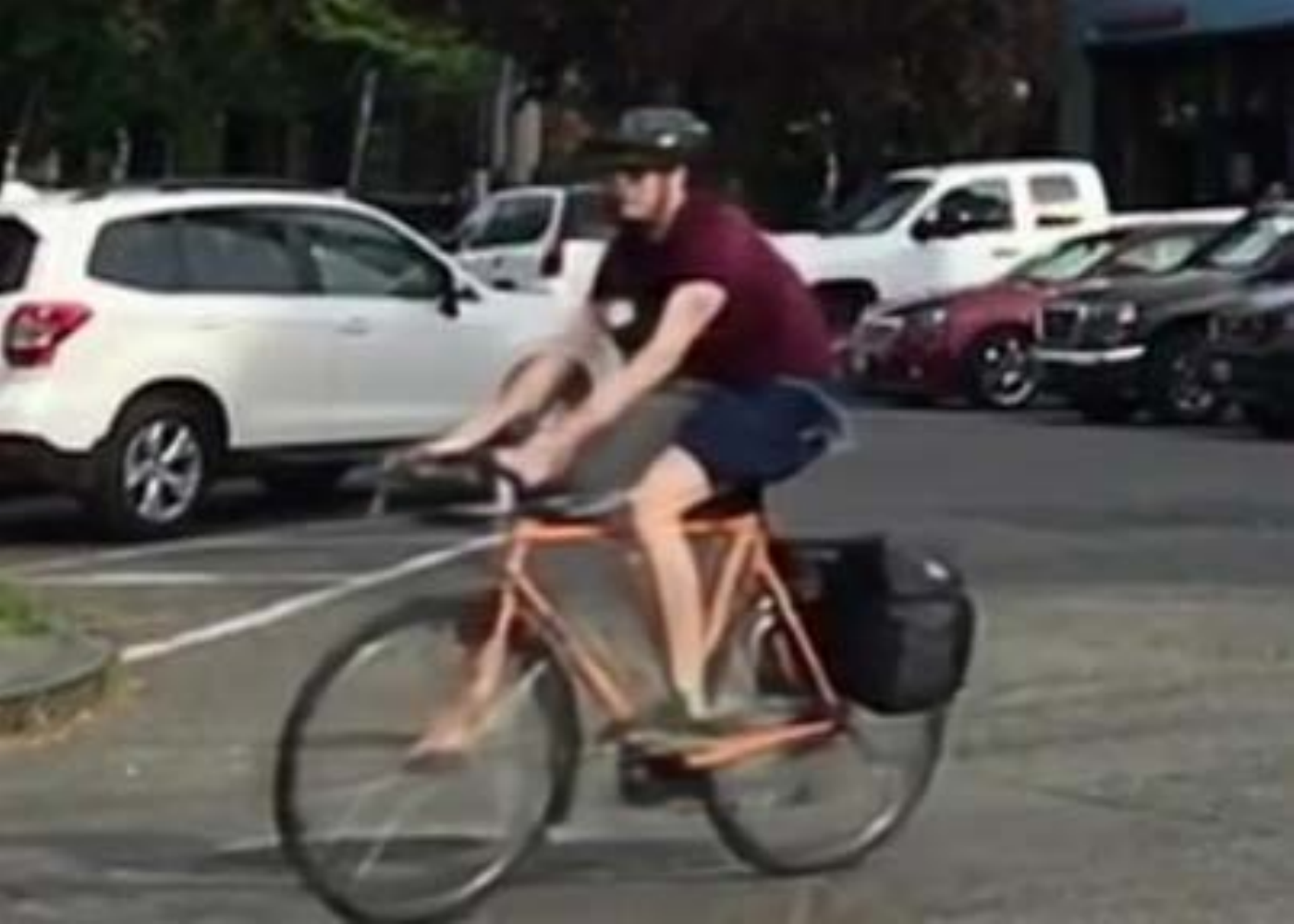} &
			\includegraphics[width=\k\linewidth]{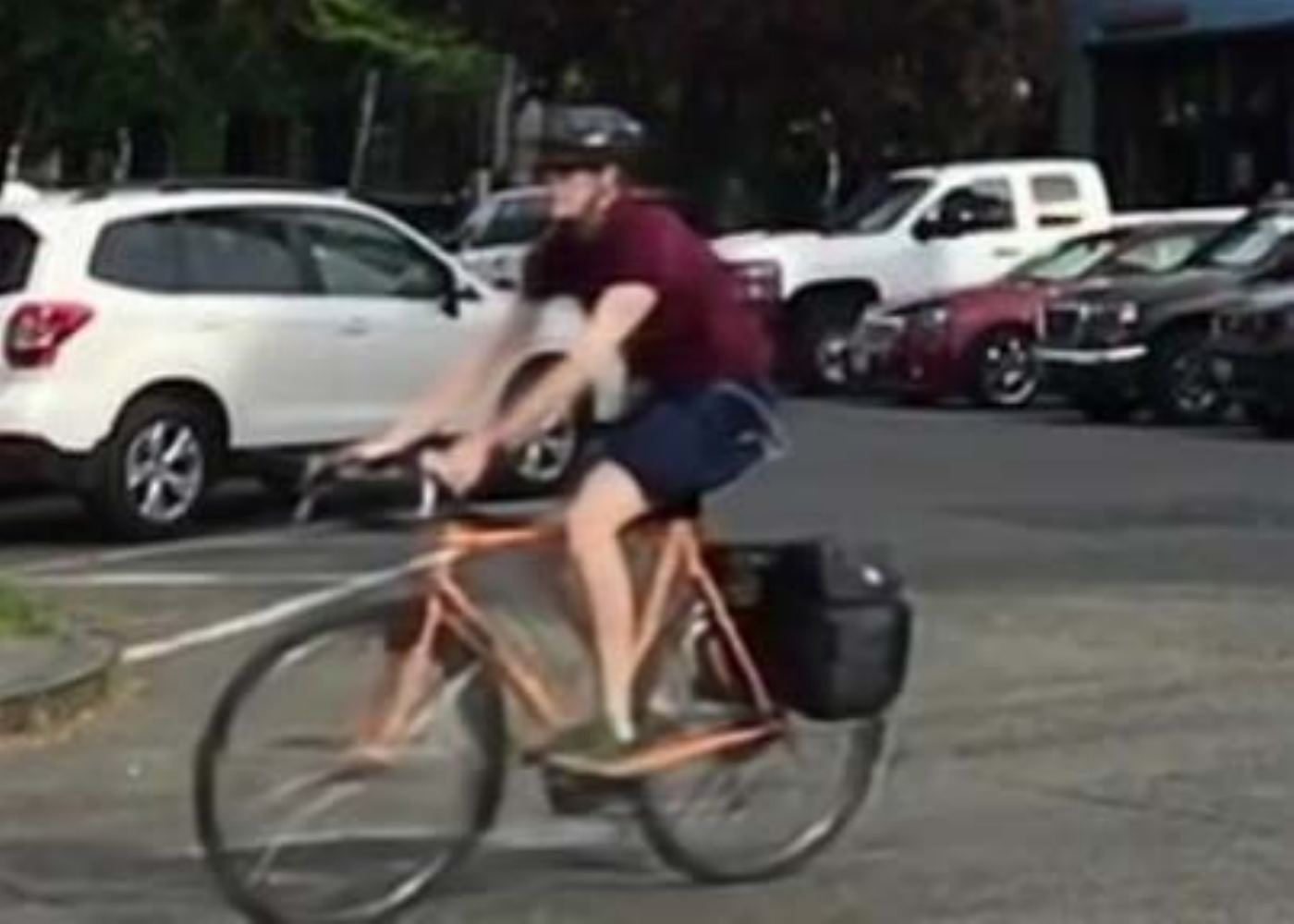} &
			\includegraphics[width=\k\linewidth]{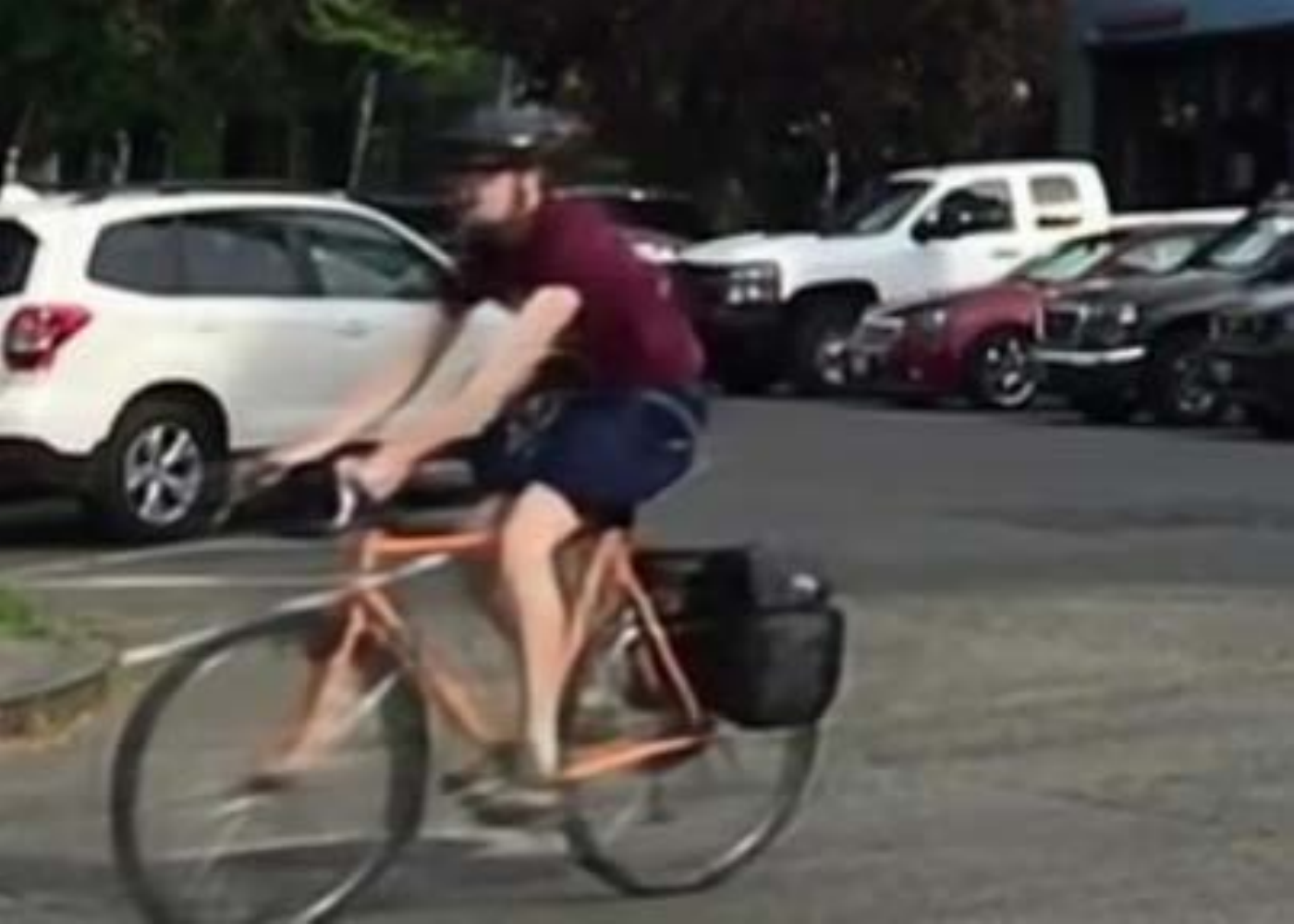} \\
			&T= 0 & T= 1 & T= 2 & T= 3& T= 4 &T= 5&T= 6
		\end{tabular}
	\end{center}%
	\vspace{-4mm}
	\caption{{\small Temporal consistency evaluation on consecutive frames from a blurry video. \textbf{(zoom in for best view)}.}}
	\label{fig:temporal}
	\vspace{-3mm}
\end{figure}

To enforce temporal consistency, we adopt the recurrent network to transfer previous feature maps over time,
and propose the FAC layer for propagating information between consecutive frames via explicit alignment.
Fig.~\ref{fig:temporal} shows that our method not only restores sharper frames but also keeps better temporal consistency.
In addition, the video results are given on our [\href{https://shangchenzhou.com/projects/stfan/}{project webpage}].
\section{Analysis and Discussions}
We have shown that the proposed algorithm performs favorably against state-of-the-art methods.
In this section, we conduct a number of comparative experiments for ablation study and analysis further.
\begin{figure*}[t]\footnotesize
\centering
\renewcommand{\tabcolsep}{1pt}
\renewcommand{\arraystretch}{1}
\begin{center}
	\begin{tabular}{ccccc}
		\includegraphics[width=0.192\linewidth]{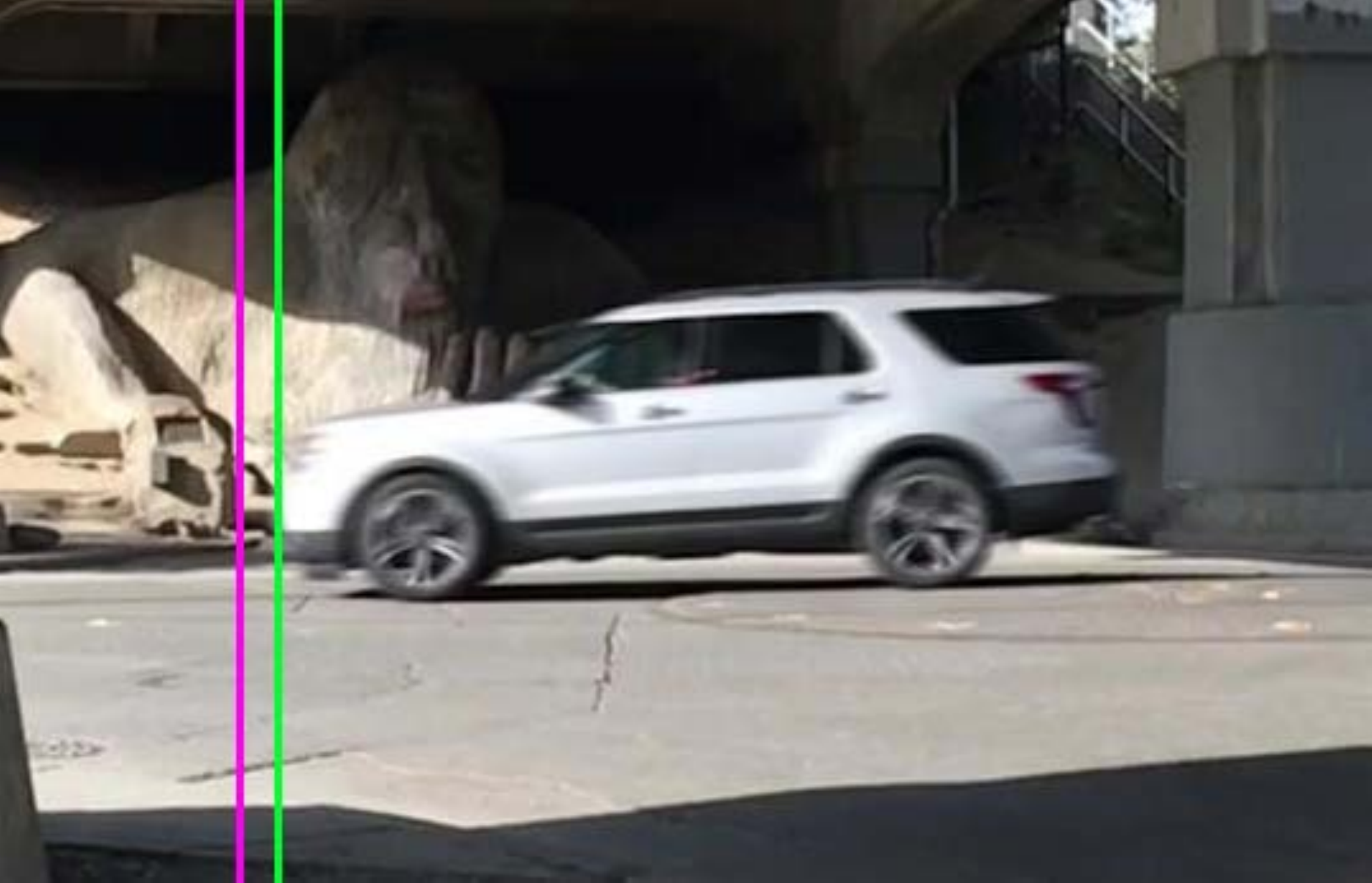} &
		\includegraphics[width=0.192\linewidth]{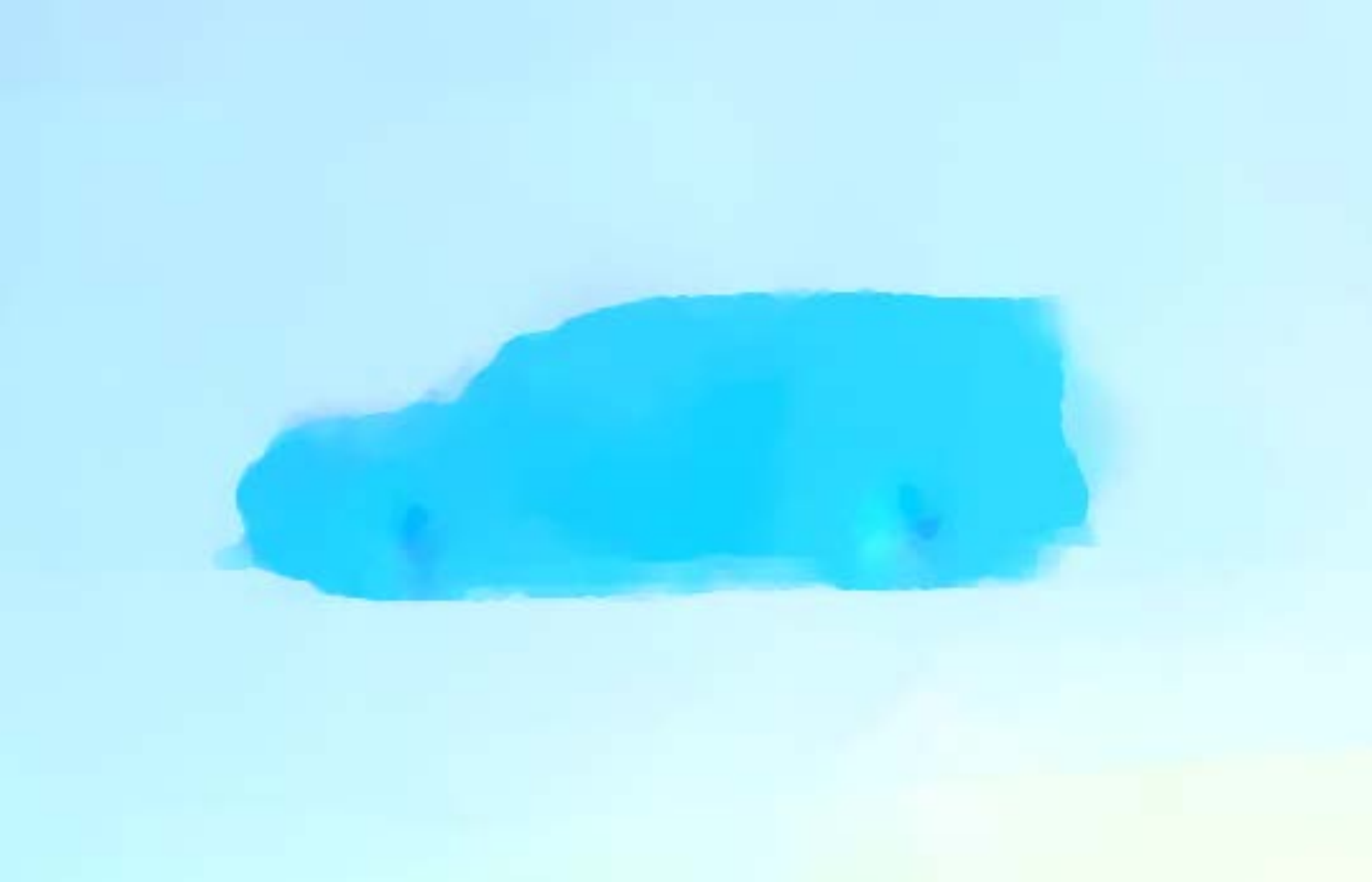} &
		\includegraphics[width=0.192\linewidth]{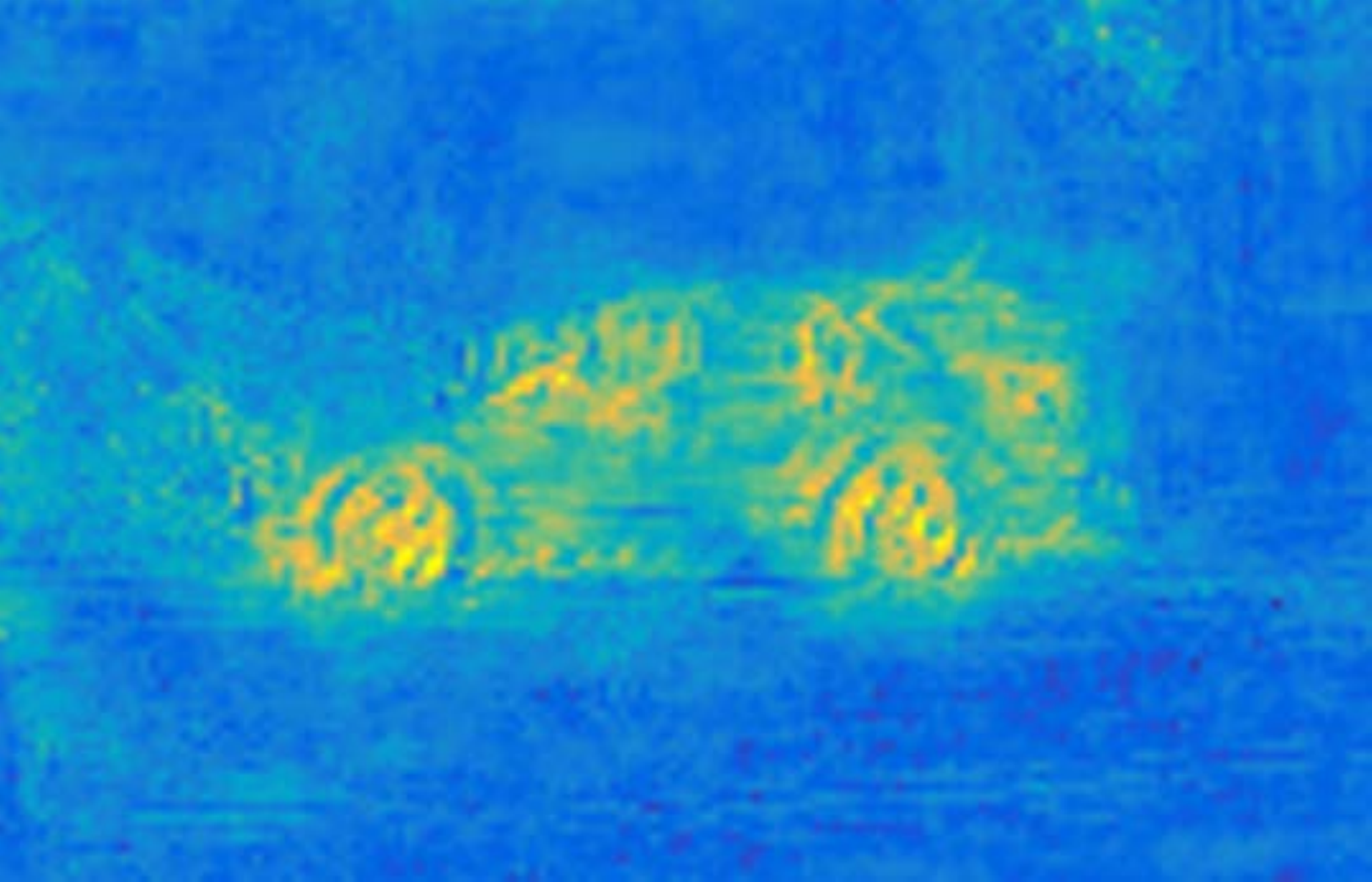} &
		\includegraphics[width=0.192\linewidth]{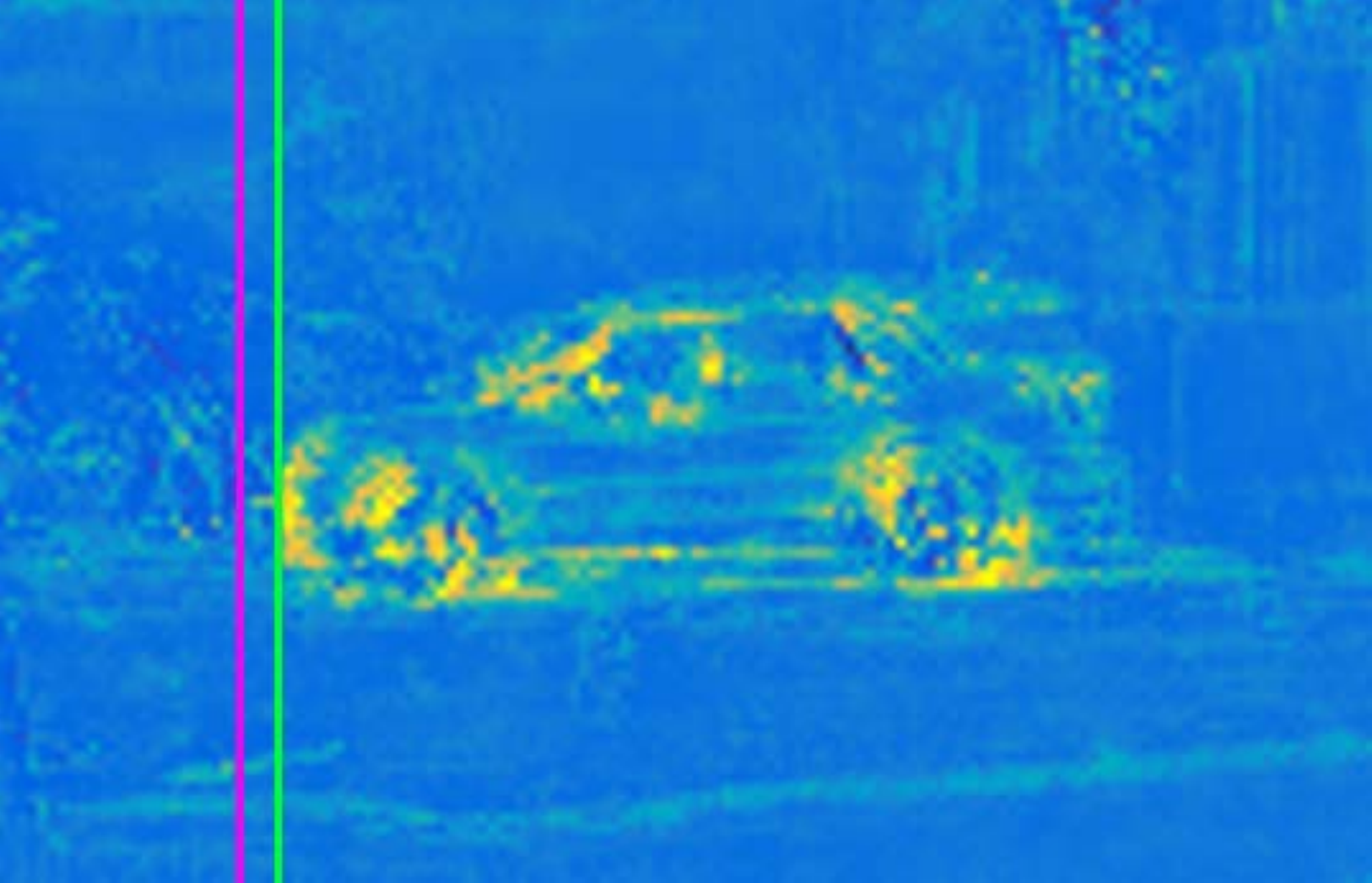} &
		\includegraphics[width=0.192\linewidth]{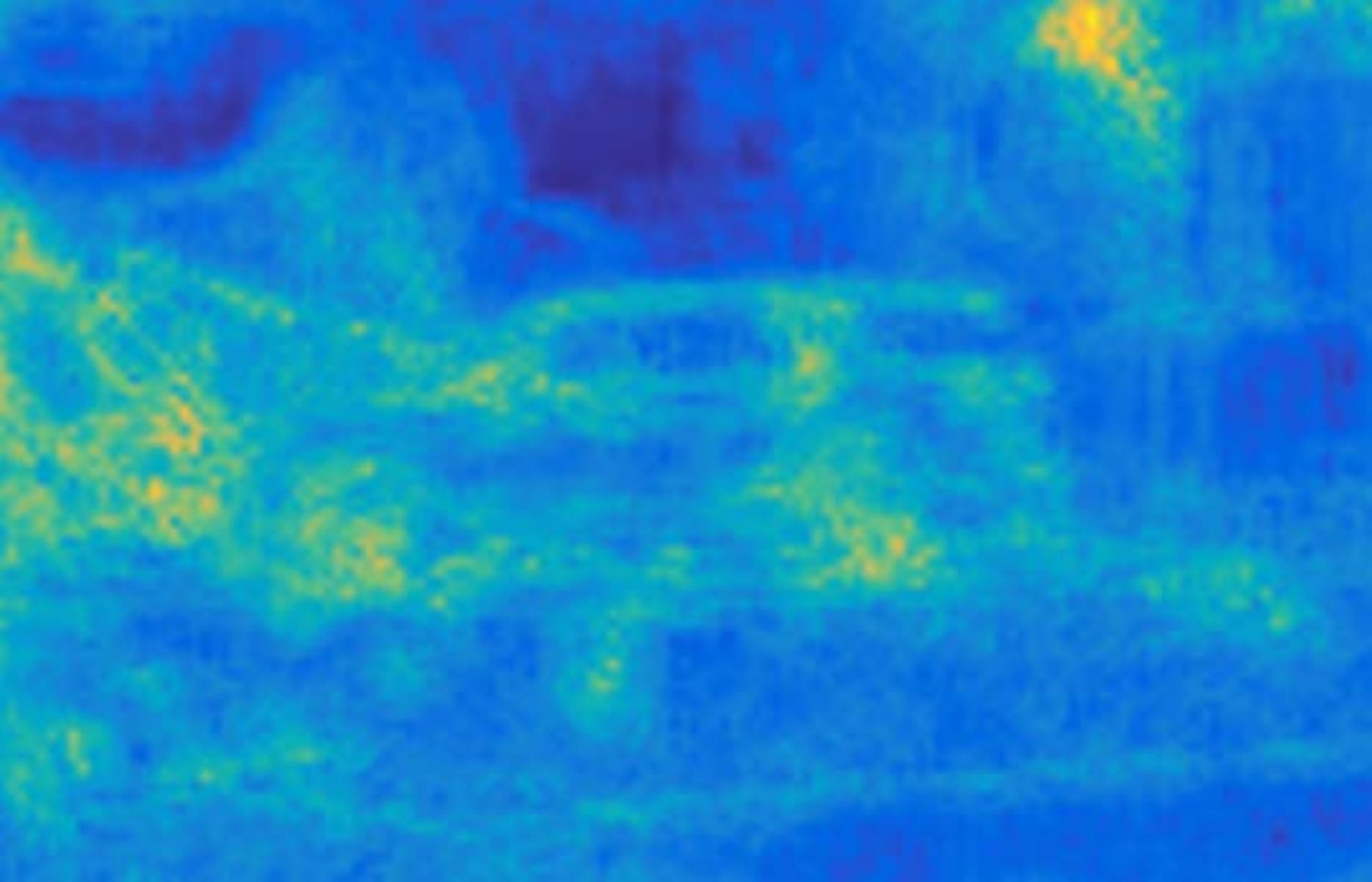} \\
		(a) Blurry image $B_{t-1}$ & (b) Optical flow & (c) Alignment filters & (d) Before alignment & (e) Before deblurring
		\vspace{1.5pt}\\
		\includegraphics[width=0.192\linewidth]{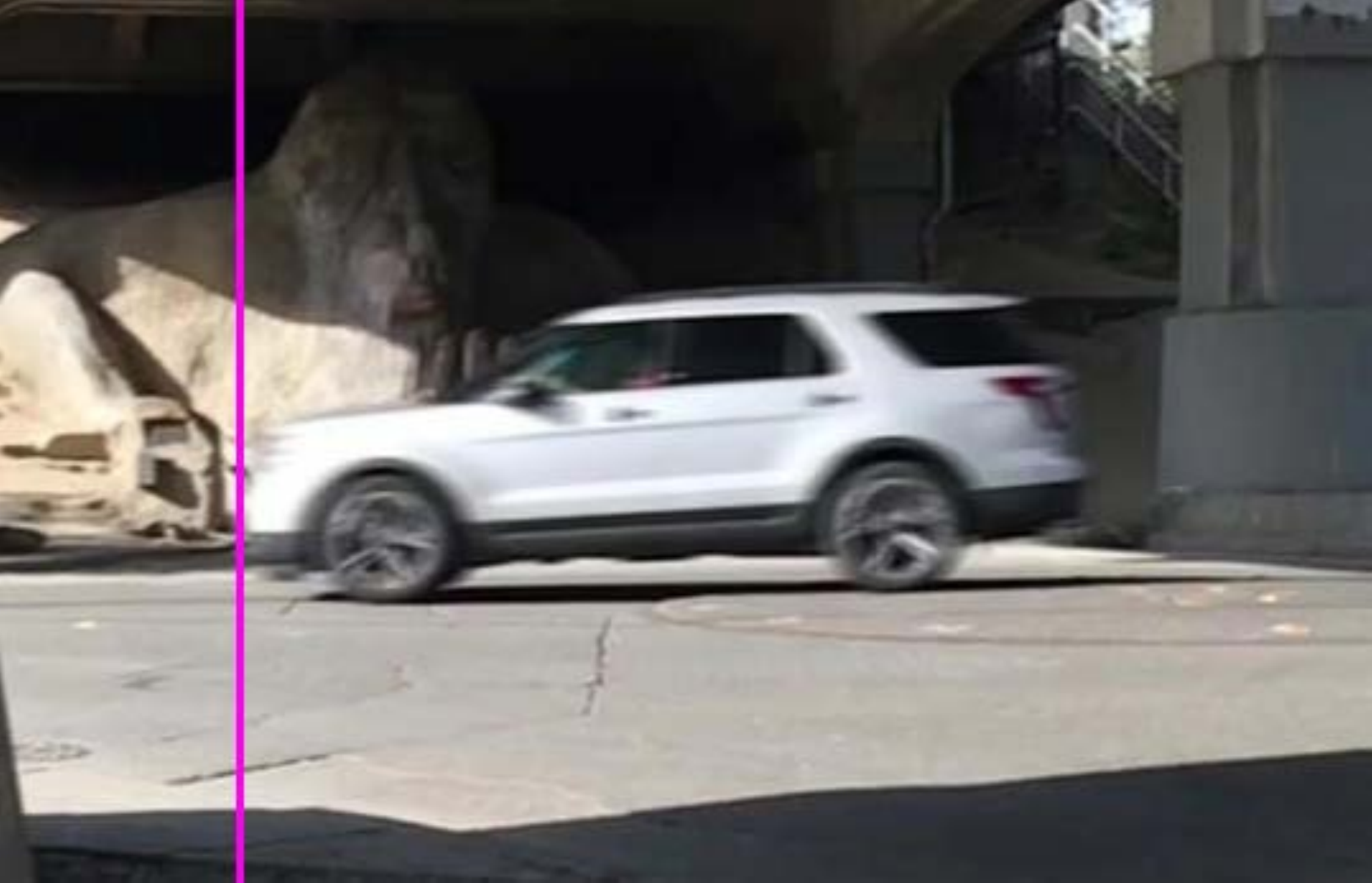} &
		\includegraphics[width=0.192\linewidth]{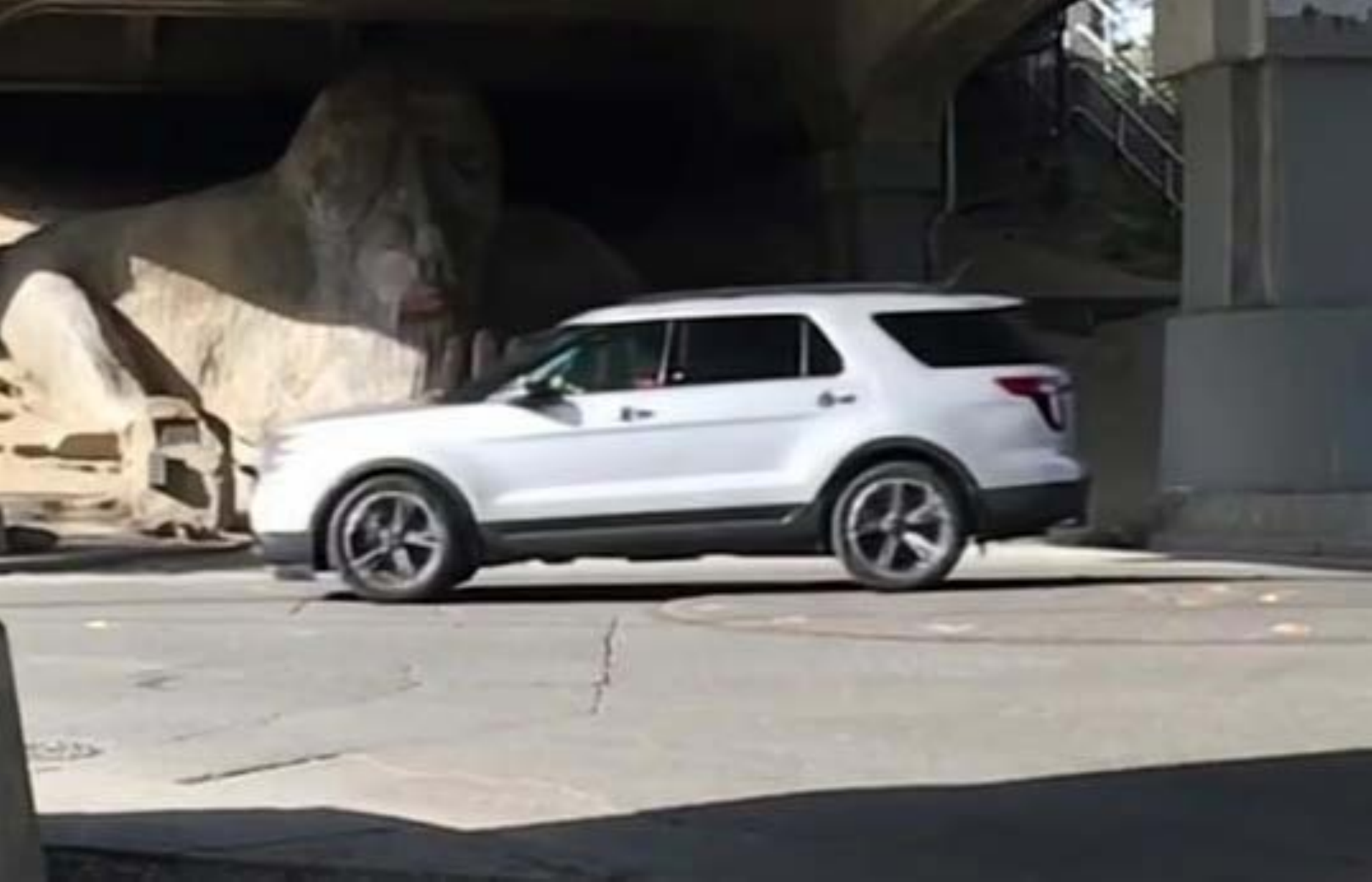}&
		\includegraphics[width=0.192\linewidth]{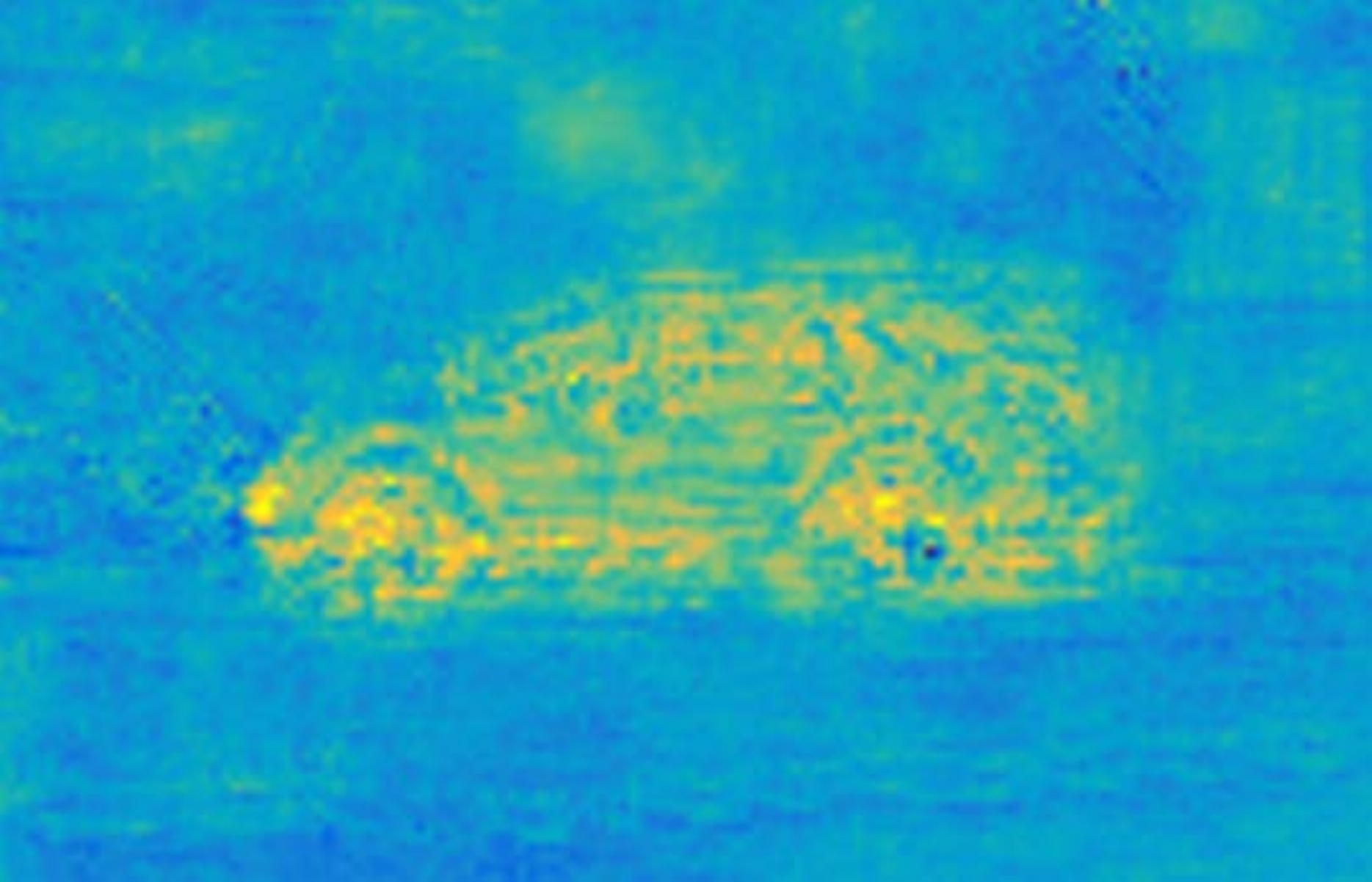}&
		\includegraphics[width=0.192\linewidth]{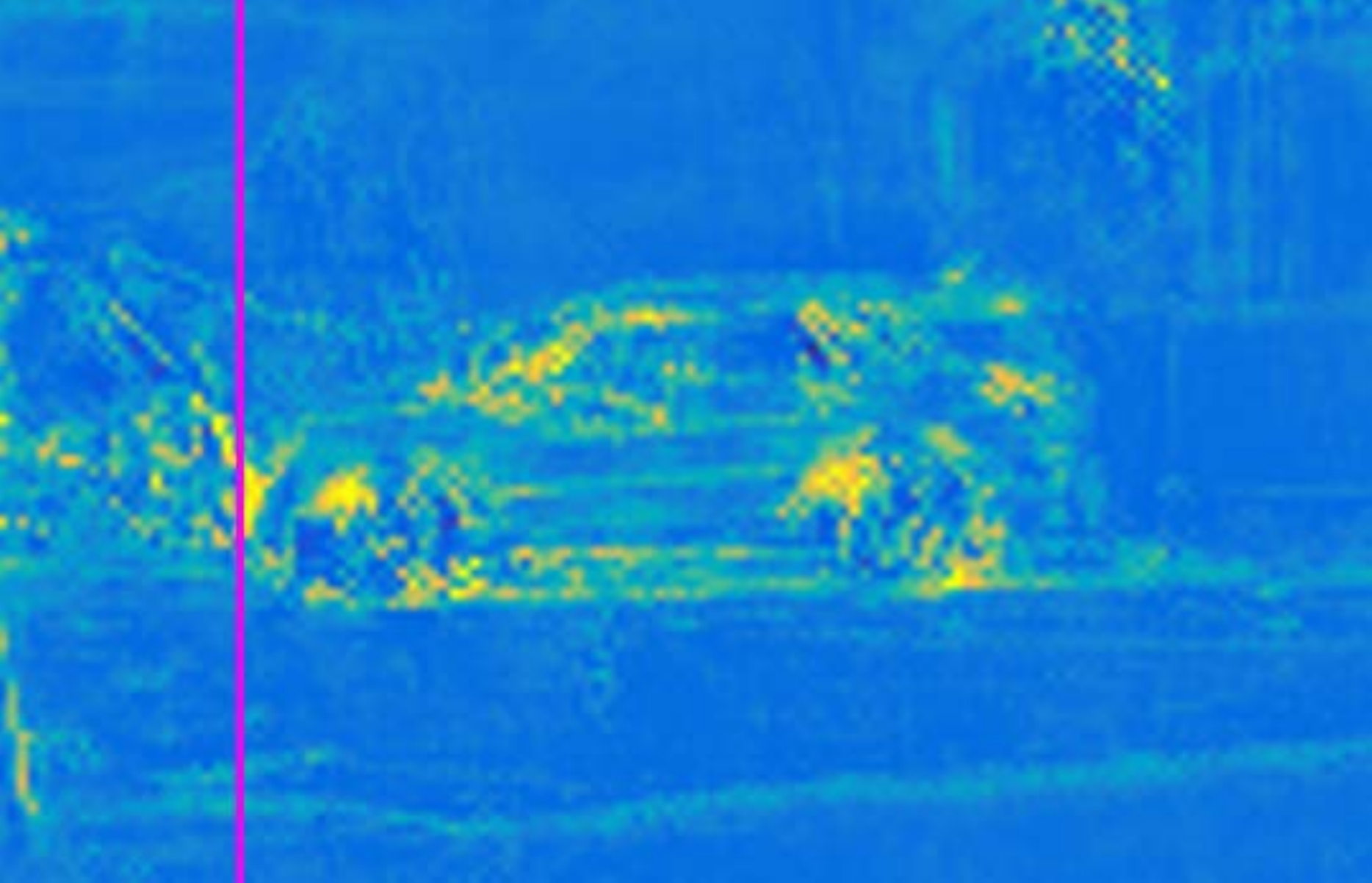}&
		\includegraphics[width=0.192\linewidth]{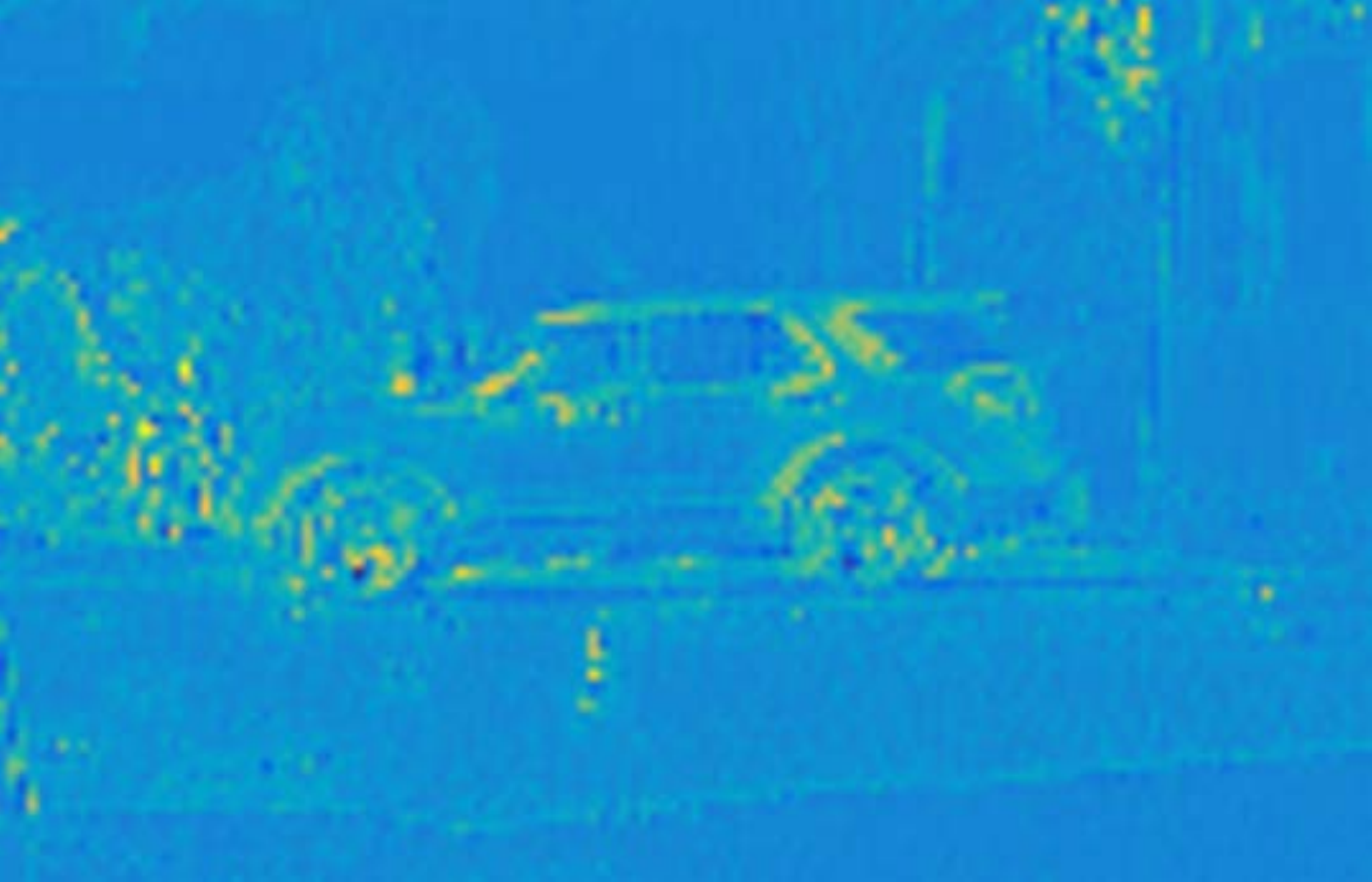}\\
		(f) Blurry image $B_{t}$  & (g) Restored image & (h) Deblurring filters & (i) After alignment  & (j) After deblurring\\
	\end{tabular}
\end{center}
\vspace{-4mm}
\caption{Effectiveness of the adaptive filter generator and  FAC layer.
	(b) is the optical flow from the adjacent input blurry frames (a) and (f) according to EpicFlow~\cite{revaud2015epicflow}.
	(c) and (h) are the visualization of the generated alignment and deblurring filters of FAC layers, respectively.
	(d) and (i) are selected feature maps before and after alignment using FAC layer.
	(e) and (j) are selected feature maps before and after deblurring using FAC layer.
	}
\label{fig:show_feas}
\vspace{-3mm}
\end{figure*}
\subsection{Effectiveness of the FAC layers}
The generated alignment filters and deblurring filters are visualized in Figure~\ref{fig:show_feas}(c) and (h), respectively.
According to the optical flow estimated by EpicFlow~\cite{revaud2015epicflow} in Figure~\ref{fig:show_feas}(b), there is a vehicle moving in the video which is coherent with the alignment filters estimated by our network.
Since removing different blur requires different operations and blur is somehow related to the optical flow, our network estimates different deblurring filters for foreground vehicle and backgrounds.

To validate the effectiveness of the FAC layer for alignment and deblurring, some intermediate features are shown in Figure~\ref{fig:show_feas}.
According to Figure~\ref{fig:show_feas}(d) and (i), the FAC layer for alignment can correctly warp the head of the vehicle from green line to purple line even without an image alignment constraint during training.
As for the transformed features in Figure~\ref{fig:show_feas}(j) for deblurring, they are sharper than those before the FAC layer in Figure~\ref{fig:show_feas}(e), which means the deblurring branch can effectively remove blur in the feature domain.

We also conduct three experiments which replace one or both the FAC layers by concatenating the corresponding features directly, without features transformation by FAC layers.
In Table~\ref{tab:structure}, (w/o A, w/ D), (w/ A, w/o D) and (w/o A, w/o D) represent removing FAC layers for feature domain alignment only, feature domain deblurring only and both of them, respectively (refer to Figure~\ref{fig:networks} for clarification).
It shows that the network performs worse without the help of the feature transformation by FAC layers.
In addition, Figure~\ref{fig:head_pic} also shows that our method cannot restore such a sharp image without using FAC layers.
\begin{table}[h]
	\centering
	\caption{Results of different variants of structures. 
		The (w/o A, w/ D), (w/ A, w/o D) and (w/o A, w/o D) represent removing FAC layers for alignment only, deblurring only and both of them, respectively. 
		Unlike the above variants still considering nonalignment features, (-, w D) and (w A, -) denote removing the features of the alignment branch and removing the features of the deblurring branch.}
	\vspace{-1mm}
	\resizebox{0.97\linewidth}{!} {
		\begin{tabular}{lcccccccc}
			\toprule
			\multirow{2}{*}{Structure} & w/o A & w/o A  & w A     && - & w A      &  \multirow{2}{*}{Ours} \\ \cline{2-4} \cline{6-7} \vspace{-3.5mm}\\
			& w/o D & w D     & w/o D  && w D    & -  &   \\
			\midrule
			PSNR          & 29.91       & 30.92       &  30.59      &&  30.80   & 30.29    & \bf{31.24}      \\
			SSIM           & 0.919       & 0.931       & 0.926       &&  0.929   & 0.924    & \bf{0.934}    \\
			\bottomrule
		\end{tabular}
	}
	\label{tab:structure}
	\vspace{-4mm}
\end{table}
\subsection{Effectiveness of the A and D Branches}
To validate the effectiveness of both alignment (A) and deblurring (D) branches, we compare our network with two variant networks:
removing the features of the alignment branch (-, w D)
and removing the features of the deblurring branch (w A, -).
According to Table~\ref{tab:structure}, these two baseline networks do not generate satisfying deblurring results compared to our proposed method.
\subsection{Effectiveness of the Triplet Input of STFAN}
\label{sec: Effectiveness of the Triplet Input of STFAN}
To generate adaptive alignment and deblurring filters, STFAN takes the triplet input (previous blurry image $B_{t-1}$, previous restored image $R_{t-1}$, and current blurry image $B_t$).
Table~\ref{tab:input} shows the results of two variants which take ($B_{t-1}, B_{t}$) and ($R_{t-1}, B_{t}$) as input, respectively.
The triplet input leads to the best performance.
As Sec.~\ref{sec:arch} discussed, the network can implicitly capture the motion and model dynamic scene blur better from the triplet input.
\begin{table}[h]
\vspace{-1mm}
\centering
\caption{Effectiveness of using triplet input of the STFAN.
	We replace the input of the STFAN by ($B_{t-1}, B_{t}$) and ($R_{t-1}, B_{t}$) as two variants of our network ($R_{t-1}, B_{t-1}, B_{t}$), respectively.}
	\vspace{-1mm}
\resizebox{0.82\linewidth}{!} {
	\begin{tabular}{lcccc}
		\toprule
		Input       & ($B_{t-1},B_{t}$)  & ($R_{t-1}, B_{t}$)  & ($R_{t-1}, B_{t-1}, B_{t}$) \\
		\midrule
		PSNR         & 30.87               & 30.85              & \bf{31.24}   \\
		SSIM          & 0.930               & 0.930              & \bf{0.934}   \\
		\bottomrule
	\end{tabular}
}
\label{tab:input}
\vspace{-1mm}
\end{table}
\begin{table}[h]
	\centering
	\caption{Results of different sizes of adaptive filters.}
	\vspace{-1mm}
	\resizebox{0.8\linewidth}{!} {
		\begin{tabular}{lccccc}
			\toprule
			Filter Size    & $k=$ 3             & $k=$ 5             & $k=$ 7       & $k=$ 9\\
			\midrule
			PSNR          & 30.95              & 31.24       & 31.27            & 31.30  \\
			SSIM           & 0.931              & 0.934       & 0.934            & 0.935  \\
			\midrule
			Receptive Field & 79             & 87            & 95      & 103\\
			Params (M) & 4.58         &  5.37        & 6.56        &8.14\\
			\bottomrule
		\end{tabular}
	}
	\label{tab:filter_size}
	\vspace{-4mm}
\end{table}
\subsection{Effectiveness of the Size of Adaptive Filters}\label{sec:filter_size}
To further investigate the proposed network, we test different sizes of adaptive filters, shown in Table~\ref{tab:filter_size}.
The larger size of the adaptive filters leads to better performance.
However, increasing the size of adaptive filters after $k=5$ only has minor performance improvement.
We empirically set $k=5$ as a trade-off among the computational complexity, model size and performance.
\section{Conclusion}
We have proposed a novel spatio-temporal network for video deblurring based on filter adaptive convolutional (FAC) layers.
The network dynamically generate element-wise alignment and deblurring filters in order.
Using the generated filters and FAC layers, our network can perform temporal alignment and deblurring in the feature domain.
We have shown that the formulation of two spatially variant problems in video deblurring (i.e., alignment and deblurring) as two filter adaptive convolution processes allows the proposed method to
utilize features obtained at different time steps without explicit motion estimation (e.g., optical flow) and enables our method to handle spatially variant blur in dynamic scenes.
The experimental results demonstrate the effectiveness of the proposed method in terms of accuracy, speed as well as model size.
\clearpage
{\small
\bibliographystyle{ieee}
\bibliography{references}
}

\end{document}